\RequirePackage{fix-cm}
\documentclass[twocolumn]{svjour3}          
\smartqed  
\usepackage{framed,multirow}
\usepackage{latexsym}
\usepackage{url}
\usepackage{algorithm}
\usepackage{algpseudocode}
\usepackage{graphicx}
\usepackage{textcomp}
\usepackage{natbib}
\usepackage{multirow}
\usepackage{amsmath,amssymb,amsfonts}
\usepackage{bbm}
\usepackage{bm}
\usepackage{booktabs}
\usepackage{array}
\usepackage{blindtext}
\usepackage{comment}
\usepackage[breaklinks=true]{hyperref}
\usepackage{pifont}
\usepackage{makecell}
\usepackage{verbatim}

\usepackage[dvipsnames]{xcolor}
\usepackage{color, colortbl}
\newcommand{\myparagraph}[1]{\vspace{0.1em}\noindent\textbf{#1}}
\makeatletter
\renewcommand\paragraph{
  \@startsection{paragraph} 
  {4} 
  {\z@} 
  {.5em \@plus1ex \@minus.2ex} 
  {-.5em} 
  {\normalfont\normalsize\bfseries} 
}
\makeatother

\definecolor{delta}{RGB}{2, 167, 81}
\newcommand{\M}{Ctrl-GenAug} 

\begin{document}
\sloppy

\title{Ctrl-GenAug: Controllable Generative Augmentation for Medical Sequence Classification}

\author{Xinrui Zhou$^{1,2,4^\dagger}$ \and
        Yuhao Huang$^{1,2^\dagger}$ \and
        Haoran Dou$^{6}$ \and
        Shijing Chen$^{1,2}$ \and
        Ao Chang$^{1,2}$ \and
        Jia Liu$^{7}$ \and
        Weiran Long$^{7}$ \and
        Jian Zheng$^{8,9}$ \and
        Erjiao Xu$^{10}$ \and
        Jie Ren$^{7}$ \and
        Alejandro F. Frangi$^{6,12,13}$ \and
        Ruobing Huang$^{1,2}$ \and
        Jun Cheng$^{1,2}$ \and
        Xiaomeng Li$^{4,5}$ \and
        Wufeng Xue$^{1,2^*}$ \and
        Dong Ni$^{2,3,11^*}$
}

\authorrunning{Xinrui Zhou et al.}

\institute{
        $^1$ National-Regional Key Technology Engineering Laboratory for Medical Ultrasound, School of Biomedical Engineering, Medical School, Shenzhen University, Shenzhen, China \\
        $^2$ Medical UltraSound Image Computing (MUSIC) Lab, Shenzhen University, Shenzhen, China \\
        $^3$ School of Artificial Intelligence, Shenzhen University, Shenzhen, China \\
        $^4$ Department of Electronic and Computer Engineering, The Hong Kong University of Science and Technology, Kowloon, Hong Kong SAR, China \\
        $^5$ Department of Computer Science and Engineering, The Hong Kong University of Science and Technology, Kowloon, Hong Kong SAR, China \\
        $^6$ Department of Computer Science, Faculty of Science and Engineering, The University of Manchester, Manchester, UK \\
        $^7$ The Third Affiliated Hospital of Sun Yat-sen University, Guangzhou, China \\
        $^8$ Longgang District People’s Hospital of Shenzhen, Shenzhen, China \\
        $^9$ The Second Affiliated Hospital of The Chinese University of Hong Kong, Shenzhen, China \\
        $^{10}$ The Eighth Affiliated Hospital of Sun Yat-sen University, Shenzhen, China \\
        $^{11}$ School of Biomedical Engineering and Informatics, Nanjing Medical University, Nanjing, China \\
        $^{12}$ Division of Informatics, Imaging and Data Science, Faculty of Biology, Medicine, and Health, The University of Manchester, Manchester, UK\\
        $^{13}$ NIHR Manchester Biomedical Research Centre, Manchester Academic Health Science Centre, Manchester, UK \\
        $^*$Corresponding authors. $^\dagger$Authors contributed equally. \\
        \email{xuewf@szu.edu.cn, nidong@szu.edu.cn}
}

\date{Received: date / Accepted: date}
\maketitle

\begin{abstract}
In the medical field, the limited availability of large-scale datasets and labor-intensive annotation processes hinder the performance of deep models.
Diffusion-based generative augmentation approaches present a promising solution to this issue, having been proven effective in advancing downstream medical recognition tasks.
Nevertheless, existing works lack sufficient semantic and sequential steerability for challenging video/3D sequence generation, and neglect quality control of noisy synthesized samples, resulting in unreliable synthetic databases and severely limiting the performance of downstream tasks.
In this work, we present \textit{\M}, a novel and general generative augmentation framework that enables highly semantic- and sequential-customized sequence synthesis and suppresses incorrectly synthesized samples, to aid medical sequence classification.
Specifically, we first design a multimodal conditions-guided sequence generator for controllably synthesizing diagnosis-promotive samples.
A sequential augmentation module is integrated to enhance the temporal/stereoscopic coherence of generated samples.
Then, we propose a noisy synthetic data filter to suppress unreliable cases at the semantic and sequential levels.
Extensive experiments on 5 medical datasets with 4 different modalities, including comparisons with 15 augmentation methods and evaluations using 11 networks trained on 3 paradigms, comprehensively demonstrate the effectiveness and generality of \textit{\M}, particularly with pronounced performance gains in underrepresented high-risk populations and out-domain conditions.
Codes, models, and synthetic databases are available at \url{https://github.com/XinRuiZhou0106/Ctrl-GenAug}.


\end{abstract}

\section{Introduction}
Dynamic information in medical imaging (e.g., video sequences) plays a vital role in clinical diagnosis.
Recent deep learning-based classifiers have shown the ability to improve the diagnostic accuracy of different diseases~\citep{zhou2023inflated,wang2024screening}.
Despite rapid advancements, the performance of current advanced solutions is still limited by several issues:
1) The scarcity of dynamic clinical cases, coupled with the high annotation cost, limits data availability;
2) Imbalanced data distribution, driven by the rarity of high-risk positive cases, skews model training;
3) Deep models are prone to brittle performance degradation when tested on out-domain data, e.g., cases from different medical centers~\citep{varoquaux2022machine,huang2022online}.
This risk presents a significant obstacle to deploying models in real-world medical settings.
To address these issues, generative augmentation paradigms have employed advanced generative models to synthesize medical samples, thereby augmenting relevant diagnostic tasks.

Denoising diffusion probabilistic models (DDPMs), which utilize explicit likelihood estimation and progressive sampling, have more well-established mathematical explanations and abilities to achieve stable, controllable, and diverse data synthesis than previous generative methods like generative adversarial networks (GANs)~\citep{croitoru2023diffusion,dhariwal2021diffusion,luo2024measurement}.
Diffusion models have garnered remarkable success in several natural image fields, including static image generation~\citep{rombach2022high,zhan2023multimodal}, video synthesis~\citep{singer2022make,zhou2022magicvideo,ge2023preserve,ho2022video,wu2023lamp,wang2024boximator,ren2024consistiv,han2024vfusion3d}, video editing~\citep{khachatryan2023text2video,zhang2023controlvideo,qi2023fatezero}, and image animation~\citep{guo2024i2v,uzolas2024motiondreamer}. 
Recently, researchers have explored high-fidelity medical sequence synthesis by applying the above approaches within the medical field~\citep{zhou2024heartbeat,li2024endora}.

Most recently, pioneers deeply investigated the impact of the diffusion-based generative augmentation scheme on solving data scarcity~\citep{luo2024measurement,ktena2024generative} and domain generalization~\citep{ktena2024generative} issues in diagnostic tasks.
However, they failed to fully ensure the reliability of synthetic medical images for classification due to limited semantic steerability and absent quality control over the generated images.
Moreover, their techniques focused on image-level synthesis rather than sequence-level synthesis, which is crucial for medical modalities like MRI, ultrasound (US), etc~\citep{luo2024measurement}.
Thus, our study plans to devise a diagnosis-reliable controllable generative augmentation framework to facilitate accurate and robust medical sequence classification. 

\begin{figure*}[!t]
	\centering
	\includegraphics[width=0.95\linewidth]{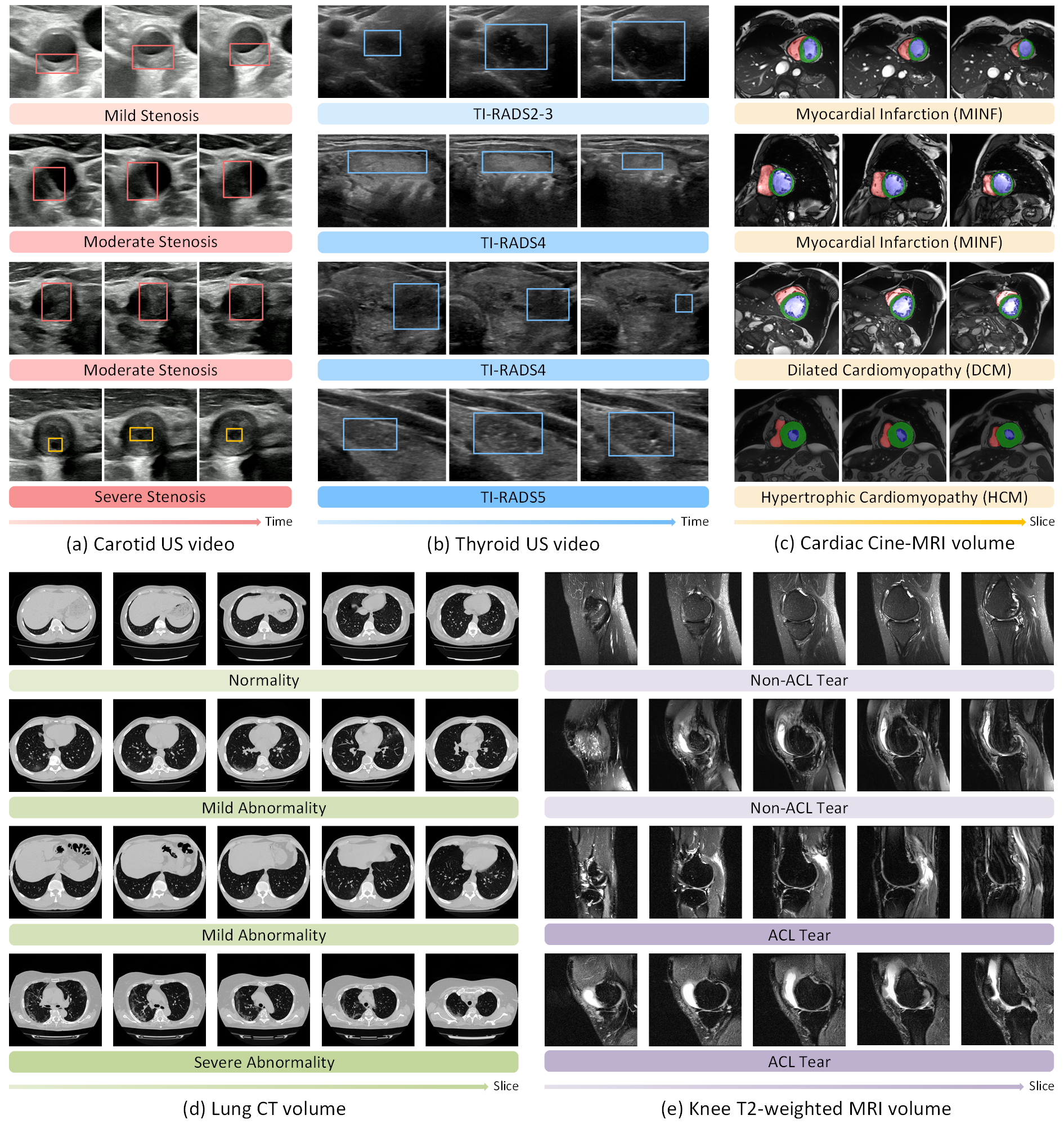}
	\caption{Datasets description: (a) Carotid US videos with various stenosis gradings. Red and yellow boxes represent plaques and residual lumens, respectively. (b) Thyroid US videos with different TI-RADS levels, where blue boxes indicate nodules. (c) Cardiac Cine-MRI volumes with distinct diseases. Three key anatomical structures associated with diagnosis are highlighted with masks, including the left ventricle (blue), myocardium (green), and right ventricle (red). (d) Lung CT volumes exhibiting varying degrees of tissue abnormalities with COVID-19. (e) Knee T2-weighted MRI volumes with and without anterior cruciate ligament (ACL) tears.}
	\label{fig:intro}
\end{figure*}

Intuitively, building customized, high-fidelity, and sequentially coherent synthetic medical databases, and effectively utilizing them, is essential for enhancing sequence recognition.
However, this task presents several challenges.
First, lesions (Fig.~\ref{fig:intro}(a-b)) or structures (Fig.~\ref{fig:intro}(c-e)) of the same disease category exhibit large visual variances (e.g., shapes, intensities, positions, etc).
This may seriously confuse diffusion learning, thus causing uncontrollable and unreliable generation. 
Second, artifacts and noises (e.g., US speckle noise) may prevent models from accurately perceiving vital anatomical targets, hindering the synthesis of high-fidelity sequences.
Third, the complex dynamic changes (Fig.~\ref{fig:intro}(a-b)) and the varying sizes of anatomical regions (Fig.~\ref{fig:intro}(c-e)) challenge the coherence of synthesized sequences.
The fourth challenge lies in the domain gap between synthetic and real samples, where achieving real-domain customization is crucial for effectively utilizing synthetic samples in downstream learning~\citep{he2023is}.
Lastly, even a well-designed sequence generator cannot always guarantee high-quality synthesis due to random sampling, and unsatisfactory synthetic samples may negatively impact subsequent classifier learning.
In summary, from model- and data-centric perspectives, the challenges of the task can be concluded as:
1) How can we design the generator architecture to achieve satisfactory sequence synthesis?
2) How can we discriminatively filter out potentially harmful synthetic samples?

In this work, we propose a novel controllable generative augmentation framework, named \textit{\M}, to facilitate medical sequence classification tasks on different organs and modalities.
Concretely, \textit{\M} synthesizes customized, high-quality sequences using conditional diffusion models and applies quality control to the generated sequences, to supplement medical datasets and improve classification.
This is a general method for enhancing medical sequence classification.
We believe this is the first comprehensive study to analyze the impact of controllable generative augmentation on multi-modal medical sequence classification.
Our contributions are three-fold:

\begin{itemize}
    \item 
    We propose a multimodal conditions-guided sequence generator to ensure controllable and high-fidelity synthesis.
    The introduced image prior serves as conditional guidance to empower the real-domain customization capacity.
    Moreover, we design a sequential augmentation module to promote dynamics modeling, thus improving the temporal or stereoscopic coherence of generated data.
    \item We introduce a noisy synthetic data filter to suppress harmful synthetic sequences at class semantics and sequential levels.
    By effectively reducing noise and enhancing the reliability of the synthetic databases, we can associate the synthetic samples with downstream tasks, thus better improving the classification performance.
    \item
    Comprehensive experiments on 5 medical datasets with 4 modalities (US, CT, Cine/T2-weighted MRI), including benchmarking against 15 augmentation methods and evaluations using 11 downstream classifiers trained on 3 paradigms, show that our approach consistently achieves superior performance across diverse diagnostic tasks, while maintaining robust effectiveness with multiple classifier architectures.
    \textit{\M} also improves diagnostic performance in underrepresented high-risk sets and enhances out-domain robustness, underscoring its practicality in real clinical scenarios.
\end{itemize}

It is noted that the sequence generator extends our prior work presented at MICCAI~\citep{zhou2024heartbeat}, with significant improvements in controllability at both semantic and sequential levels.
We highlight that customized semantics and consistent sequential dynamics are crucial for enhancing the learning capability of downstream classifiers.
Hence, we incorporate attribute texts and class labels to provide fine-grained semantic guidance, while employing a computation-efficient motion field to enhance dynamic control.
Besides, we propose a sequential augmentation module to further ensure the sequential consistency and smoothness of generated samples.

\section{Related Works}
In this part, we briefly review the controllable sequence synthesis approaches using diffusion models.
Given the strong inter-slice association in 3D medical volumes~\citep{zhu2024medical}, we treat them as video sequence data in our study and summarize related video-based methods.
Finally, we deeply involve existing diffusion-based generative augmentation schemes.

\subsection{Controllable Video Synthesis with Diffusion Models}
Diffusion models, a class of generative models, have recently attracted significant attention, especially text-to-video (T2V) synthesis~\citep{blattmann2023align,ho2022imagen,singer2022make,zhou2022magicvideo}.
However, these methods did not achieve precise control over the visual appearance and dynamics of the synthesized video, resulting in limited practicality.
To mitigate this issue, researchers have concentrated on incorporating multimodal conditions into T2V frameworks to guide controllable synthesis.
Specifically, AnimateDiff~\citep{guo2023animatediff} and MoonShot~\citep{zhang2024moonshot} introduced text and image conditions to control the visual appearance.
Most recently, several studies~\citep{zhang2023controlvideo,zhao2023controlvideo,wang2023videocomposer} achieved fine-grained spatial compositional control by integrating additional diverse conditions (i.e., sketch, mask, and depth sequences).
Meanwhile, two spatial-sequential (S$^2$) modeling strategies (fully attention layers~\citep{zhang2023controlvideo} and S$^2$ condition encoder~\citep{wang2023videocomposer}) were introduced to capture the dynamics of sequential conditions, promoting the inter-frame features interaction. 
Besides, dense optical flow~\citep{zach2007duality} was adopted to control sequential consistency~\citep{liang2023flowvid,ni2023conditional}.
The above methods have shown promising controllability in natural video synthesis.
However, for medical synthesis, acquiring abundant dense annotations (e.g., mask sequences) as control signals is impractical and unaffordable.
Additionally, medical videos have unique attributes compared with natural ones, such as blurred anatomical regions and complex structure variations.
Hence, they may not suit the medical video synthesis tasks.

\begin{figure*}[!htbp]
	\centering
	\includegraphics[width=1.0\linewidth]{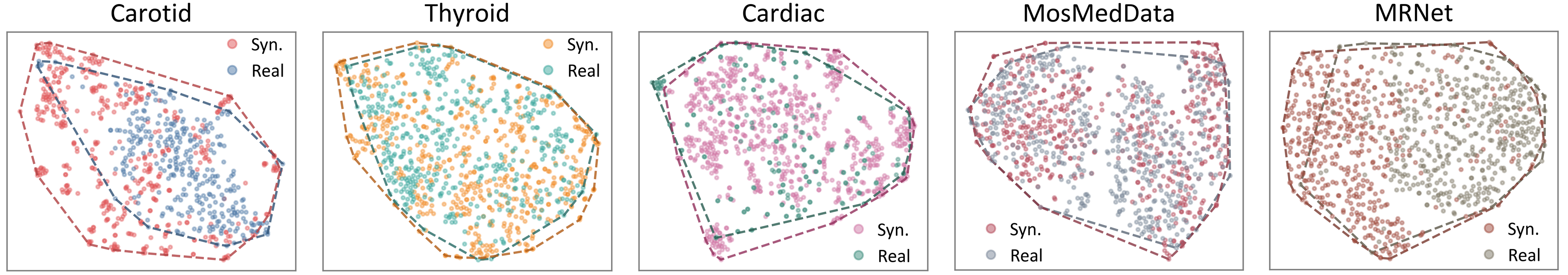}
	\caption{t-SNE visualization of features of synthetic and real training data by a pre-trained I3D~\citep{carreira2017quo} model on our five datasets. Our \textit{\M} produces synthetic samples with diverse feature patterns that enrich the training distribution.}
	\label{fig:tsne}
\end{figure*}

In the medical field, several diffusion-based methods have been proposed to generate photorealistic echocardiography videos using the conditional guidance of a single frame~\citep{reynaud2023feature}, ejection fractions~\citep{reynaud2023feature}, single semantic mask~\citep{van2024echocardiography,nguyen2024training}, and single sketch image~\citep{zhou2024heartbeat}.~\cite{zhou2024heartbeat} additionally leveraged mitral valve (MV) skeletons to control the complex motion trajectories of MV in the generated echocardiograms.
Besides,~\cite{li2024endora} proposed to generate endoscopy videos that simulate clinical scenes by integrating a spatial-temporal transformer with 2D vision foundation model priors.
However, the above methods may have the following drawbacks: 
unsteerable semantics due to the lack of attribute guidance (e.g., category or target shape), poor control over sequences, etc.
Thus, they may be unsuitable for direct adoption in the generative augmentation tasks.

\subsection{Generative Augmentation with Diffusion Models}
Synthetic data can provide intrinsically diverse characteristics and visual appearances~\citep{luo2024measurement,pan20232d}, thus enhancing model learning.
This aligns with the observation shown in Fig.~\ref{fig:tsne}.
Previous works have succeeded in utilizing synthetic data by diffusion models to enhance natural image classification~\citep{he2023is,Sariyildiz_2023_CVPR,trabucco2023effective,azizi2023synthetic}, segmentation~\citep{sankaranarayanan2018learning,wu2023diffumask,yang2024freemask}, and detection~\citep{feng2024instagen}.
Recently,~\cite{singh2024synthetic} performed an in-depth analysis of models trained with synthetic data across various robustness measures, and verified that they achieved good suitability in real-world settings.

To address data scarcity and promote diagnostic performance, the intelligent medical field has raised increasing interest in exploring diffusion-based generative augmentation for aiding different downstream diagnostic tasks~\citep{shang2023synfundus,farooq2024derm,ktena2024generative,zhang2024diffboost,dorjsembe2024polyp,li2024fairdiff}. \cite{luo2024measurement} marked the first focus on comprehensively analyzing the impact of their designed uncertainty-guided diffusion models on downstream diagnostic tasks.
Nevertheless, current studies have two main limitations:

On the one hand, due to the lack of adequate semantic and sequential guidance during sampling, coupled with the absence of post-sampling quality control for generated samples, existing works could hardly fully ensure the reliability of synthetic databases for downstream disease recognition.
For instance, descriptive text (e.g., lesion morphology) can provide informative guidance related to disease grading.
Unlike previous solutions, we aim to develop a generative augmentation framework that enables highly semantic- and sequential-customized sequence synthesis and suppresses incorrectly synthesized samples to better facilitate downstream classifier learning.

On the other hand, few studies comprehensively explore the effect of generative models on downstream medical tasks, and they solely focus on image-level generation.
Considering the value of dynamic information for clinical diagnosis, we 
concentrate on diagnosis-reliable sequence synthesis.
This is more challenging than image-related works as synthetic data not only meets high-fidelity requirements but also ensures temporal or stereoscopic consistency and coherence for effectively promoting downstream S$^2$ modeling capacity.
A brief review of studies in the medical field utilizing diffusion models to promote downstream tasks can be found in the \textbf{Supplementary Material}.

\section{Method}
\begin{figure*}[!t]
    \centering
    \includegraphics[width=1.0\linewidth]{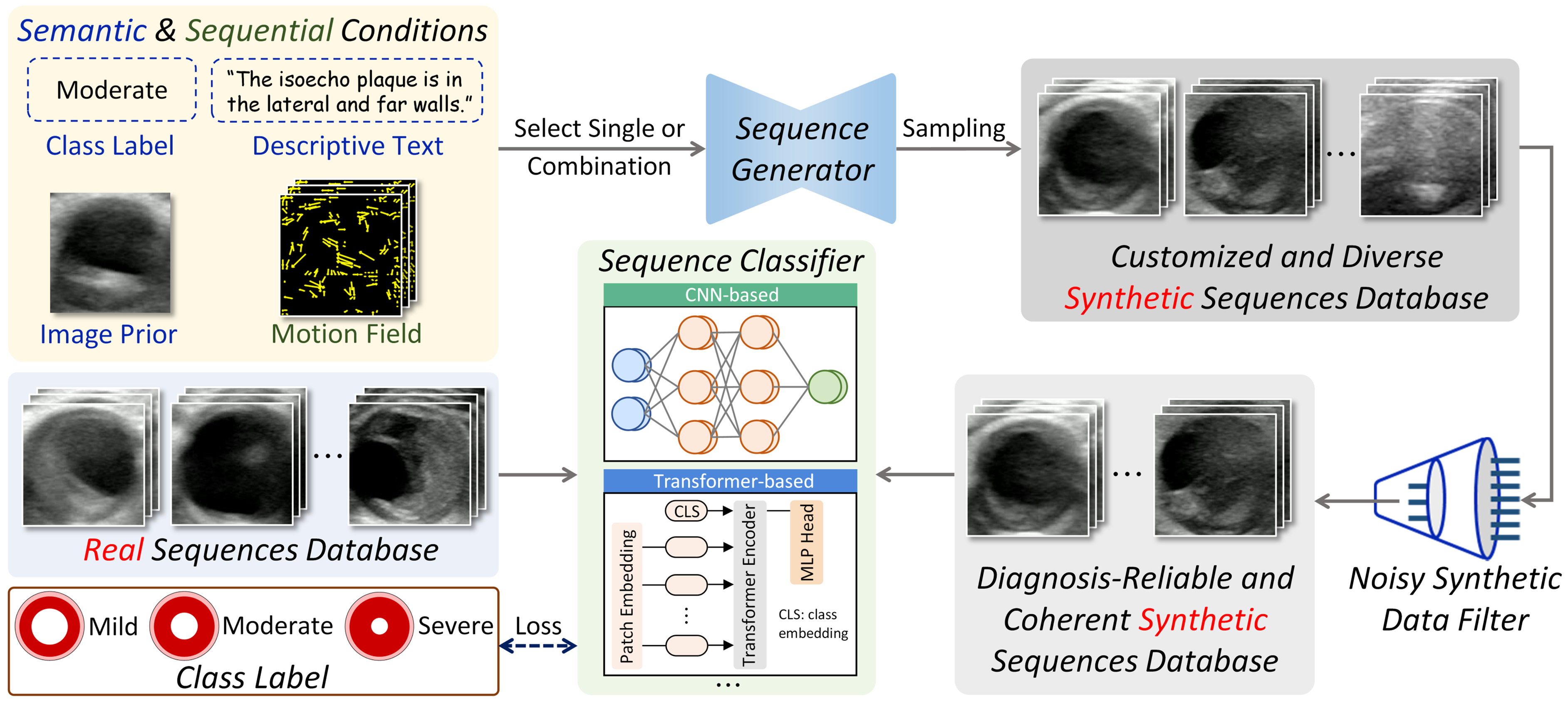}	
    \caption{Pipeline of using our proposed framework to facilitate medical sequence recognition, which can be worked with a variety of classifiers. Here, we use the carotid plaque US video sequence as an example to demonstrate the overall process.}
    \label{fig:framework}
\end{figure*}

\subsection{Model Overview}
Fig.~\ref{fig:framework} shows the pipeline of using our proposed controllable generative augmentation framework to aid medical sequence classification.
Our augmentation system is composed of two key components, including a sequence generator and a noisy synthetic data filter.
First, the sequence generator perceives multiple semantic (including class label, descriptive text, and image prior) and sequential conditions (i.e., motion field) to produce customized and diverse synthetic databases.
Then, the noisy synthetic data filter is employed to effectively suppress unsatisfactory generated sequences, resulting in reliability-enhanced and coherent synthetic databases for downstream diagnosis.
In the application stage, the quality-controlled synthetic sequences and real ones work together to improve the performance of arbitrary classifiers.
To formulate customized, high-fidelity, and coherent synthetic databases to boost classification, we design the whole framework from model-centric, i.e., sequence generator (Secs.~\ref{sec:3B},~\ref{sec:3C}, and~\ref{sec:3D}) and data-centric, i.e., data filter (Sec.~\ref{sec:3E}) perspectives.

\subsection{Basic Architecture of Sequence Generator}
\label{sec:3B}
\begin{figure*}[!t]
    \centering
    \includegraphics[width=1.0\linewidth]{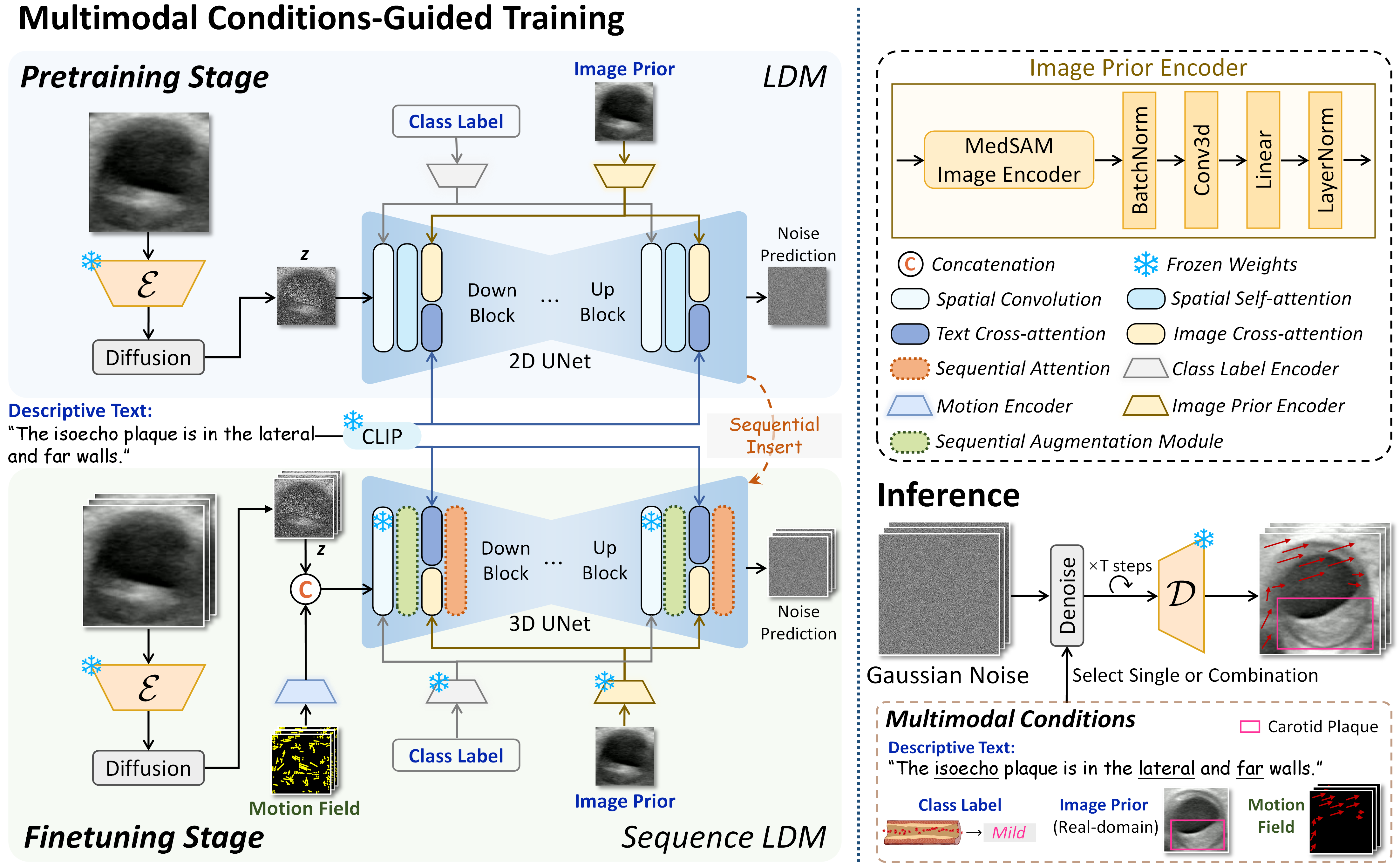}	
    \caption{Pipeline of our proposed sequence generator. The MedSAM image encoder~\citep{huang2024segment} is used for domain-specific image prior feature extraction. During inference, given a bank of customized conditions (e.g., descriptive text with underlined carotid plaque \underline{attributes}, the class label \textit{mild}, image prior, and motion field), our generator produces a real-domain style sequence that faithfully adheres to all specified conditions.}
    \label{fig:generator}
\end{figure*}

Fig.~\ref{fig:generator} shows the pipeline of our proposed sequence generator. 
It supports customized and high-quality medical sequence generation via multimodal conditions, using a two-stage training scheme.
In the pretraining stage, it attends to high-fidelity \textit{visual features learning} for controllable image synthesis.  
While in the finetuning stage, the domain-specific visual knowledge acquired from the previous stage is reused and focuses on \textit{sequential patterns modeling} for customized sequence synthesis. 
During inference, given a single or combination of multimodal conditions as control signal inputs, high-quality and steerable sequence generation from Gaussian noise can work.
We then provide a short background of video diffusion models, followed by a detailed description of the basic architecture of our generator.

\myparagraph{Preliminaries of Video DDPMs.}
\label{ddpm_background}
Video DDPMs aim to learn complex distributions by iteratively denoising corrupted inputs using diffusion methods~\citep{sohl2015deep, ho2020denoising}.
Given a video $\mathit{x}_0\sim\mathit{q}$ and timestep $\mathit{T}$, DDPMs first produce a sequence of noisy inputs via diffusion process $\mathit{q}(\mathit{x}_t|\mathit{x}_0, \mathit{t}), \mathit{t} \in {1, 2, ..., \mathit{T}}$, which progressively adds Gaussian noise $\epsilon \in \mathcal{N}(\mathbf{0}, \mathbf{I})$ to $\mathit{x}_0$.
To ease the computational burden of traditional pixel-space training, video latent diffusion models~\citep{blattmann2023align} perform the diffusion process in latent space of a variational autoencoder~\citep{esser2021taming} (VAE).
The model is then trained to estimate the parameterized Gaussian transition $\mathit{p}(\mathit{x}_{t-1}|\mathit{x}_t)$ via a denoising network $\theta$.
Mathematically, the optimized objective can be a simplified variant of the variational lower bound:
\begin{equation}
    \begin{split}
        \min_\theta \mathbb{E}_{\mathit{z}_0,\epsilon\sim\mathcal{N}(\mathbf{0}, \mathbf{I}), \mathbf{c}, \mathit{t}}||\epsilon - \epsilon_\theta(\mathit{z}_t, \mathbf{c}, \mathit{t})||^2,
    \end{split}
\end{equation}	
where $\mathit{z}_0$ is the latent code of $\mathit{x}_0$.
$\epsilon_\theta$ and $\epsilon$ represent the predicted and target noise, respectively.
$\mathbf{c}$ denotes the control signal (optional). 
For our generator, $\mathbf{c}$ involves multimodal conditions including semantic and sequential ones.

\myparagraph{Factorized Learning of Visual Features and Sequential Patterns.}
\label{factorized_learning}
Developing a generator with pure 3D architecture design is an intuitive scheme for sequence synthesis~\citep{blattmann2023align}.
However, due to the scarcity and complexity of medical data, learning visual features and sequential patterns at the same time is very challenging and imposes high training costs.
To this end, we propose to factorize both learning by first pretraining a latent diffusion model (LDM) and then finetuning its sequence counterpart.
In this way, the generator enables realistic sequence synthesis, while easing the training burden.

\begin{figure}[!h]
	\centering
	\includegraphics[width=0.99\linewidth]{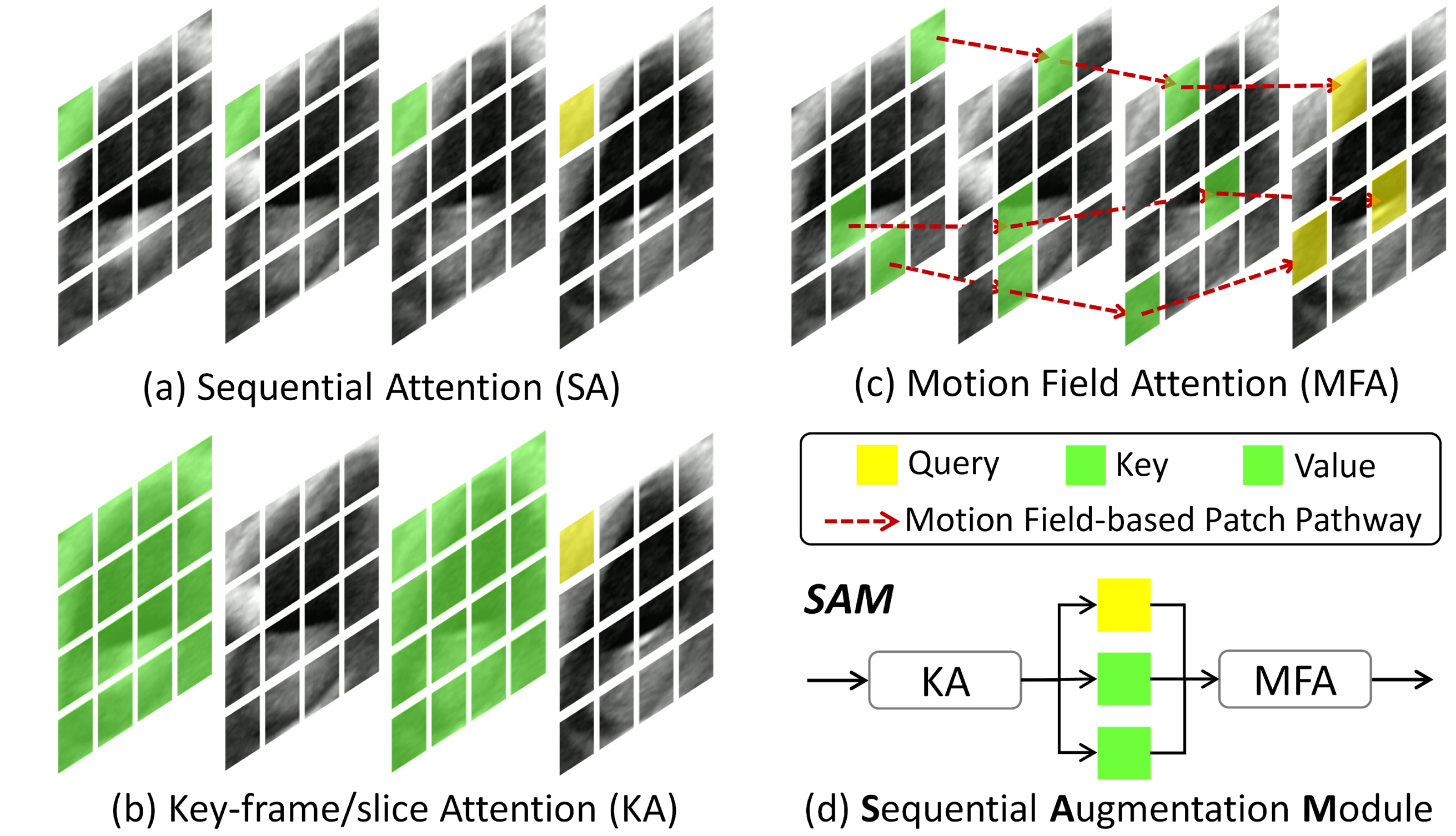}
	\caption{Schematics of three attention mechanisms for sequential modeling (a-c) and our proposed sequential augmentation module in this study (d).}
	\label{fig:attn}
\end{figure}

\myparagraph{2D-to-3D Model Inflation.}
\label{model_inflation}
During pretraining, we develop a 2D UNet~\citep{ronneberger2015u} in LDM to predict noises for image synthesis.
Referring to~\citep{ho2022video}, we extend the 2D UNet to a 3D version via an inflation scheme to build the sequence LDM.
Specifically, all the spatial convolution layers are inflated to pseudo-3D counterparts by expanding the kernels at the sequential dimension (e.g., $3\times3 \rightarrow 1\times3\times3$ kernel).
Besides, we perform sequential insertion by adding sequential attention (SA) layers (see Fig.~\ref{fig:generator}).
Fig.~\ref{fig:attn}(a) shows the principle of the SA mechanism, where each patch queries to those at the same spatial position and across frames/slices.
This design encourages the generator to model the sequential patterns, while not significantly altering visual feature distribution baked in the LDM~\citep{zhang2024moonshot}.
Overall, sequence LDM enables inheriting the rich visual concepts preserved in LDM and focusing on sequential pattern aggregation, making the model learning efficient.

\subsection{Multimodal Conditions Guidance}
\label{sec:3C}
Sequence generators guided by a single condition (e.g., text) have the following drawbacks: 
1) poor controllability and  2) sequential inconsistency. 
To solve these issues, we introduce multimodal conditions to guide our sequence generator for customized, realistic, and sequential-consistent synthesis.

We consider four multimodal conditions to ensure comprehensive and accurate control over the sequence synthesis procedure (see Figs.~\ref{fig:framework}-\ref{fig:generator}).
Specifically, these conditions are divided into semantic and sequential ones for visual appearance control and serial guidance, respectively.
We highlight that our generator supports composable synthesis by allowing users to flexibly choose any single condition or composition during inference. 
This flexibility makes our generator particularly user-friendly, as it enables high-quality synthesis even when certain conditions are missing.
Details are described below.

\myparagraph{Semantic Conditions for Visual Appearance Control.}
As shown in Fig.~\ref{fig:generator}, the sequence generator exploits three semantic conditions to perform visual appearance control, thus achieving controllable and high-fidelity synthesis.

- \textit{Descriptive Text:} 
It provides an intuitive indication of the coarse-grained semantic concepts of sequences.  
In our implementation, we pre-align the visual and textual features by finetuning the widely used CLIP model~\citep{radford2021learning} using our medical datasets to mitigate the large gap between them.
Then, semantic information of texts can feasibly query relevant medical visual patterns, thus easing subsequent training.
Similar to~\citep{rombach2022high}, the text guides the generator via cross-attention layers.

- \textit{Class Label:}
Generating medical data with the expected disease class is crucial for enhancing downstream classification, particularly improving the diagnostic accuracy of underrepresented high-risk sets.
Thus, we propose encouraging the model to focus on the grading signal.
Specifically, rather than merging the class label with descriptive text, we treat it as a separate tag to directly provide disease-specific guidance, as merging with text may dilute the class signal.
We further propose a dedicated insertion method for this purpose.
First, similar to the timestep embedding process~\citep{rombach2022high} in diffusion models, we adopt a class label encoder composed of a label discretization layer and an embedding layer to obtain its embeddings.
The addition between the timestep and class label embeddings is then input to the spatial convolution layer in each UNet block for visual appearance control.

- \textit{Image Prior:}
Merely using the above conditions faces challenges of insufficient semantic control and an inevitable domain gap between real and synthetic sequences, constraining the capability of generative augmentation.
Hence, we introduce the first frame/slice of real-domain sequences as image priors to provide rich semantic guidance and yield real-domain style sequences.
As image prior offers global semantics like texts, we align image prior features $f_{i}$ extracted from image prior encoder (see Fig.~\ref{fig:generator}) with text embeddings $f_{t}$ generated by our medical data-specific CLIP text encoder for joint guidance.
Specifically, motivated by~\citep{ye2023ip}, we replace the text-guided cross-attention layers in original UNet blocks with decoupled counterparts that handle texts and image priors in parallel and then merge the results by addition.
The decoupled attention can be formulated as:
\begin{equation}
    \left\{
    \begin{aligned}
        & \mathbf{Q_{g}}=W^{Q_{g}}f_{g}, \mathbf{K_{g}}=W^{K_{g}}f_{g}, \mathbf{V_{g}}=W^{V_{g}}f_{g}, g\in\left\{\mathbf{t},\mathbf{i}\right\}, \\
        & Attention\left(\mathbf{Q_{g}},\mathbf{K_{g}},\mathbf{V_{g}}\right)=\sum_{g}\left[\sigma(\frac{\mathbf{Q_{g}K_{g}^{\top}}}{\sqrt{d}}\mathbf{V_{g}})\right],
    \end{aligned}
    \right.
\end{equation}
where $W^{Q}, W^{K}, W^{V}$ denote the trainable linear projection matrices.
$\mathbf{t}, \mathbf{i}$ represent text and image prior conditions. $\mathit{d}$ denotes dimension of latent features. 
$\sigma(\cdot)$, softmax function.

\myparagraph{Sequential Condition for Serial Guidance.}
\label{sequential_condition}
Apart from perceiving visual concepts, modeling sequential knowledge is also important in sequence synthesis, with sequential cues playing a vital role.
Most existing methods~\citep{liang2023flowvid, ni2023conditional} adopted dense optical flow~\citep{zach2007duality} to promote sequential dynamics modeling. 
However, optical flow extraction requires high computational demands~\citep{wang2023videocomposer}.
Thus, we instead introduce computation-efficient \textit{motion field} as a sequential condition in sequence LDM.
It explicitly showcases the pixel-wise motions between adjacent frames/slices (Figs.~\ref{fig:framework}-\ref{fig:generator}).
In our approach, we first extract the motion fields of real sequences using a Python package~\citep{9248145}.
Then, a motion encoder (Fig.~\ref{fig:generator}) receives the motion fields and produces motion features.
Last, the features are concatenated with latent representations $\mathit{z}$ from VAE along the channel dimension for serial guidance.

\subsection{Sequential Augmentation Module}
\label{sec:3D}
Solely equipping with SA layers and sequential cues in sequence LDM may present inadequate consistency and coherence across synthetic frames/slices.
This issue may arise from insufficient sequential modeling of input noisy latents and motion fields, constrained by a small parameter space, overly burdening the SA layers.
To solve the issue, we propose a sequential augmentation module (SAM) that enables the generator to more effectively model sequential dependencies.
As shown in Fig.~\ref{fig:attn}(d), SAM integrates two attention mechanisms in cascade for sequential augmentation.

\myparagraph{Key-frame/slice Attention.}
The common spatial self-attention layers can lead to sequential inconsistency due to the lack of interaction across frames/slices.
To augment \textit{sequential consistency}, we introduce a key-frame/slice attention (KA) mechanism, where two selected key-frames/slices act as references to propagate S$^2$ information throughout the sequence.
Specifically, for any frame/slice, we select its previous and the first counterpart of the sequence as the references and transform spatial self-attention into KA, aiming to align the latent features $z_{l}$ of the \textit{l}-th frame/slice with $z_{1}$ and $z_{l-1}$.
We obtain query from $z_{l}$, key and value features from $z_{1}$ and $z_{l-1}$, and compute $Attention\left(\mathbf{Q},\mathbf{K},\mathbf{V}\right)$ using:
\begin{equation}
    \begin{split}
        \mathbf{Q}=W^{Q}z_{l}, \mathbf{K}=W^{K}\left\{z_{1},z_{l-1}\right\}, \mathbf{V}=W^{V}\left\{z_{1},z_{l-1}\right\},
    \end{split}
\end{equation}
where $W^{Q}, W^{K}, W^{V}$ are initialized on the original spatial self-attention weights for inheriting the semantic perception capability of LDM in the finetuning stage. 
$\left\{\cdot\right\}$ represents concatenation operation. 
It is highlighted that KA retains low computational complexity compared with full attention~\citep{zhang2023controlvideo}.
Please refer to Fig.~\ref{fig:attn}(b) for a visual illustration.

\myparagraph{Motion Field Attention.}
To further boost the \textit{sequential coherence} of the generated sequences, we propose to reuse the motion field condition (refer to Sec.~\ref{sequential_condition}) by introducing a motion field attention (MFA) mechanism after KA.
As shown in Fig.~\ref{fig:attn}(c), MFA requires the patches to communicate with those in the same motion field-based pathway including itself, which eliminates flickers of the generated sequences to make the contents visually smooth.
Inspired by~\citep{cong2024flatten}, MFA is implemented in two steps: a) motion field-based patch pathway sampling and b) attention calculation.
In \textbf{step 1}, we sample the patch pathways based on the \textit{m}-scaled downsampled motion fields.
Take a video sequence as an example, for a patch $p_{l}$ on the \textit{l}-th frame of a \textit{f}-frame video, the $path$ can be derived from the motion field.
Since the sampling procedure inevitably generates multiple pathways for the same patch, we randomly sample a pathway to ensure its uniqueness to which each patch belongs.
In this setting, let $H,W$ the height and width of the input video frame, then the size of the pathway set after sampling equals $\frac{H\times W}{m^{2}}\times f$.
In \textbf{step 2}, we calculate $Attention(\mathbf{Q},\mathbf{K},\mathbf{V})$ with:
\begin{equation}
    \begin{split}
        \mathbf{Q}=z_{p_{l}}, \mathbf{K}=\mathbf{V}=z_{path},
    \end{split}
\end{equation}
where $z_{path}$ represents the latent features of patches on the $path$, as follows:
\begin{equation}
    \begin{split}
        z_{path}=\left[z_{p_{0}},z_{p_{1}},\dots,z_{p_{l}},\dots,z_{p_{f-1}},z_{p_{f}}\right].
    \end{split}
\end{equation}
The output of KA is directly fed into MFA without being handled by the query, key, and value projection functions. Hence, MFA can comfortably decrease extra computation.

\begin{algorithm}[!t]
    \caption{Noisy synthetic data filtering.}
    \small
    \label{alg:data_filter}
    \begin{algorithmic}[1]
    \Require synthetic samples set $\mathrm{S}$, real data-trained classifier $p$.
    \State \# Stage 1: Class semantics misalignment filtering
    \State $\mathrm{S_1}=\varnothing$ \Comment{synthetic samples set obtained after stage 1}
    \For {$i \in \left\{1,...,n\right\}$}
        \State $c \gets$ class in the conditions bank of the group $\mathrm{S}[i]$
        \State $l_x = -\sum c\mathrm{log}\sigma(p(\mathrm{S}[i][x]))$ \Comment{loss of the sample indexed $x$}
        \State $L_{c}=\frac{1}{M}\sum_{x=1}^{M}l_{x}$ \Comment{noise identification threshold}
        \Statex \Comment{$M$ clips in the group}
        \For {$x \in \left\{1,...,M\right\}$}
            \If {$l_{x}>L_{c}$}
                \State $continue$
            \Else
                \State $\mathrm{S_1} \gets \mathrm{S_1}\cup\{\mathrm{S}[i][x] \}$ \Comment{save less noisy sample}
            \EndIf
        \EndFor
    \EndFor
    \State \textbf{return} $\mathrm{S_1}$ \Comment{with $n_1(< n)$ groups totaling $N_1(< N)$ clips}
    \State \# Stage 2: Inner-sequence filtering
    \State $\mathrm{S_{2}}=\varnothing$ \Comment{synthetic samples set obtained after stage 2}
    \Statex $\mathrm{A}=\left\{A_1,...,A_{N}\right\}$ \Comment{VAE-Seq values set of $\mathrm{S}$}
    \Statex $\mathrm{B}=\left\{B_1,...,B_{N_1}\right\}$ \Comment{VAE-Seq values set of $\mathrm{S_1}$}
    \State $t_{l},t_{h}\gets\mathbf{KMeans}(\mathrm{A}, \mathbf{K}=4)$ \Comment{boundary thresholds}
    \For {$j \in \left\{1,...,N_1\right\}$}
        \If {$B_j\ge t_l\And B_j\le t_h$}
            \State $\mathrm{S_2} \gets \mathrm{S_2}\cup\{y_j|y_j \in \mathrm{S_1} \}$ \Comment{save gentle dynamic sample}
        \Else
            \State $continue$
        \EndIf
    \EndFor
    \State \textbf{return} $\mathrm{S_2}$ \Comment{with $n_2(< n_1)$ groups totaling $N_2(< N_1)$ clips}
    \State \# Stage 3: Inter-sequence filtering
    \State $\mathrm{S_{3}}=\varnothing$ \Comment{synthetic samples set obtained after stage 3}
    \For {$k \in \left\{1,...,n_2\right\}$}
        \State $\mathrm{S_3} \gets \mathrm{S_3}\cup\mathrm{S_2}[k][1]$ 
        \For {$q \in \left\{2,...,M'\right\}$} \Comment{$M'$ clips in the group $\mathrm{S_2}[k]$}
        \If {$\forall z \in \mathrm{S_3},\Theta\footnotemark (\mathrm{S_2}[k][q],z)<98$}
            \State $\mathrm{S_3} \gets \mathrm{S_3}\cup\mathrm{S_2}[k][q]$ \Comment{save diverse sample}
        \Else
            \State $continue$
        \EndIf
        \EndFor
    \EndFor
    \State \textbf{return} $\mathrm{S_3}$ \Comment{final synthetic database for downstream training}
    \end{algorithmic}
\end{algorithm}
\footnotetext{$\Theta$ calculates inter-clip similarity.}

\subsection{Noisy Synthetic Data Filter}
\label{sec:3E}
We use the proposed sequence generator to constitute the synthetic samples set.
Concretely, assuming there are $n$ training clips in the target dataset, with a bank of conditions derived from each clip, we synthesize a group of clips guided by each conditions bank.
Eventually, we obtain $n$ groups of clips to form our synthetic sample set, with a total of $N$ clips.

\begin{figure*}[!t]
	\centering
    \includegraphics[width=1.0\linewidth]{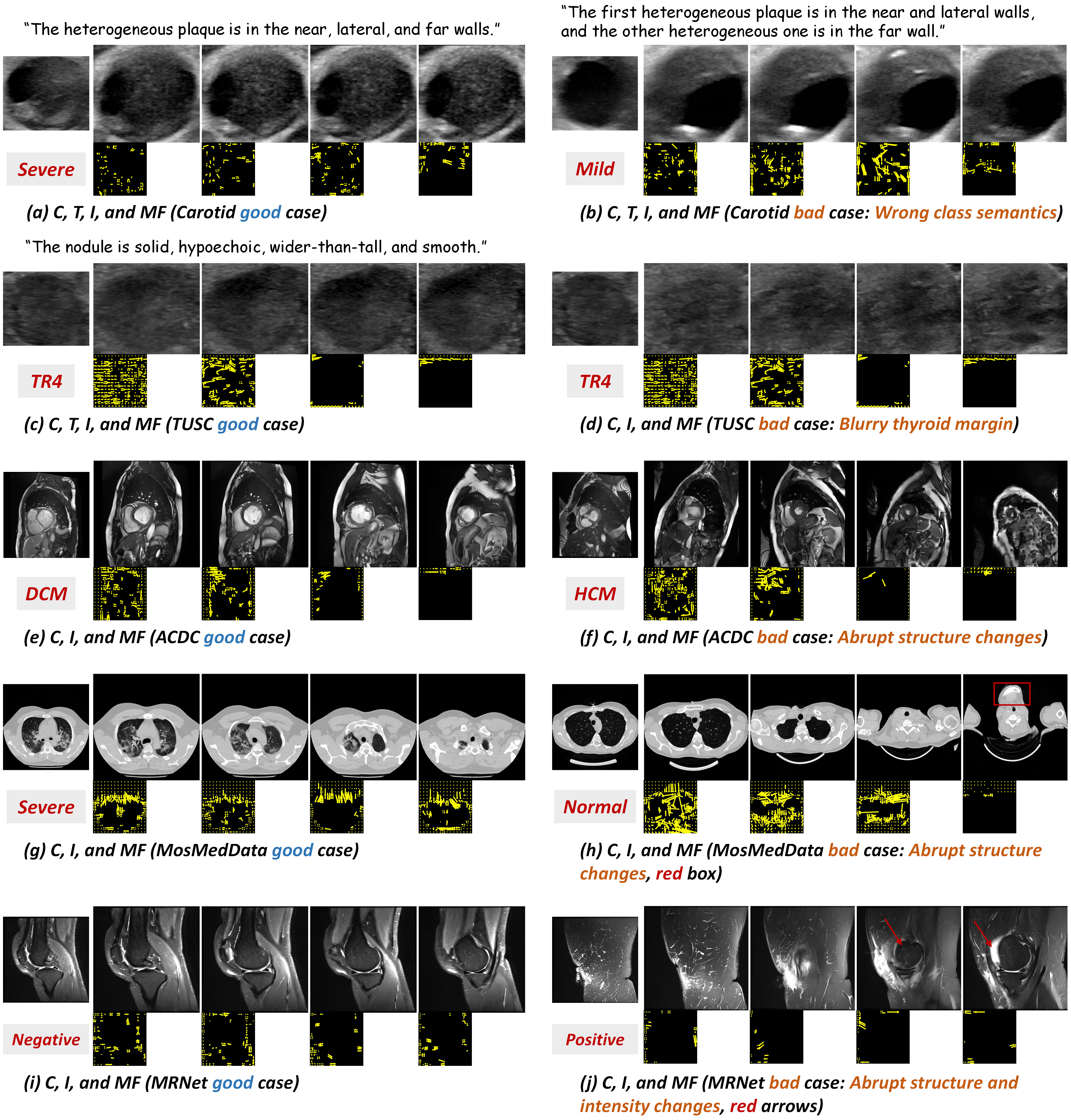}
	\caption{Typical good and bad synthetic samples of five datasets using different banks of conditions, with representative failure patterns highlighted. Conditional controls include class label (C), text (T), image prior (I), and motion field (MF). See more synthetic results in \textbf{Supplementary Material}.}
	\label{fig:syn_visual}
\end{figure*}

Although visually realistic and smooth (good cases, Fig.~\ref{fig:syn_visual}), sequence synthesis may still suffer from class semantics misalignment, cross-frame/slice inconsistency or over-consistency (i.e., almost static clip), and inter-clip similarity.
For instance, in Fig.~\ref{fig:syn_visual}(b), the synthesized carotid clip category is wrong, which should be moderate rather than mild.
In Fig.~\ref{fig:syn_visual}(f, h, j), the generated clip includes abrupt changes in anatomical structures.
Hence, blindly using all synthetic clips for classifier learning will significantly cause a performance drop due to noisy samples. 
Our work proposes a noisy synthetic data filter to adaptively remove harmful generated clips at class semantics and sequential levels, as illustrated in Algorithm~\ref{alg:data_filter}.
This can also link the synthetic data to downstream tasks, thus potentially achieving a higher performance upper bound.

\myparagraph{Class Semantics Misalignment Filtering.}
\label{semantic_filter}
This scheme is designed to filter the generated clips whose visual contents mismatch the corresponding class labels.
Motivated by anomaly detection~\citep{kwon2020backpropagated}, we consider that such noisy synthetic samples will produce large losses if fed into a powerful classifier trained on real ones.
Hence, a loss threshold can be set to identify and filter the anomalistic clips.
Specifically, we first adopt a well-trained classifier in the real data domain to calculate case-wise cross-entropy loss values $\left\{l_{x}\right\}_{x=1}^{M}$ based on corresponding category $c$ for all synthetic clips in each group.
Then, we set the average loss as the noise identification threshold of $c$.
Last, we conduct class semantics misalignment filtering at the group level.
If the loss of a synthetic clip with $c$ exceeds the threshold $L_{c}$, it will be viewed as a noisy sample and excluded from the synthetic pool.
In our experiments, the classifier employed for semantic filtering is identical to the downstream classifier by default.

\myparagraph{Sequential Filtering.}
\label{sequential_filtering}
This strategy aims to filter noisy samples at a) inner-sequence and b) inter-sequence levels. 
\textit{For a)}, we screen out synthetic clips with gentle dynamic, avoiding those that are either inconsistent or over-consistent affecting downstream learning.
Specifically, we retain a synthetic clip whose cross-frame/slice consistency falls in a pre-computed range.
It is determined by K-means clustering~\citep{macqueen1967some} based on the cross-frame/slice consistency of all synthetic clips.
To assess the video sequence consistency,~\cite{wang2023videocomposer} tends to use CLIP~\citep{radford2021learning} image embeddings to compute the average cosine similarity across consecutive frames.
However, this method leads to inadequate evaluation due to the usage of limited informative CLIP embeddings. 
To resolve the problem, we propose a metric termed VAE-Seq that utilizes latent embeddings from pretrained VAE, instead of deriving embeddings from CLIP space, to assess cross-frame/slice consistency.
Therefore, with the higher dimension of latents (e.g., 4$\times$32$\times$32 for input size 256$\times$256 compared to 768 in CLIP embeddings), VAE-Seq reflects an accurate evaluation using more fine-grained and informative features.
\textit{For b)}, we seek to diversify the samples in the synthetic set to prevent overfitting and avoid wasting computational resources in downstream training. 
To this end, we perform the inter-sequence filtering based on inter-clip similarity in each group, as shown in Algorithm~\ref{alg:data_filter}.
We quantify the inter-clip similarity by calculating the cosine distance of VAE latent embeddings between frame/slice pairs in two clips and averaging the obtained distance values.

\section{Experiments}
We first benchmarked \textit{\M} against 15 augmentation methods across 5 datasets with 4 medical modalities to comprehensively validate its effectiveness and generality.
We further systematically assessed its downstream robustness using 11 popular classifiers trained under 3 paradigms.
We then evaluated its performance in underrepresented high-risk sets and out-domain conditions.
Finally, we conducted extensive ablation studies to examine the contributions of each component and the behavior of \textit{\M} under different diffusion settings, backbone selections in the semantics misalignment filter, and varying annotation requirements.

\begin{table*}[!htbp]
    \centering
    \caption{Experimental datasets and settings.}
    \label{tab:dataset_setting}
    \resizebox{1.0\textwidth}{!}{
    \begin{tabular}{c c c c c c}
        \toprule
        & Carotid & TUSC & ACDC & MosMedData & MRNet \\
        \midrule
        Modality & US & US & Cine-MRI & CT & T2-weighted MRI \\
        Organ & Carotid artery & Thyroid & Heart & Lung & Knee \\
        Patient & 231 & 167 & 150 & 1,110 & 1,199 \\
        Clip & 486 & 633 & 314 & 2,573 & 1,958 \\
        \midrule
        & \textcolor{gray}{Mild/Moderate/Severe} & \textcolor{gray}{TI-RADS2-3/4/5} & \textcolor{gray}{NOR/MINF/DCM/HCM/ARV} & \textcolor{gray}{Normal/Mild/Severe} & \textcolor{gray}{Negative/Positive} \\
        Class (train) & 141/104/64 & 146/189/113 & 40/40/40/40/45 & 512/1,226/317 & 1,449/320 \\
        Class (test) & 51/27/20 & 64/73/48 & 21/20/20/20/28 & 127/314/77 & 109/80 \\
        Class (OD) & 37/24/18 & - & - & - & - \\
        \midrule
        \multirow{6.5}{*}{\makecell[c]{Sequence\\generator}} & \multicolumn{5}{l}{\textit{Training}} \\
        \cmidrule{2-6}
        & \multicolumn{5}{c}{Downsampling rate of VAE(8) \hspace{0.2em} $\mathit{T}$(1,000) \hspace{0.2em} AdamW(lr=1e-4) \hspace{0.2em} lr scheduler(Cosine)} \\
        & \multicolumn{5}{c}{Warm(500 iterations) \hspace{0.2em} Epochs(200) \hspace{0.2em} Pretraining batch size(64) \hspace{0.2em} Finetuning batch size(8)} \\
        \cmidrule{2-6}
        & \multicolumn{5}{l}{\textit{Inference}} \\
        \cmidrule{2-6}
        & \multicolumn{5}{c}{Classifier-free guidance~\citep{ho2022classifier} factor(7.5) \hspace{0.4em} Sampling steps(200)} \\
        \midrule
        \multirow{2}{*}{Classifiers} & \multicolumn{5}{c}{Adam(default lr=1e-4, unless specified) \hspace{0.2em} Batch size(default=8, unless specified) \hspace{0.2em} lr scheduler(Cosine)} \\
        & \multicolumn{5}{c}{Epochs(100) \hspace{0.2em} Standard augmentation practices(Random Color/Move/Gaussian/Rotation/Flip)}\\
        \bottomrule
    \end{tabular}}
\end{table*}

\subsection{Datasets and Settings} 
\textbf{Carotid} was collected by three medical centers with approval from local institutional review boards, including a) The Third Affiliated Hospital of Sun Yat-sen University, b) Shenzhen Longgang District People's Hospital, and c) The Eighth Affiliated Hospital of Sun Yat-sen University.
It consists of 273 US video sequences from 231 patients with carotid plaque-induced stenosis.
The severity of carotid stenosis is graded into three stages, i.e., mild, moderate, and severe.
We uniformly sampled several non-overlapping 8-frame clips from each sequence. 
The final dataset involved an in-domain (ID) subset (407 clips from 193 patients collected by Center (a)) and an out-domain (OD) subset (79 clips from 38 patients acquired from Center (b)-(c)).
The ID/OD subset was randomly split into 309/39 and 98/40 clips for training and testing at the patient level.
Each clip has up to two plaques and was labeled with stenosis grading and descriptive text by experienced sonographers using Pair annotation software package~\citep{liang2022sketch}.  
The text annotations indicated the plaque characteristics of echogenicity and location (see Fig.~\ref{fig:framework}).
We cropped the region of interest based on the pretrained carotid vessel detector~\citep{liu2023deep} for easing the model learning.

\textbf{TUSC}~\citep{TUSC} contains 192 thyroid US video sequences from 167 patients collected by the Stanford University Medical Center, with each including one nodule.
Each sequence provided a TI-RADS level and nodule descriptors~\citep{Tessler2017ACR} (e.g., shape), the latter of which was used to form descriptive text.
The TI-RADS level is ranging from 1 to 5.
Considering the imbalanced distribution of TI-RADS levels, as suggested by the experienced sonographers, we rebuilt the dataset by removing the TI-RADS1 video sequence and combining the levels 2-3 sequences into a unified category named TI-RADS2-3.
We uniformly sampled the sequences, obtaining 633 8-frame clips, with 448 for training and 185 for testing.
Each sampled clip inherited the same TI-RADS level and text as the original sequence.
The bounding box of the nodule region for each frame was extracted by radiologists, similar to~\citep{yamashita2022toward}.

\textbf{ACDC}~\citep{bernard2018deep} includes 4D cardiac cine-MRI from 150 patients acquired at the University Hospital of Dijon.
They were evenly divided into five classes: normal patients (NOR), myocardial infarction (MINF), dilated cardiomyopathy (DCM), hypertrophic cardiomyopathy (HCM), and abnormal right ventricle (ARV).
We first extracted volume sequences at end-systole and diastole phases from each case and assigned corresponding classes.
Then, 8-slice clips were uniformly sampled from each sequence along the slice axis.
Finally, 205 and 109 clips were obtained for training and testing.
Text prompts were set to empty strings during generator training.

\textbf{MosMedData}~\citep{morozov2020mosmeddata} contains 3D lung CT sequences from 1,110 patients collected by municipal hospitals in Moscow.
Each sequence was labeled according to the severity of lung tissue abnormalities with COVID-19, originally graded from CT-0 to CT-4 (CT-0: normal, CT-1: mild, CT-2/CT-3/CT-4: increasing severity).
For analysis, CT-2, CT-3, and CT-4 were merged into a single \textit{severe} category to alleviate class imbalance.
We uniformly sampled the sequences to obtain 2,573 8-slice clips, of which 2,055 were used for training and 518 for testing.

\textbf{MRNet}~\citep{bien2018deep} publicly releases 1,250 knee T2-weighted MRI exams, each comprising three orthogonal planes, from 1,199 patients acquired at the Stanford University Medical Center.
Given the complexity of handling multiple imaging planes simultaneously, as suggested in~\citep{khandelwal2018diagnostic}, we only included sagittal-plane sequences in the analysis for diagnosing anterior cruciate ligament (ACL) tears.
Each sequence was then uniformly sampled into several 8-slice clips.
The final dataset was split following the official protocol, resulting in 1,769 training clips and 189 testing clips.

Table~\ref{tab:dataset_setting} presents a detailed description of experimental datasets and settings.
Using all available conditions, we synthesized 300 clips per category for ACDC, 500 clips per class for TUSC and Carotid, 1,500 clips per category for MosMedData, and 2,000 clips per class for MRNet.
All frames/slices were resized to 256$\times$256.
By default, we set \textit{m}=8 for the MFA mechanism.
For the sequence generator, the pretraining LDM was built upon Stable Diffusion~\citep{rombach2022high} and initialized with the public weights\footnote{\url{https://huggingface.co/CompVis/stable-diffusion-v1-4}}.
During sequence LDM finetuning, we updated the parameters of SAM, SA layers, and motion encoder.
We further refined the text- and image prior-sequence alignment by keeping the query projection matrices in cross-attention layers trainable.
During generator training, we employed the multimodal conditions joint training strategy~\citep{huang2023composer}.
This allows users to flexibly select any single condition or combination during sampling, without finetuning the model for each specific combination.
All datasets were employed to benchmark the proposed \textit{\M} against a wide range of augmentation methods and to comprehensively examine its effectiveness across multiple classifiers.
Motivated by~\citep{yang2024freemask}, we adopted three paradigms to explore the impacts of the synthetic samples for downstream classifiers.
\textbf{Baseline}: merely trained with real clips.
\textbf{Real-finetune}: initially pretrained with synthetic clips and then finetuned with real ones.
\textbf{Joint-train}: real clips were first over-sampled to match the number of synthetic clips, the classifier was then trained on the real-synthetic mixed set.
For classification experiments on the MosMedData and MRNet datasets, we set the batch size to 16 and the initial learning rate to 2e-4.
All diagnostic experiments were conducted on MMAction2~\citep{2020mmaction2} except for FTC~\citep{ahmadi2023transformer} and CSG-3DCT~\citep{zhou2023inflated}.

\begin{table*}[!htbp]
    \centering
    \caption{Diagnostic performance comparison of diverse augmentation methods using I3D~\citep{carreira2017quo} on multi-organ and multi-modal datasets. Classifiers with GAN- and diffusion-based augmentations were trained on \textit{Real-finetune} paradigm for diagnosis. 
    \underline{Underlined results} show no significant difference compared with \textit{Ours}, all others indicate significant differences ($p<0.05$).
    95\% confidence intervals for both metrics are reported.
    The best results are shown in bold. Acc., accuracy ($\%$).}
    \label{tab:aug_compare}
    \resizebox{1.0\textwidth}{!}{
    \begin{tabular}{l c c c c c c c c c c}
        \toprule
        \multirow{2.5}{*}{\bf{Augmentation Method}} & \multicolumn{2}{c}{\bf{Carotid}} & \multicolumn{2}{c}{\textbf{TUSC}} & \multicolumn{2}{c}{\textbf{ACDC}} & \multicolumn{2}{c}{\textbf{MosMedData}} & \multicolumn{2}{c}{\textbf{MRNet}} \\
        \cmidrule(lr){2-3} \cmidrule(lr){4-5} \cmidrule(lr){6-7} \cmidrule(lr){8-9} \cmidrule(lr){10-11}
        & Acc. & AUROC & Acc. & AUROC & Acc. & AUROC & Acc. & AUROC & Acc. & AUROC
        \\
        \midrule
        \multicolumn{11}{l}{\textit{\textcolor{gray!70}{Traditional Augmentation}}} \\
        \addlinespace[2pt]
        Basic
        & \makecell{79.59 \\ {[73.47, 85.71]}} & \makecell{0.737 \\ {[0.660, 0.815]}}
        & \makecell{60.72 \\ {[56.04, 65.41]}} & \makecell{0.559 \\ {[0.506, 0.613]}}
        & \makecell{83.49 \\ {[79.82, 87.17]}} & \makecell{0.733 \\ {[0.678, 0.791]}}
        & \makecell{67.18 \\ {[64.35, 70.01]}} & \makecell{0.566 \\ {[0.535, 0.597]}}
        & \makecell{63.49 \\ {[56.08, 70.37]}} & \makecell{0.592 \\ {[0.532, 0.647]}}
        \\
        Random Elastic Deformation
        & \makecell{80.27 \\ {[74.15, 85.71]}} & \makecell{0.747 \\ {[0.672, 0.817]}}
        & \makecell{60.72 \\ {[56.04, 65.41]}} & \makecell{0.554 \\ {[0.502, 0.610]}}
        & \makecell{83.12 \\ {[79.45, 86.79]}} & \makecell{0.725 \\ {[0.671, 0.781]}}
        & \makecell{68.08 \\ {[65.25, 70.91]}} & \makecell{0.584 \\ {[0.553, 0.615]}}
        & \makecell{66.67 \\ {[59.26, 73.54]}} & \makecell{0.623 \\ {[0.563, 0.680]}}
        \\
        Average
        & 79.93 & 0.742
        & 60.72 & 0.557
        & 83.31 & 0.729
        & 67.63 & 0.575
        & 65.08 & 0.608
        \\
        \midrule
        \multicolumn{11}{l}{\textit{\textcolor{gray!70}{Composition-based Augmentation}}} \\
        \addlinespace[2pt]
        \makecell[l]{AutoAugment~\citep{cubuk2019autoaugment}}
        & \makecell{81.63 \\ {[75.62, 86.37]}} & \makecell{0.740 \\ {[0.682, 0.823]}}
        & \makecell{61.08 \\ {[57.84, 66.52]}} & \makecell{0.567 \\ {[0.512, 0.624]}}
        & \makecell{85.68 \\ {[82.26, 89.74]}} & \makecell{0.756 \\ {[0.702, 0.800]}}
        & \makecell{68.72 \\ {[65.84, 71.13]}} & \makecell{0.598 \\ {[0.569, 0.630]}}
        & \makecell{66.13 \\ {[58.25, 72.10]}} & \makecell{0.611 \\ {[0.553, 0.621]}}
        \\
        \makecell[l]{AdaAug~\citep{cheung2021adaaug}}
        & \makecell{82.99 \\ {[76.54, 87.21]}} & \makecell{0.763 \\ {[0.712, 0.841]}}
        & \makecell{60.36 \\ {[56.31, 65.83]}} & \makecell{0.549 \\ {[0.489, 0.603]}}
        & \makecell{84.59 \\ {[81.35, 89.05]}} & \makecell{0.741 \\ {[0.685, 0.792]}}
        & \makecell{69.37 \\ {[66.20, 71.85]}} & \makecell{0.590 \\ {[0.558, 0.624]}} 
        & \makecell{66.67 \\ {[59.26, 72.43]}} & \makecell{0.604 \\ {[0.540, 0.619]}}
        \\
        \makecell[l]{GRA-Video~\citep{an2022group}}
        & \makecell{80.95 \\ {[76.19, 85.71]}} & \makecell{0.722 \\ {[0.663, 0.798]}}
        & \makecell{61.44 \\ {[58.00, 67.34]}} & \makecell{0.556 \\ {[0.508, 0.618]}}
        & \makecell{84.22 \\ {[80.64, 88.12]}} & \makecell{0.747 \\ {[0.689, 0.784]}}
        & \makecell{70.79 \\ {[67.59, 72.20]}} & \makecell{0.584 \\ {[0.551, 0.619]}}
        & \makecell{66.13 \\ {[58.87, 72.58]}} & \makecell{0.617 \\ {[0.569, 0.630]}}
        \\
        \makecell[l]{Sample-aware RandAugment~\citep{xiao2025sample}}
        & \makecell{82.31 \\ {[76.19, 87.76]}} & \makecell{0.758 \\ {[0.700, 0.834]}}
        & \makecell{61.80 \\ {[58.85, 67.76]}} & \makecell{0.587 \\ {[0.529, 0.631]}}
        & \makecell{83.85 \\ {[79.98, 87.25]}} & \makecell{0.739 \\ {[0.689, 0.790]}}
        & \makecell{70.27 \\ {[67.18, 71.94]}} & \makecell{0.593 \\ {[0.562, 0.627]}}
        & \makecell{66.67 \\ {[59.26, 72.43]}} & \makecell{0.625 \\ {[0.559, 0.674]}}
        \\
        Average
        & 81.97 & 0.746
        & 61.17 & 0.565
        & 84.59 & 0.746
        & 69.79 & 0.591
        & 66.40 & 0.614
        \\
        \midrule
        \multicolumn{11}{l}{\textit{\textcolor{gray!70}{Mixing-based Augmentation}}} \\
        \addlinespace[2pt]
        \makecell[l]{Mixup~\citep{zhang2017mixup}}
        & \makecell{83.67 \\ {[77.51, 88.20]}} & \makecell{0.772 \\ {[0.715, 0.839]}}
        & \makecell{62.52 \\ {[57.84, 67.57]}} & \makecell{0.591 \\ {[0.534, 0.648]}}
        & \makecell{84.95 \\ {[81.10, 89.23]}} & \makecell{0.759 \\ {[0.707, 0.805]}}  
        & \makecell{71.17 \\ {[68.05, 72.83]}} & \makecell{0.589 \\ {[0.551, 0.614]}}
        & \makecell{67.72 \\ {[59.99, 74.06]}} & \makecell{0.631 \\ {[0.575, 0.685]}}
        \\
        \makecell[l]{CutMix~\citep{yun2019cutmix}}
        & \makecell{82.31 \\ {[76.19, 87.76]}} & \makecell{0.765 \\ {[0.695, 0.833]}}
        & \makecell{62.52 \\ {[57.84, 67.57]}} & \makecell{0.581 \\ {[0.526, 0.638]}}
        & \makecell{83.12 \\ {[79.45, 86.79]}} & \makecell{0.726 \\ {[0.676, 0.778]}}
        & \makecell{69.24 \\ {[66.41, 72.20]}} & \makecell{0.555 \\ {[0.524, 0.587]}} 
        & \makecell{68.25 \\ {[61.38, 74.60]}} & \makecell{0.635 \\ {[0.579, 0.688]}}
        \\
        \makecell[l]{VideoMix~\citep{yun2020videomix}}
        & \makecell{83.67 \\ {[77.51, 88.20]}} & \makecell{0.779 \\ {[0.723, 0.860]}}
        & \makecell{63.24 \\ {[60.00, 68.00]}} & \makecell{0.571 \\ {[0.514, 0.626]}}
        & \makecell{85.32 \\ {[82.00, 89.51]}} & \makecell{0.766 \\ {[0.718, 0.820]}}
        & \makecell{71.04 \\ {[68.00, 73.58]}} & \makecell{0.573 \\ {[0.536, 0.607]}}
        & \makecell{69.31 \\ {[63.02, 75.73]}} & \makecell{0.657 \\ {[0.595, 0.719]}}
        \\
        \makecell[l]{LayerMix~\citep{ahmad2025layermix}}
        & \makecell{84.35 \\ {[78.23, 89.80]}} & \makecell{0.792 \\ {[0.720, 0.858]}}
        & \makecell{61.44 \\ {[56.76, 66.13]}} & \makecell{0.570 \\ {[0.514, 0.621]}}
        & \makecell{83.85 \\ {[80.18, 87.52]}} & \makecell{0.740 \\ {[0.686, 0.796]}}
        & \makecell{71.17 \\ {[68.21, 74.00]}} & \makecell{0.567 \\ {[0.536, 0.600]}}
        & \makecell{\underline{69.84} \\ {[63.48, 76.19]}} & \makecell{\underline{0.662} \\ {[0.601, 0.721]}}
        \\
        Average
        & 83.50 & 0.777
        & 62.43 & 0.578
        & 84.31 & 0.748
        & 70.66 & 0.571
        & 68.78 & 0.646
        \\
        \midrule
        \multicolumn{11}{l}{\textit{\textcolor{gray!70}{GAN-based Augmentation}}} \\
        \addlinespace[2pt]
        \makecell[l]{TGANv2~\citep{saito2020train}}
        & \makecell{80.95 \\ {[74.83, 87.07]}} & \makecell{0.728 \\ {[0.675, 0.803]}}
        & \makecell{60.72 \\ {[56.28, 64.90]}} & \makecell{0.564 \\ {[0.508, 0.621]}}
        & \makecell{85.68 \\ {[82.26, 89.74]}} & \makecell{0.760 \\ {[0.703, 0.805]}}
        & \makecell{71.30 \\ {[68.84, 73.12]}} & \makecell{0.587 \\ {[0.553, 0.619]}}
        & \makecell{68.25 \\ {[61.29, 74.65]}} & \makecell{0.626 \\ {[0.566, 0.681]}}
        \\
        \makecell[l]{StyleGAN-V~\citep{skorokhodov2022stylegan}}
        & \makecell{82.31 \\ {[76.19, 87.76]}} & \makecell{0.749 \\ {[0.694, 0.822]}}
        & \makecell{61.80 \\ {[58.85, 67.34]}} & \makecell{0.593 \\ {[0.536, 0.645]}}
        & \makecell{85.32 \\ {[81.35, 89.74]}} & \makecell{0.768 \\ {[0.721, 0.823]}}
        & \makecell{71.43 \\ {[69.22, 73.65]}} & \makecell{0.592 \\ {[0.555, 0.612]}}
        & \makecell{68.25 \\ {[61.19, 74.62]}} & \makecell{0.633 \\ {[0.578, 0.688]}}
        \\
        Average
        & 81.63 & 0.739
        & 61.26 & 0.579
        & 85.50 & 0.764
        & 71.37 & 0.590
        & 68.25 & 0.630
        \\
        \midrule
        \multicolumn{11}{l}{\textit{\textcolor{gray!70}{Diffusion-based Augmentation}}} \\
        \addlinespace[2pt]
        \makecell[l]{VideoComposer~\citep{wang2023videocomposer}}
        & \makecell{80.95 \\ {[74.83, 87.07]}} & \makecell{0.764 \\ {[0.697, 0.833]}}
        & \makecell{63.60 \\ {[58.92, 68.29]}} & \makecell{0.590 \\ {[0.538, 0.645]}}
        & \makecell{85.32 \\ {[81.65, 88.99]}} & \makecell{0.761 \\ {[0.710, 0.815]}}
        & \makecell{71.56 \\ {[68.73, 74.39]}} & \makecell{0.591 \\ {[0.559, 0.626]}}
        & \makecell{68.78 \\ {[61.90, 75.13]}} & \makecell{0.653 \\ {[0.588, 0.712]}}
        \\
        \makecell[l]{Endora~\citep{li2024endora}}
        & \makecell{82.31 \\ {[76.19, 87.76]}} & \makecell{0.772 \\ {[0.700, 0.842]}}
        & \makecell{\underline{64.68} \\ {[60.00, 69.73]}} & \makecell{\underline{0.601} \\ {[0.546, 0.658]}}
        & \makecell{84.22 \\ {[80.55, 87.53]}} & \makecell{0.747 \\ {[0.693, 0.803]}}
        & \makecell{69.24 \\ {[66.41, 72.08]}} & \makecell{0.581 \\ {[0.547, 0.617]}}
        & \makecell{69.31 \\ {[62.43, 75.66]}} & \makecell{0.657 \\ {[0.596, 0.717]}}
        \\
        \makecell[l]{3D MedDiff~\citep{wang20253d}}
        & \makecell{82.99 \\ {[77.51, 87.21]}} & \makecell{0.786 \\ {[0.739, 0.867]}}
        & \makecell{61.80 \\ {[57.94, 66.55]}} & \makecell{0.573 \\ {[0.510, 0.632]}}
        & \makecell{84.58 \\ {[81.15, 89.34]}} & \makecell{0.755 \\ {[0.700, 0.796]}}
        & \makecell{\underline{71.68} \\ {[69.55, 74.89]}} & \makecell{\underline{0.601} \\ {[0.568, 0.638]}}
        & \makecell{67.72 \\ {[60.15, 74.00]}} & \makecell{0.638 \\ {[0.581, 0.694]}}
        \\
        Average
        & 82.08 & 0.774
        & 63.36 & 0.588
        & 84.71 & 0.754
        & 70.83 & 0.591
        & 68.60 & 0.649
        \\
        \midrule
        Ours
        & \makecell{\bf{85.03} \\ {[79.59, 90.48]}} & \makecell{\bf{0.813} \\ {[0.744, 0.880]}}
        & \makecell{\bf{65.41} \\ {[60.36, 70.09]}} & \makecell{\bf{0.611} \\ {[0.556, 0.664]}}
        & \makecell{\bf{87.16} \\ {[83.49, 90.46]}} & \makecell{\bf{0.795} \\ {[0.742, 0.843]}}
        & \makecell{\bf{72.59} \\ {[69.88, 75.42]}} & \makecell{\bf{0.609} \\ {[0.575, 0.646]}}
        & \makecell{\bf{71.43} \\ {[65.08, 77.78]}} & \makecell{\bf{0.674} \\ {[0.616, 0.733]}}
        \\
        \bottomrule
    \end{tabular}}
\end{table*}

\begin{table*}[!t]
    \centering
    \caption{Diagnostic performance comparison using eleven classifiers trained on \textit{Baseline} (A), \textit{Real-finetune} (B), and \textit{Joint-train} (C) paradigms in Carotid, TUSC~\citep{TUSC}, and ACDC~\citep{bernard2018deep} datasets.
    \underline{Underlined results} show no significant difference compared with \textit{Baseline} (A), all others indicate significant differences ($p<0.05$).
    95\% confidence intervals for both metrics are reported.
    “Hybrid”, the classifier with CNN-Transformer design. Acc., accuracy ($\%$).}
    \label{tab:results_full_shot_threesets}
    \resizebox{1.0\textwidth}{!}{
    \begin{tabular}{c c c c c c c c c c c c c c c}
        \toprule
        \multirow{2.5}{*}{\bf{Classifier}} & \multirow{2.5}{*}{\bf{Backbone}} & \multirow{2.5}{*}{\bf{\makecell[c]{Training\\Paradigm}}} & \multicolumn{4}{c}{\bf{Carotid}} & \multicolumn{4}{c}{\textbf{TUSC}} & \multicolumn{4}{c}{\textbf{ACDC}} \\
        \cmidrule(lr){4-7} \cmidrule(lr){8-11} \cmidrule(lr){12-15}
        & & & Acc. & \textcolor{delta}{$\Delta$} & AUROC & \textcolor{delta}{$\Delta$} & Acc. & \textcolor{delta}{$\Delta$} & AUROC & \textcolor{delta}{$\Delta$} & Acc. & \textcolor{delta}{$\Delta$} & AUROC & \textcolor{delta}{$\Delta$} \\
        \midrule
        \multirow{5.1}{*}{\makecell[c]{I3D\\~\citep{carreira2017quo}}} & \multirow{5.1}{*}{CNN} 
            & A
            & \makecell{79.59 \\ {[73.47, 85.71]}} & & \makecell{0.737 \\ {[0.660, 0.815]}} & 
            & \makecell{60.72 \\ {[56.04, 65.41]}} & & \makecell{0.559 \\ {[0.506, 0.613]}} & 
            & \makecell{83.49 \\ {[79.82, 87.17]}} & & \makecell{0.733 \\ {[0.678, 0.791]}} & 
            \\
            & & B
            & \vspace{2pt}\cellcolor{gray!15}\makecell{85.03 \\ {[79.59, 90.48]}} & \textcolor{delta}{$\uparrow$ 5.44} & \cellcolor{gray!15}\makecell{0.813 \\ {[0.744, 0.880]}} & \textcolor{delta}{$\uparrow$ 0.076} 
            & \cellcolor{gray!15}\makecell{65.41 \\ {[60.36, 70.09]}} & \textcolor{delta}{$\uparrow$ 4.69} & \cellcolor{gray!15}\makecell{0.611 \\ {[0.556, 0.664]}} & \textcolor{delta}{$\uparrow$ 0.052}
            & \cellcolor{gray!15}\makecell{87.16 \\ {[83.49, 90.46]}} & \textcolor{delta}{$\uparrow$ 3.67} & \cellcolor{gray!15}\makecell{0.795 \\ {[0.742, 0.843]}} & \textcolor{delta}{$\uparrow$ 0.062} 
            \\
            & & C 
            & \cellcolor{gray!15}\makecell{84.35 \\ {[78.91, 89.80]}} & \textcolor{delta}{$\uparrow$ 4.76} & \cellcolor{gray!15}\makecell{0.811 \\ {[0.741, 0.874]}} & \textcolor{delta}{$\uparrow$ 0.074} 
            & \cellcolor{gray!15}\makecell{62.88 \\ {[58.20, 67.57]}} & \textcolor{delta}{$\uparrow$ 2.16} & \cellcolor{gray!15}\makecell{0.578 \\ {[0.525, 0.634]}} & \textcolor{delta}{$\uparrow$ 0.019}
            & \cellcolor{gray!15}\makecell{85.32 \\ {[81.65, 88.99]}} & \textcolor{delta}{$\uparrow$ 1.83} & \cellcolor{gray!15}\makecell{0.762 \\ {[0.706, 0.818]}} & \textcolor{delta}{$\uparrow$ 0.029}
            \\
        \midrule
        \multirow{5.1}{*}{\makecell[c]{R(2+1)D\\~\citep{tran2018closer}}} & \multirow{5.1}{*}{CNN} 
            & A
            & \makecell{82.99 \\ {[76.87, 89.12]}} & & \makecell{0.781 \\ {[0.707, 0.854]}} & 
            & \makecell{63.96 \\ {[59.27, 68.65]}} & & \makecell{0.595 \\ {[0.541, 0.649]}} & 
            & \makecell{83.85 \\ {[80.18, 87.52]}} & & \makecell{0.737 \\ {[0.681, 0.792]}} & 
            \\
            & & B
            & \vspace{2pt}\cellcolor{gray!15}\makecell{83.67 \\ {[77.55, 89.12]}} & \textcolor{delta}{$\uparrow$ 0.68} & \cellcolor{gray!15}\makecell{0.797 \\ {[0.728, 0.866]}} & \textcolor{delta}{$\uparrow$ 0.016} 
            & \cellcolor{gray!15}\makecell{64.68 \\ {[60.00, 69.73]}} & \textcolor{delta}{$\uparrow$ 0.72} & \cellcolor{gray!15}\makecell{0.601 \\ {[0.548, 0.659]}} & \textcolor{delta}{$\uparrow$ 0.006}  
            & \cellcolor{gray!15}\makecell{85.69 \\ {[82.39, 88.99]}} & \textcolor{delta}{$\uparrow$ 1.84} & \cellcolor{gray!15}\makecell{0.769 \\ {[0.714, 0.823]}} & \textcolor{delta}{$\uparrow$ 0.032} 
            \\
            & & C 
            & \cellcolor{gray!15}\makecell{85.04 \\ {[78.91, 90.48]}} & \textcolor{delta}{$\uparrow$ 2.05} & \cellcolor{gray!15}\makecell{0.804 \\ {[0.737, 0.868]}} & \textcolor{delta}{$\uparrow$ 0.023} 
            & \cellcolor{gray!15}\makecell{65.41 \\ {[60.72, 70.45]}} & \textcolor{delta}{$\uparrow$ 1.45} & \cellcolor{gray!15}\makecell{0.610 \\ {[0.554, 0.665]}} & \textcolor{delta}{$\uparrow$ 0.015} 
            & \cellcolor{gray!15}\makecell{85.69 \\ {[82.02, 89.36]}} & \textcolor{delta}{$\uparrow$ 1.84} & \cellcolor{gray!15}\makecell{0.771 \\ {[0.723, 0.821]}} & \textcolor{delta}{$\uparrow$ 0.034} 
            \\
        \midrule
        \multirow{5.1}{*}{\makecell[c]{SlowFast\\~\citep{feichtenhofer2019slowfast}}} & \multirow{5.1}{*}{CNN} 
            & A
            & \makecell{80.95 \\ {[74.83, 86.41]}} & & \makecell{0.742 \\ {[0.669, 0.814]}} & 
            & \makecell{62.88 \\ {[57.84, 67.93]}} & & \makecell{0.579 \\ {[0.524, 0.637]}} & 
            & \makecell{81.65 \\ {[77.61, 85.32]}} & & \makecell{0.702 \\ {[0.650, 0.753]}} & 
            \\
            & & B
            & \vspace{2pt}\cellcolor{gray!15}\makecell{\underline{81.63} \\ {[75.51, 87.07]}} & \textcolor{delta}{$\uparrow$ 0.68} & \cellcolor{gray!15}\makecell{\underline{0.770} \\ {[0.697, 0.838]}} & \textcolor{delta}{$\uparrow$ 0.028} 
            & \cellcolor{gray!15}\makecell{65.76 \\ {[61.08, 70.45]}} & \textcolor{delta}{$\uparrow$ 2.88} & \cellcolor{gray!15}\makecell{0.609 \\ {[0.555, 0.664]}} & \textcolor{delta}{$\uparrow$ 0.030}  
            & \cellcolor{gray!15}\makecell{86.05 \\ {[82.39, 89.36]}} & \textcolor{delta}{$\uparrow$ 4.40} & \cellcolor{gray!15}\makecell{0.772 \\ {[0.719, 0.823]}} & \textcolor{delta}{$\uparrow$ 0.070}
            \\
            & & C  
            & \cellcolor{gray!15}\makecell{84.35 \\ {[78.91, 89.80]}} & \textcolor{delta}{$\uparrow$ 3.40} & \cellcolor{gray!15}\makecell{0.811 \\ {[0.740, 0.876]}} & \textcolor{delta}{$\uparrow$ 0.069} 
            & \cellcolor{gray!15}\makecell{65.41 \\ {[60.36, 70.09]}} & \textcolor{delta}{$\uparrow$ 2.53} & \cellcolor{gray!15}\makecell{0.609 \\ {[0.552, 0.662]}} & \textcolor{delta}{$\uparrow$ 0.030} 
            & \cellcolor{gray!15}\makecell{83.85 \\ {[80.54, 87.52]}} & \textcolor{delta}{$\uparrow$ 2.20} & \cellcolor{gray!15}\makecell{0.742 \\ {[0.689, 0.794]}} & \textcolor{delta}{$\uparrow$ 0.040} 
            \\
        \midrule
        \multirow{5.1}{*}{\makecell[c]{CSN\\~\citep{tran2019video}}} & \multirow{5.1}{*}{CNN} 
            & A
            & \makecell{80.27 \\ {[74.15, 86.39]}} & & \makecell{0.757 \\ {[0.683, 0.828]}} & 
            & \makecell{61.80 \\ {[57.12, 66.85]}} & & \makecell{0.570 \\ {[0.514, 0.627]}} & 
            & \makecell{85.69 \\ {[82.02, 89.36]}} & & \makecell{0.765 \\ {[0.710, 0.821]}} & 
            \\
            & & B
            & \vspace{2pt}\cellcolor{gray!15}\makecell{84.35 \\ {[78.91, 89.80]}} & \textcolor{delta}{$\uparrow$ 4.08} & \cellcolor{gray!15}\makecell{0.784 \\ {[0.710, 0.853]}} & \textcolor{delta}{$\uparrow$ 0.027} 
            & \cellcolor{gray!15}\makecell{67.21 \\ {[62.16, 72.25]}} & \textcolor{delta}{$\uparrow$ 5.41} & \cellcolor{gray!15}\makecell{0.631 \\ {[0.575, 0.688]}} & \textcolor{delta}{$\uparrow$ 0.061}  
            & \cellcolor{gray!15}\makecell{87.15 \\ {[83.49, 90.83]}} & \textcolor{delta}{$\uparrow$ 1.46} & \cellcolor{gray!15}\makecell{0.794 \\ {[0.743, 0.845]}} & \textcolor{delta}{$\uparrow$ 0.029} 
            \\
            & & C 
            & \cellcolor{gray!15}\makecell{87.07 \\ {[81.63, 91.84]}} & \textcolor{delta}{$\uparrow$ 6.80} & \cellcolor{gray!15}\makecell{0.843 \\ {[0.774, 0.905]}} & \textcolor{delta}{$\uparrow$ 0.086} 
            & \cellcolor{gray!15}\makecell{65.41 \\ {[60.72, 70.10]}} & \textcolor{delta}{$\uparrow$ 3.61} & \cellcolor{gray!15}\makecell{0.610 \\ {[0.554, 0.665]}} & \textcolor{delta}{$\uparrow$ 0.040} 
            & \cellcolor{gray!15}\makecell{88.62 \\ {[84.95, 91.93]}} & \textcolor{delta}{$\uparrow$ 2.93} & \cellcolor{gray!15}\makecell{0.814 \\ {[0.761, 0.865]}} & \textcolor{delta}{$\uparrow$ 0.049} 
            \\
        \midrule
        \multirow{5.1}{*}{\makecell[c]{TPN\\~\citep{yang2020temporal}}} & \multirow{5.1}{*}{CNN} 
            & A
            & \makecell{82.99 \\ {[76.87, 88.44]}} & & \makecell{0.784 \\ {[0.711, 0.853]}} & 
            & \makecell{64.32 \\ {[59.64, 69.37]}} & & \makecell{0.602 \\ {[0.546, 0.657]}} & 
            & \makecell{83.85 \\ {[80.18, 87.52]}} & & \makecell{0.735 \\ {[0.681, 0.787]}} & 
            \\
            & & B
            & \vspace{2pt}\cellcolor{gray!15}\makecell{87.76 \\ {[82.99, 92.52]}} & \textcolor{delta}{$\uparrow$ 4.77} & \cellcolor{gray!15}\makecell{0.856 \\ {[0.795, 0.912]}} & \textcolor{delta}{$\uparrow$ 0.072} 
            & \cellcolor{gray!15}\makecell{66.85 \\ {[62.16, 71.89]}} & \textcolor{delta}{$\uparrow$ 2.53} & \cellcolor{gray!15}\makecell{0.633 \\ {[0.580, 0.688]}} & \textcolor{delta}{$\uparrow$ 0.031}  
            & \cellcolor{gray!15}\makecell{85.32 \\ {[81.65, 88.62]}} & \textcolor{delta}{$\uparrow$ 1.47} & \cellcolor{gray!15}\makecell{0.764 \\ {[0.708, 0.816]}} & \textcolor{delta}{$\uparrow$ 0.029} 
            \\
            & & C 
            & \cellcolor{gray!15}\makecell{85.71 \\ {[79.59, 90.48]}} & \textcolor{delta}{$\uparrow$ 2.72} & \cellcolor{gray!15}\makecell{0.823 \\ {[0.751, 0.887]}} & \textcolor{delta}{$\uparrow$ 0.039} 
            & \cellcolor{gray!15}\makecell{65.41 \\ {[60.72, 70.09]}} & \textcolor{delta}{$\uparrow$ 1.09} & \cellcolor{gray!15}\makecell{0.609 \\ {[0.556, 0.663]}} & \textcolor{delta}{$\uparrow$ 0.007} 
            & \cellcolor{gray!15}\makecell{84.95 \\ {[81.28, 88.62]}} & \textcolor{delta}{$\uparrow$ 1.10} & \cellcolor{gray!15}\makecell{0.756 \\ {[0.702, 0.812]}} & \textcolor{delta}{$\uparrow$ 0.021} 
            \\
        \midrule
        \multirow{5.1}{*}{\makecell[c]{TimeSformer\\~\citep{bertasius2021space}}} & \multirow{5.1}{*}{Transformer} 
            & A
            & \makecell{77.55 \\ {[70.75, 83.67]}} & & \makecell{0.721 \\ {[0.646, 0.794]}} & 
            & \makecell{65.41 \\ {[60.36, 70.45]}} & & \makecell{0.614 \\ {[0.558, 0.674]}} & 
            & \makecell{72.48 \\ {[69.17, 76.15]}} & & \makecell{0.562 \\ {[0.513, 0.615]}} & 
            \\
            & & B
            & \vspace{2pt}\cellcolor{gray!15}\makecell{80.95 \\ {[74.83, 86.39]}} & \textcolor{delta}{$\uparrow$ 3.40} & \cellcolor{gray!15}\makecell{0.753 \\ {[0.686, 0.819]}} & \textcolor{delta}{$\uparrow$ 0.032} 
            & \cellcolor{gray!15}\makecell{66.49 \\ {[61.80, 71.17]}} & \textcolor{delta}{$\uparrow$ 1.08} & \cellcolor{gray!15}\makecell{0.626 \\ {[0.574, 0.678]}} & \textcolor{delta}{$\uparrow$ 0.012}  
            & \cellcolor{gray!15}\makecell{76.51 \\ {[72.84, 80.18]}} & \textcolor{delta}{$\uparrow$ 4.03} & \cellcolor{gray!15}\makecell{0.626 \\ {[0.575, 0.680]}} & \textcolor{delta}{$\uparrow$ 0.064} 
            \\
            & & C 
            & \cellcolor{gray!15}\makecell{79.59 \\ {[72.79, 85.71]}} & \textcolor{delta}{$\uparrow$ 2.04} & \cellcolor{gray!15}\makecell{0.730 \\ {[0.650, 0.805]}} & \textcolor{delta}{$\uparrow$ 0.009} 
            & \cellcolor{gray!15}\makecell{66.49 \\ {[61.80, 71.53]}} & \textcolor{delta}{$\uparrow$ 1.08} & \cellcolor{gray!15}\makecell{0.618 \\ {[0.564, 0.675]}} & \textcolor{delta}{$\uparrow$ 0.004} 
            & \cellcolor{gray!15}\makecell{77.62 \\ {[73.94, 81.28]}} & \textcolor{delta}{$\uparrow$ 5.14} & \cellcolor{gray!15}\makecell{0.642 \\ {[0.589, 0.698]}} & \textcolor{delta}{$\uparrow$ 0.080} 
            \\
        \midrule
        \multirow{5.1}{*}{\makecell[c]{MViTv2\\~\citep{li2022mvitv2}}} & \multirow{5.1}{*}{Transformer} 
            & A
            & \makecell{74.15 \\ {[68.03, 80.27]}} & & \makecell{0.663 \\ {[0.593, 0.735]}} & 
            & \makecell{59.64 \\ {[54.95, 64.32]}} & & \makecell{0.543 \\ {[0.490, 0.597]}} & 
            & \makecell{77.98 \\ {[74.31, 81.65]}} & & \makecell{0.642 \\ {[0.590, 0.695]}} & 
            \\
            & & B
            & \vspace{2pt}\cellcolor{gray!15}\makecell{78.91 \\ {[72.79, 85.03]}} & \textcolor{delta}{$\uparrow$ 4.76} & \cellcolor{gray!15}\makecell{0.740 \\ {[0.665, 0.810]}} & \textcolor{delta}{$\uparrow$ 0.077} 
            & \cellcolor{gray!15}\makecell{63.96 \\ {[59.28, 68.65]}} & \textcolor{delta}{$\uparrow$ 4.32} & \cellcolor{gray!15}\makecell{0.596 \\ {[0.540, 0.647]}} & \textcolor{delta}{$\uparrow$ 0.053}  
            & \cellcolor{gray!15}\makecell{\underline{78.72} \\ {[75.05, 82.39]}} & \textcolor{delta}{$\uparrow$ 0.74} & \cellcolor{gray!15}\makecell{\underline{0.653} \\ {[0.601, 0.710]}} & \textcolor{delta}{$\uparrow$ 0.011} 
            \\
            & & C 
            & \cellcolor{gray!15}\makecell{76.87 \\ {[70.75, 82.99]}} & \textcolor{delta}{$\uparrow$ 2.72} & \cellcolor{gray!15}\makecell{0.701 \\ {[0.625, 0.772]}} & \textcolor{delta}{$\uparrow$ 0.038} 
            & \cellcolor{gray!15}\makecell{62.17 \\ {[57.12, 67.21]}} & \textcolor{delta}{$\uparrow$ 2.53} & \cellcolor{gray!15}\makecell{0.568 \\ {[0.512, 0.624]}} & \textcolor{delta}{$\uparrow$ 0.025} 
            & \cellcolor{gray!15}\makecell{80.18 \\ {[76.51, 84.22]}} & \textcolor{delta}{$\uparrow$ 2.20} & \cellcolor{gray!15}\makecell{0.680 \\ {[0.626, 0.738]}} & \textcolor{delta}{$\uparrow$ 0.038} 
            \\
        \midrule
        \multirow{5.1}{*}{\makecell[c]{VideoSwin\\~\citep{liu2022video}}} & \multirow{5.1}{*}{Transformer} 
            & A
            & \makecell{79.59 \\ {[73.47, 85.71]}} & & \makecell{0.732 \\ {[0.658, 0.807]}} & 
            & \makecell{58.92 \\ {[54.23, 63.61]}} & & \makecell{0.535 \\ {[0.480, 0.591]}} & 
            & \makecell{74.31 \\ {[70.64, 77.98]}} & & \makecell{0.572 \\ {[0.543, 0.604]}} & 
            \\
            & & B
            & \vspace{2pt}\cellcolor{gray!15}\makecell{83.67 \\ {[78.23, 89.12]}} & \textcolor{delta}{$\uparrow$ 4.08} & \cellcolor{gray!15}\makecell{0.789 \\ {[0.722, 0.858]}} & \textcolor{delta}{$\uparrow$ 0.057} 
            & \cellcolor{gray!15}\makecell{61.08 \\ {[56.40, 65.77]}} & \textcolor{delta}{$\uparrow$ 2.16} & \cellcolor{gray!15}\makecell{0.551 \\ {[0.497, 0.605]}} & \textcolor{delta}{$\uparrow$ 0.016}  
            & \cellcolor{gray!15}\makecell{79.45 \\ {[75.78, 83.12]}} & \textcolor{delta}{$\uparrow$ 5.14} & \cellcolor{gray!15}\makecell{0.663 \\ {[0.612, 0.718]}} & \textcolor{delta}{$\uparrow$ 0.091} 
            \\
            & & C 
            & \cellcolor{gray!15}\makecell{83.67 \\ {[77.55, 89.12]}} & \textcolor{delta}{$\uparrow$ 4.08} & \cellcolor{gray!15}\makecell{0.776 \\ {[0.701, 0.846]}} & \textcolor{delta}{$\uparrow$ 0.044} 
            & \cellcolor{gray!15}\makecell{61.44 \\ {[56.76, 66.49]}} & \textcolor{delta}{$\uparrow$ 2.52} & \cellcolor{gray!15}\makecell{0.562 \\ {[0.508, 0.618]}} & \textcolor{delta}{$\uparrow$ 0.027} 
            & \cellcolor{gray!15}\makecell{77.62 \\ {[73.94, 81.28]}} & \textcolor{delta}{$\uparrow$ 3.31} & \cellcolor{gray!15}\makecell{0.632 \\ {[0.588, 0.674]}} & \textcolor{delta}{$\uparrow$ 0.060} 
            \\
        \midrule
        \multirow{5.1}{*}{\makecell[c]{UniFormerV2\\~\citep{li2022uniformerv2}}} & \multirow{5.1}{*}{Hybrid} 
            & A
            & \makecell{76.87 \\ {[70.75, 82.31]}} & & \makecell{0.710 \\ {[0.638, 0.782]}} & 
            & \makecell{65.41 \\ {[60.71, 70.09]}} & & \makecell{0.611 \\ {[0.554, 0.664]}} & 
            & \makecell{75.78 \\ {[72.11, 79.45]}} & & \makecell{0.614 \\ {[0.561, 0.665]}} & 
            \\
            & & B
            & \vspace{2pt}\cellcolor{gray!15}\makecell{80.27 \\ {[74.15, 85.73]}} & \textcolor{delta}{$\uparrow$ 3.40} & \cellcolor{gray!15}\makecell{0.743 \\ {[0.672, 0.815]}} & \textcolor{delta}{$\uparrow$ 0.033} 
            & \cellcolor{gray!15}\makecell{\underline{65.77} \\ {[61.07, 70.81]}} & \textcolor{delta}{$\uparrow$ 0.36} & \cellcolor{gray!15}\makecell{\underline{0.618} \\ {[0.562, 0.674]}} & \textcolor{delta}{$\uparrow$ 0.007}  
            & \cellcolor{gray!15}\makecell{77.25 \\ {[73.58, 80.92]}} & \textcolor{delta}{$\uparrow$ 1.47} & \cellcolor{gray!15}\makecell{0.645 \\ {[0.592, 0.699]}} & \textcolor{delta}{$\uparrow$ 0.031} 
            \\
            & & C  
            & \cellcolor{gray!15}\makecell{81.63 \\ {[76.17, 87.07]}} & \textcolor{delta}{$\uparrow$ 4.76} & \cellcolor{gray!15}\makecell{0.760 \\ {[0.689, 0.830]}} & \textcolor{delta}{$\uparrow$ 0.050} 
            & \cellcolor{gray!15}\makecell{\underline{66.13} \\ {[61.43, 70.81]}} & \textcolor{delta}{$\uparrow$ 0.72} & \cellcolor{gray!15}\makecell{\underline{0.617} \\ {[0.562, 0.671]}} & \textcolor{delta}{$\uparrow$ 0.006} 
            & \cellcolor{gray!15}\makecell{77.61 \\ {[73.94, 81.28]}} & \textcolor{delta}{$\uparrow$ 1.83} & \cellcolor{gray!15}\makecell{0.637 \\ {[0.591, 0.686]}} & \textcolor{delta}{$\uparrow$ 0.023} 
            \\
        \midrule
        \multirow{5.1}{*}{\makecell[c]{FTC\\~\citep{ahmadi2023transformer}}} & \multirow{5.1}{*}{Hybrid} 
            & A
            & \makecell{76.99 \\ {[70.86, 80.12]}} & & \makecell{0.723 \\ {[0.669, 0.776]}} & 
            & \makecell{62.09 \\ {[57.71, 64.98]}} & & \makecell{0.584 \\ {[0.523, 0.625]}} & 
            & \makecell{80.95 \\ {[77.93, 83.20]}} & & \makecell{0.689 \\ {[0.618, 0.713]}} & 
            \\
            & & B
            & \vspace{2pt}\cellcolor{gray!15}\makecell{80.91 \\ {[76.44, 82.97]}} & \textcolor{delta}{$\uparrow$ 3.92} & \cellcolor{gray!15}\makecell{0.750 \\ 
            {[0.685, 0.782]}} & \textcolor{delta}{$\uparrow$ 0.027} 
            & \cellcolor{gray!15}\makecell{64.09 \\ 
            {[58.56, 67.31]}} & \textcolor{delta}{$\uparrow$ 2.00} & \cellcolor{gray!15}\makecell{0.593 \\ 
            {[0.541, 0.630]}} & \textcolor{delta}{$\uparrow$ 0.009}  
            & \cellcolor{gray!15}\makecell{82.38 \\ 
            {[79.30, 85.24]}} & \textcolor{delta}{$\uparrow$ 1.43} & \cellcolor{gray!15}\makecell{0.703 \\ 
            {[0.638, 0.732]}} & \textcolor{delta}{$\uparrow$ 0.014} 
            \\
            & & C  
            & \cellcolor{gray!15}\makecell{81.80 \\ 
            {[77.23, 83.56]}} & \textcolor{delta}{$\uparrow$ 4.81} & \cellcolor{gray!15}\makecell{0.758 \\ 
            {[0.702, 0.798]}} & \textcolor{delta}{$\uparrow$ 0.035} 
            & \cellcolor{gray!15}\makecell{65.66 \\ {[60.28, 68.53]}} & \textcolor{delta}{$\uparrow$ 3.57} & \cellcolor{gray!15}\makecell{0.614 \\ 
            {[0.567, 0.659]}} & \textcolor{delta}{$\uparrow$ 0.030} 
            & \cellcolor{gray!15}\makecell{84.38 \\ {[80.21, 86.10]}} & \textcolor{delta}{$\uparrow$ 3.43} & \cellcolor{gray!15}\makecell{0.718 \\ 
            {[0.651, 0.743]}} & \textcolor{delta}{$\uparrow$ 0.029} 
            \\
        \midrule
        \multirow{5.1}{*}{\makecell[c]{CSG-3DCT\\~\citep{zhou2023inflated}}} & \multirow{5.1}{*}{Hybrid} 
            & A
            & \makecell{84.35 \\ {[78.23, 89.80]}} & & \makecell{0.812 \\ {[0.740, 0.879]}} & 
            & \makecell{61.44 \\ {[56.76, 66.49]}} & & \makecell{0.571 \\ {[0.515, 0.626]}} & 
            & \makecell{87.16 \\ {[83.49, 90.46]}} & & \makecell{0.797 \\ {[0.747, 0.846]}} & 
            \\
            & & B
            & \vspace{2pt}\cellcolor{gray!15}\makecell{87.07 \\ {[81.63, 91.84]}} & \textcolor{delta}{$\uparrow$ 2.72} & \cellcolor{gray!15}\makecell{0.831 \\ {[0.762, 0.897]}} & \textcolor{delta}{$\uparrow$ 0.019} 
            & \cellcolor{gray!15}\makecell{63.96 \\ {[59.28, 68.65]}} & \textcolor{delta}{$\uparrow$ 2.52} & \cellcolor{gray!15}\makecell{0.590 \\ {[0.537, 0.644]}} & \textcolor{delta}{$\uparrow$ 0.019}  
            & \cellcolor{gray!15}\makecell{88.62 \\ {[85.32, 91.93]}} & \textcolor{delta}{$\uparrow$ 1.46} & \cellcolor{gray!15}\makecell{0.819 \\ {[0.764, 0.872]}} & \textcolor{delta}{$\uparrow$ 0.022} 
            \\
            & & C  
            & \cellcolor{gray!15}\makecell{87.76 \\ {[82.31, 92.52]}} & \textcolor{delta}{$\uparrow$ 3.41} & \cellcolor{gray!15}\makecell{0.837 \\ {[0.769, 0.902]}} & \textcolor{delta}{$\uparrow$ 0.025} 
            & \cellcolor{gray!15}\makecell{65.05 \\ {[60.00, 69.73]}} & \textcolor{delta}{$\uparrow$ 3.61} & \cellcolor{gray!15}\makecell{0.609 \\ {[0.551, 0.664]}} & \textcolor{delta}{$\uparrow$ 0.038} 
            & \cellcolor{gray!15}\makecell{88.63 \\ {[85.32, 91.93]}} & \textcolor{delta}{$\uparrow$ 1.47} & \cellcolor{gray!15}\makecell{0.820 \\ {[0.768, 0.872]}} & \textcolor{delta}{$\uparrow$ 0.023} 
            \\
        \bottomrule
    \end{tabular}}
\end{table*}

\begin{table*}[!t]
    \centering
    \caption{Diagnostic performance comparison using eleven classifiers trained on \textit{Baseline} (A), \textit{Real-finetune} (B), and \textit{Joint-train} (C) paradigms in MosMedData~\citep{morozov2020mosmeddata} and MRNet~\citep{bien2018deep} datasets. 
    \underline{Underlined results} show no significant difference compared with \textit{Baseline} (A), all others indicate significant differences ($p<0.05$).
    95\% confidence intervals for both metrics are reported.
    “Hybrid”, the classifier with CNN-Transformer design. Acc., accuracy ($\%$).}
    \label{tab:results_full_shot_twosets}
    \resizebox{1.0\textwidth}{!}{
    \begin{tabular}{c c c c c c c c c c c}
        \toprule
        \multirow{2.5}{*}{\bf{Classifier}} & \multirow{2.5}{*}{\bf{Backbone}} & \multirow{2.5}{*}{\bf{\makecell[c]{Training\\Paradigm}}} & \multicolumn{4}{c}{\textbf{MosMedData}} & \multicolumn{4}{c}{\textbf{MRNet}} \\
        \cmidrule(lr){4-7} \cmidrule(lr){8-11}
        & & & Acc. & \textcolor{delta}{$\Delta$} & AUROC & \textcolor{delta}{$\Delta$} & Acc. & \textcolor{delta}{$\Delta$} & AUROC & \textcolor{delta}{$\Delta$} \\
        \midrule
        \multirow{3}{*}{\makecell[c]{I3D\\~\citep{carreira2017quo}}} & \multirow{3}{*}{CNN} 
            & A
            & 67.18 [64.35, 70.01] & & 0.566 [0.535, 0.597] & 
            & 63.49 [56.08, 70.37] & & 0.592 [0.532, 0.647] & 
            \\
            & & B
            & \cellcolor{gray!15}72.59 [69.88, 75.42] & \textcolor{delta}{$\uparrow$ 5.41} & \cellcolor{gray!15}0.609 [0.575, 0.646] & \textcolor{delta}{$\uparrow$ 0.043} 
            & \cellcolor{gray!15}71.43 [65.08, 77.78] & \textcolor{delta}{$\uparrow$ 7.94} & \cellcolor{gray!15}0.674 [0.616, 0.733] & \textcolor{delta}{$\uparrow$ 0.082}
            \\
            & & C 
            & \cellcolor{gray!15}69.50 [66.67, 72.33] & \textcolor{delta}{$\uparrow$ 2.32} & \cellcolor{gray!15}0.585 [0.551, 0.620] & \textcolor{delta}{$\uparrow$ 0.019} 
            & \cellcolor{gray!15}70.37 [64.02, 76.72] & \textcolor{delta}{$\uparrow$ 6.88} & \cellcolor{gray!15}0.665 [0.606, 0.723] & \textcolor{delta}{$\uparrow$ 0.073}
            \\
        \midrule
        \multirow{3}{*}{\makecell[c]{R(2+1)D\\~\citep{tran2018closer}}} & \multirow{3}{*}{CNN} 
            & A
            & 71.30 [68.34, 74.26] & & 0.572 [0.541, 0.603] & 
            & 67.72 [60.83, 74.07] & & 0.639 [0.579, 0.698] & 
            \\
            & & B
            & \cellcolor{gray!15}\underline{72.33} [69.50, 75.16] & \textcolor{delta}{$\uparrow$ 1.03} & \cellcolor{gray!15}\underline{0.610} [0.577, 0.644] & \textcolor{delta}{$\uparrow$ 0.038} 
            & \cellcolor{gray!15}69.31 [62.43, 76.19] & \textcolor{delta}{$\uparrow$ 1.59} & \cellcolor{gray!15}0.651 [0.592, 0.710] & \textcolor{delta}{$\uparrow$ 0.012}  
            \\
            & & C 
            & \cellcolor{gray!15}74.00 [71.30, 76.71] & \textcolor{delta}{$\uparrow$ 2.70} & \cellcolor{gray!15}0.601 [0.569, 0.633] & \textcolor{delta}{$\uparrow$ 0.029} 
            & \cellcolor{gray!15}\underline{68.78} [61.90, 75.15] & \textcolor{delta}{$\uparrow$ 1.06} & \cellcolor{gray!15}\underline{0.643} [0.584, 0.701] & \textcolor{delta}{$\uparrow$ 0.004} 
            \\
        \midrule
        \multirow{3}{*}{\makecell[c]{SlowFast\\~\citep{feichtenhofer2019slowfast}}} & \multirow{3}{*}{CNN} 
            & A
            & 65.38 [62.55, 68.21] & & 0.576 [0.541, 0.614] & 
            & 67.72 [60.85, 74.60] & & 0.632 [0.574, 0.690] & 
            \\
            & & B
            & \cellcolor{gray!15}73.75 [70.91, 76.45] & \textcolor{delta}{$\uparrow$ 8.37} & \cellcolor{gray!15}0.622 [0.586, 0.657] & \textcolor{delta}{$\uparrow$ 0.046} 
            & \cellcolor{gray!15}73.02 [66.67, 79.37] & \textcolor{delta}{$\uparrow$ 5.30} & \cellcolor{gray!15}0.688 [0.630, 0.744] & \textcolor{delta}{$\uparrow$ 0.056}  
            \\
            & & C  
            & \cellcolor{gray!15}69.11 [66.28, 72.07] & \textcolor{delta}{$\uparrow$ 3.73} & \cellcolor{gray!15}0.604 [0.568, 0.641] & \textcolor{delta}{$\uparrow$ 0.028} 
            & \cellcolor{gray!15}69.84 [63.49, 76.19] & \textcolor{delta}{$\uparrow$ 2.12} & \cellcolor{gray!15}0.662 [0.600, 0.720] & \textcolor{delta}{$\uparrow$ 0.030} 
            \\
        \midrule
        \multirow{3}{*}{\makecell[c]{CSN\\~\citep{tran2019video}}} & \multirow{3}{*}{CNN} 
            & A
            & 70.79 [67.95, 73.49] & & 0.636 [0.602, 0.671] & 
            & 71.96 [65.08, 78.31] & & 0.685 [0.623, 0.744] & 
            \\
            & & B
            & \cellcolor{gray!15}74.13 [71.17, 76.83] & \textcolor{delta}{$\uparrow$ 3.34} & \cellcolor{gray!15}0.643 [0.607, 0.679] & \textcolor{delta}{$\uparrow$ 0.007} 
            & \cellcolor{gray!15}75.66 [69.31, 81.48] & \textcolor{delta}{$\uparrow$ 3.70} & \cellcolor{gray!15}0.719 [0.663, 0.779] & \textcolor{delta}{$\uparrow$ 0.034}  
            \\
            & & C 
            & \cellcolor{gray!15}73.36 [70.52, 76.06] & \textcolor{delta}{$\uparrow$ 2.57} & \cellcolor{gray!15}0.641 [0.606, 0.679] & \textcolor{delta}{$\uparrow$ 0.005} 
            & \cellcolor{gray!15}73.54 [66.67, 79.38] & \textcolor{delta}{$\uparrow$ 1.58} & \cellcolor{gray!15}0.694 [0.636, 0.751] & \textcolor{delta}{$\uparrow$ 0.009} 
            \\
        \midrule
        \multirow{3}{*}{\makecell[c]{TPN\\~\citep{yang2020temporal}}} & \multirow{3}{*}{CNN} 
            & A
            & 74.65 [71.81, 77.48] & & 0.604 [0.574, 0.635] & 
            & 71.43 [64.55, 77.78] & & 0.682 [0.618, 0.744] & 
            \\
            & & B
            & \cellcolor{gray!15}76.32 [73.62, 79.28] & \textcolor{delta}{$\uparrow$ 1.67} & \cellcolor{gray!15}0.622 [0.590, 0.658] & \textcolor{delta}{$\uparrow$ 0.018} 
            & \cellcolor{gray!15}77.78 [71.96, 83.60] & \textcolor{delta}{$\uparrow$ 6.35} & \cellcolor{gray!15}0.742 [0.684, 0.798] & \textcolor{delta}{$\uparrow$ 0.060}  
            \\
            & & C 
            & \cellcolor{gray!15}76.96 [74.26, 79.67] & \textcolor{delta}{$\uparrow$ 2.31} & \cellcolor{gray!15}0.673 [0.639, 0.710] & \textcolor{delta}{$\uparrow$ 0.069} 
            & \cellcolor{gray!15}75.66 [69.31, 81.48] & \textcolor{delta}{$\uparrow$ 4.23} & \cellcolor{gray!15}0.722 [0.665, 0.784] & \textcolor{delta}{$\uparrow$ 0.040} 
            \\
        \midrule
        \multirow{3}{*}{\makecell[c]{TimeSformer\\~\citep{bertasius2021space}}} & \multirow{3}{*}{Transformer} 
            & A
            & 73.75 [71.04, 76.45] & & 0.500 [0.500, 0.500] & 
            & 58.20 [50.79, 65.08] & & 0.506 [0.500, 0.521] & 
            \\
            & & B
            & \cellcolor{gray!15}74.39 [71.69, 77.22] & \textcolor{delta}{$\uparrow$ 0.64} & \cellcolor{gray!15}0.511 [0.502, 0.521] & \textcolor{delta}{$\uparrow$ 0.011} 
            & \cellcolor{gray!15}66.14 [59.79, 72.49] & \textcolor{delta}{$\uparrow$ 7.94} & \cellcolor{gray!15}0.683 [0.620, 0.744] & \textcolor{delta}{$\uparrow$ 0.177}  
            \\
            & & C 
            & \cellcolor{gray!15}74.91 [72.20, 77.61] & \textcolor{delta}{$\uparrow$ 1.16} & \cellcolor{gray!15}0.520 [0.509, 0.534] & \textcolor{delta}{$\uparrow$ 0.020} 
            & \cellcolor{gray!15}67.20 [59.79, 74.07] & \textcolor{delta}{$\uparrow$ 9.00} & \cellcolor{gray!15}0.627 [0.570, 0.687] & \textcolor{delta}{$\uparrow$ 0.121} 
            \\
        \midrule
        \multirow{3}{*}{\makecell[c]{MViTv2\\~\citep{li2022mvitv2}}} & \multirow{3}{*}{Transformer} 
            & A
            & 73.75 [70.91, 76.45] & & 0.512 [0.498, 0.527] & 
            & 57.67 [50.26, 64.55] & & 0.500 [0.500, 0.500] & 
            \\
            & & B
            & \cellcolor{gray!15}74.52 [71.69, 77.22] & \textcolor{delta}{$\uparrow$ 0.77} & \cellcolor{gray!15}0.523 [0.507, 0.539] & \textcolor{delta}{$\uparrow$ 0.011} 
            & \cellcolor{gray!15}61.38 [53.97, 68.27] & \textcolor{delta}{$\uparrow$ 3.71} & \cellcolor{gray!15}0.544 [0.514, 0.578] & \textcolor{delta}{$\uparrow$ 0.044}  
            \\
            & & C 
            & \cellcolor{gray!15}\underline{74.13} [71.30, 76.71] & \textcolor{delta}{$\uparrow$ 0.38} & \cellcolor{gray!15}\underline{0.536} [0.515, 0.556] & \textcolor{delta}{$\uparrow$ 0.024} 
            & \cellcolor{gray!15}61.90 [54.50, 68.78] & \textcolor{delta}{$\uparrow$ 4.23} & \cellcolor{gray!15}0.552 [0.518, 0.589] & \textcolor{delta}{$\uparrow$ 0.052} 
            \\
        \midrule
        \multirow{3}{*}{\makecell[c]{VideoSwin\\~\citep{liu2022video}}} & \multirow{3}{*}{Transformer} 
            & A
            & 74.13 [71.30, 76.83] & & 0.515 [0.502, 0.529] & 
            & 64.55 [57.14, 71.43] & & 0.601 [0.543, 0.660] & 
            \\
            & & B
            & \cellcolor{gray!15}\underline{74.52} [71.69, 77.09] & \textcolor{delta}{$\uparrow$ 0.39} & \cellcolor{gray!15}\underline{0.528} [0.511, 0.547] & \textcolor{delta}{$\uparrow$ 0.013} 
            & \cellcolor{gray!15}\underline{64.65} [57.67, 71.43] & \textcolor{delta}{$\uparrow$ 0.10} & \cellcolor{gray!15}\underline{0.608} [0.544, 0.673] & \textcolor{delta}{$\uparrow$ 0.007}  
            \\
            & & C 
            & \cellcolor{gray!15}75.80 [72.84, 78.51] & \textcolor{delta}{$\uparrow$ 1.67} & \cellcolor{gray!15}0.628 [0.592, 0.665] & \textcolor{delta}{$\uparrow$ 0.113} 
            & \cellcolor{gray!15}66.67 [59.79, 73.02] & \textcolor{delta}{$\uparrow$ 2.12} & \cellcolor{gray!15}0.654 [0.584, 0.720] & \textcolor{delta}{$\uparrow$ 0.053} 
            \\
        \midrule
        \multirow{3}{*}{\makecell[c]{UniFormerV2\\~\citep{li2022uniformerv2}}} & \multirow{3}{*}{Hybrid} 
            & A
            & 74.26 [71.56, 76.96] & & 0.510 [0.501, 0.520] & 
            & 57.67 [50.26, 64.55] & & 0.500 [0.500, 0.500] & 
            \\
            & & B
            & \cellcolor{gray!15}\underline{74.91} [72.07, 77.61] & \textcolor{delta}{$\uparrow$ 0.65} & \cellcolor{gray!15}\underline{0.528} [0.511, 0.544] & \textcolor{delta}{$\uparrow$ 0.018} 
            & \cellcolor{gray!15}\underline{58.20} [51.32, 65.08] & \textcolor{delta}{$\uparrow$ 0.53} & \cellcolor{gray!15}\underline{0.619} [0.560, 0.677] & \textcolor{delta}{$\uparrow$ 0.019}  
            \\
            & & C  
            & \cellcolor{gray!15}\underline{74.52} [71.69, 77.09] & \textcolor{delta}{$\uparrow$ 0.26} & \cellcolor{gray!15}\underline{0.540} [0.520, 0.561] & \textcolor{delta}{$\uparrow$ 0.030} 
            & \cellcolor{gray!15}59.79 [52.38, 66.67] & \textcolor{delta}{$\uparrow$ 2.12} & \cellcolor{gray!15}0.547 [0.492, 0.602] & \textcolor{delta}{$\uparrow$ 0.047} 
            \\
        \midrule
        \multirow{3}{*}{\makecell[c]{FTC\\~\citep{ahmadi2023transformer}}} & \multirow{3}{*}{Hybrid} 
            & A
            & 74.39 [71.68, 77.10] & & 0.597 [0.581, 0.624] & 
            & 63.49 [60.57, 69.81] & & 0.588 [0.552, 0.651]  & 
            \\
            & & B
            & \cellcolor{gray!15}76.32 [73.24, 78.33] & \textcolor{delta}{$\uparrow$ 1.93} & \cellcolor{gray!15}0.630 [0.605, 0.663]& \textcolor{delta}{$\uparrow$ 0.033} 
            & \cellcolor{gray!15}66.13 [62.24, 70.99] & \textcolor{delta}{$\uparrow$ 2.64} & \cellcolor{gray!15}0.607 [0.563, 0.674] & \textcolor{delta}{$\uparrow$ 0.019} 
            \\
            & & C  
            & \cellcolor{gray!15}77.09 [74.70, 79.54] & \textcolor{delta}{$\uparrow$ 2.70} & \cellcolor{gray!15}0.641 [0.615, 0.675] & \textcolor{delta}{$\uparrow$ 0.044} 
            & \cellcolor{gray!15}68.25 [64.57, 72.10] & \textcolor{delta}{$\uparrow$ 4.76} & \cellcolor{gray!15}0.623 [0.577, 0.691] & \textcolor{delta}{$\uparrow$ 0.035} 
            \\
        \midrule
        \multirow{3}{*}{\makecell[c]{CSG-3DCT\\~\citep{zhou2023inflated}}} & \multirow{3}{*}{Hybrid} 
            & A
            & 72.33 [69.50, 75.03] & & 0.616 [0.579, 0.651] & 
            & 71.43 [64.55, 77.78] & & 0.677 [0.616, 0.735] & 
            \\
            & & B
            & \cellcolor{gray!15}74.00 [71.43, 76.83] & \textcolor{delta}{$\uparrow$ 1.67} & \cellcolor{gray!15}0.622 [0.589, 0.658] & \textcolor{delta}{$\uparrow$ 0.006} 
            & \cellcolor{gray!15}73.02 [66.67, 79.37] & \textcolor{delta}{$\uparrow$ 1.59} & \cellcolor{gray!15}0.688 [0.632, 0.745] & \textcolor{delta}{$\uparrow$ 0.011}  
            \\
            & & C  
            & \cellcolor{gray!15}74.78 [71.94, 77.61] & \textcolor{delta}{$\uparrow$ 2.45} & \cellcolor{gray!15}0.622 [0.587, 0.656] & \textcolor{delta}{$\uparrow$ 0.006} 
            & \cellcolor{gray!15}72.49 [66.14, 78.84] & \textcolor{delta}{$\uparrow$ 1.06} & \cellcolor{gray!15}0.688 [0.629, 0.748] & \textcolor{delta}{$\uparrow$ 0.011} 
            \\
        \bottomrule
    \end{tabular}}
\end{table*}

\subsection{Evaluation Metrics}
Existing sequence synthesis works evaluate the synthetic quality by mostly using Fr\'{e}chet Inception Distance (FID)~\citep{heusel2017gans} and Fr\'{e}chet Video Distance (FVD)~\citep{unterthiner2018towards}.
However, researchers have verified that these metrics do not consistently correlate with performance metrics on downstream tasks~\citep{yang2024freemask,luo2024measurement}.
To rigorously validate the effectiveness of our augmentation strategy and to thoroughly assess the value of synthetic data to downstream tasks, we conduct augmentation benchmarks and employ multiple popular classifiers to compare diagnostic accuracy and area under the receiver operating characteristic curve (AUROC).
Besides, we adopt the proposed VAE-Seq to assess the cross-frame/slice consistency.
Moreover, motivated by~\citep{huang2024vbench}, we quantify the smoothness of generated clips using the metric termed \textit{Dynamic Smoothness}, which is the mean absolute error between the reconstructed frames/slices from the interpolation model~\citep{li2023amt} and the original real ones.

\subsection{Statistical Analysis}
In our study, statistical significance between two classification methods evaluated on the same test set is assessed using McNemar’s test~\citep{dietterich1998approximate}.
We further report the error bars defined as 95\% confidence intervals (CIs) for accuracy and AUROC.
Uncertainty is quantified using non-parametric percentile bootstrap resampling of the test set (2,000 iterations), where samples are drawn with replacement and the metrics are recomputed for each resample.
Then, the 95\% CIs are obtained from the 2.5\textit{th} and 97.5\textit{th} percentiles. 
Therefore, the reported error bars (i.e., CIs) capture test-set sampling variability under a single overall run with fixed experimental conditions.

\subsection{Comparison with Augmentation Methods}
We conducted an extensive set of experiments comparing our method with 15 data augmentation techniques on all five datasets (Table~\ref{tab:aug_compare}).
Specifically, we grouped these methods into five categories following the augmentation taxonomy in~\citep{cheung2023survey}, which include: 
a) \textbf{Traditional-based}, i.e., Basic setting (refer to standard augmentation practices in Table~\ref{tab:dataset_setting}), Random elastic deformation; 
b) \textbf{Composition-based}, i.e., AutoAugment~\citep{cubuk2019autoaugment}, AdaAug~\citep{cheung2021adaaug}, GRA-Video~\citep{an2022group}, Sample-aware RandAugment~\citep{xiao2025sample}; 
c) \textbf{Mixing-based}, i.e., Mixup~\citep{zhang2017mixup}, CutMix~\citep{yun2019cutmix}, VideoMix~\citep{yun2020videomix}, LayerMix~\citep{ahmad2025layermix}; 
d) \textbf{GAN-based}, i.e., TGANv2~\citep{saito2020train}, StyleGAN-V~\citep{skorokhodov2022stylegan};
e) \textbf{Diffusion-based}, i.e., VideoComposer~\citep{wang2023videocomposer}, Endora~\citep{li2024endora}, 3D MedDiff~\citep{wang20253d}. Note that since few studies comprehensively examine the effect of video/3D generative models on downstream tasks, we adopted synthesis-oriented generators from both natural and medical domains as the generative augmentation schemes in our experiments.
All augmentation methods were based on the Basic setting and employed I3D~\citep{carreira2017quo} as a strong backbone for classification.
We kept the training datasets for all augmentation methods the same size to ensure fairness.

As shown in Table~\ref{tab:aug_compare}, our method consistently achieves the highest performance across five datasets on both evaluation metrics, with the majority of performance differences statistically significant ($p < 0.05$).
Our model achieves the best overall accuracy (Carotid: 85.03\%, TUSC: 65.41\%, ACDC: 87.16\%, MosMedData: 72.59\%, MRNet: 71.43\%) and AUROC (Carotid: 0.813, TUSC: 0.611, ACDC: 0.795, MosMedData: 0.609, MRNet: 0.674), significantly outperforming both traditional augmentations.
Notably, adding random elastic deformation does not yield noticeable performance improvements across all datasets (Basic: Average Acc. 70.89\%, Elastic deformation: Average Acc. 71.77\%), likely because such local geometric perturbations introduce only limited semantic variation.
Compared with composition-based augmentations, our method improves the average accuracy by 3.06\%, 4.24\%, 2.57\%, 2.80\%, and 5.03\%, and AUROC by 0.067, 0.046, 0.049, 0.018, and 0.060 on the Carotid, TUSC, ACDC, MosMedData, and MRNet datasets, respectively.
Although mixing-based augmentations yield better diagnostic results than composition-based ones in most cases, a clear performance gap remains compared with our method, with differences in mean accuracy/AUROC of 1.53\%/0.036, 2.98\%/0.033, 2.85\%/0.047, 1.93\%/0.038, and 2.65\%/0.028 across the five datasets.
Furthermore, as shown in Table~\ref{tab:aug_compare}, generative augmentation paradigms (including GAN- and diffusion-based variants) tend to achieve higher average performance than other augmentation methods, highlighting the potential of leveraging synthetic data to enhance classifier learning.
Within generative augmentations, GAN-based variants generally underperform diffusion-based ones, indicating that the limited fidelity and diversity of GAN-generated samples constrain the magnitude of the generative augmentation gains.
In contrast, diffusion-based competitors display stronger overall recognition performance due to their more faithful and diverse synthesis, reaching mean accuracies of 82.08\%, 63.36\%, 84.71\%, 70.83\%, and 68.60\%, and AUROCs of 0.774, 0.588, 0.754, 0.591, and 0.649 on the Carotid, TUSC, ACDC, MosMedData, and MRNet datasets, respectively.
Based on the experimental results, with the combination of customized sequence synthesis and pruning of noisy synthetic samples, our diffusion-based approach functions as a powerful and general data augmentation suite across multiple organs and medical modalities.

\begin{figure*}[!t]
	\centering
    \includegraphics[width=0.95\linewidth]{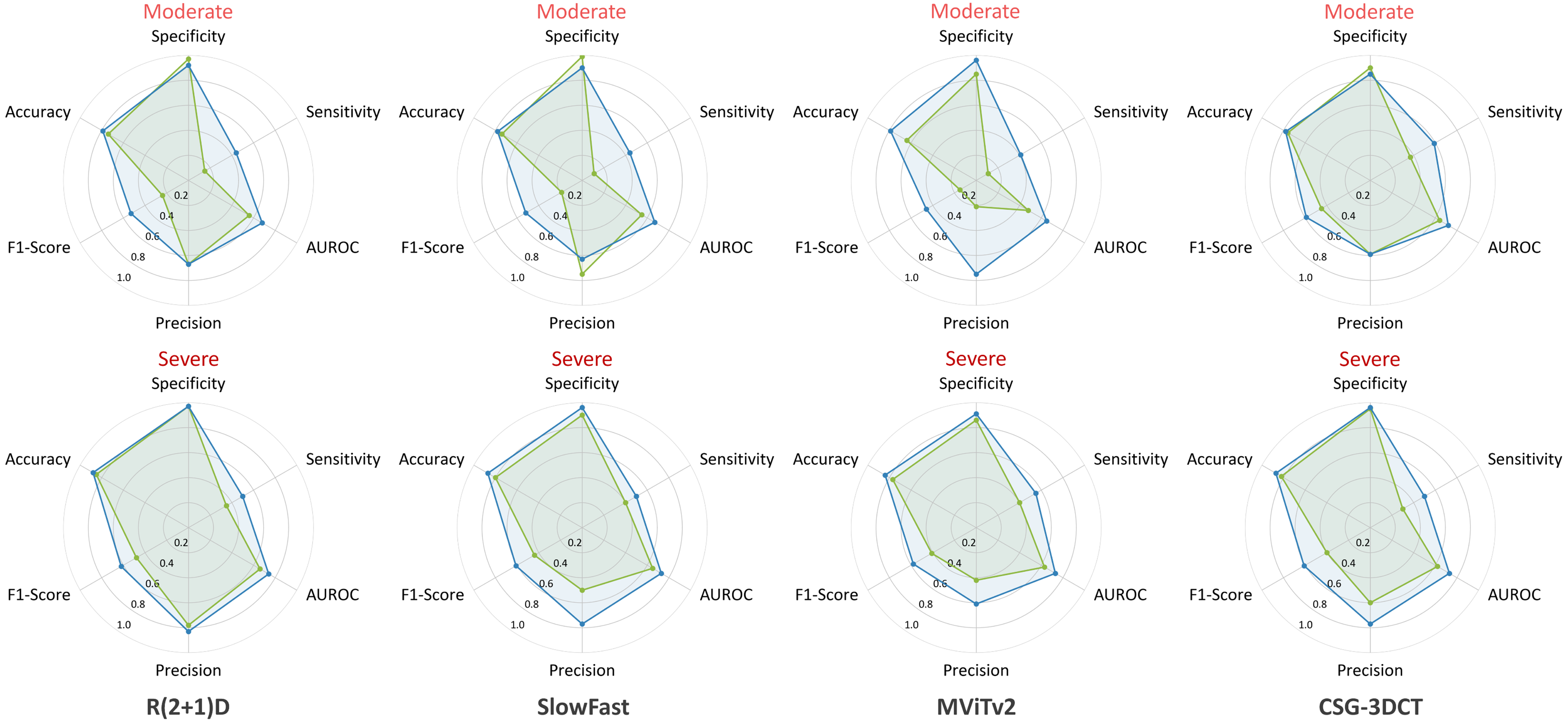}
	\caption{Carotid performance radar map, using 4 models trained on \textit{Baseline} (green) and \textit{Joint-train} (blue) paradigms in underrepresented high-risk sets. Our proposed framework generally enhances diagnostic performance across multiple evaluation metrics within these challenging subsets.}
	\label{fig:exp2_radar}
\end{figure*}

\subsection{Generative Augmentation for Diagnosis with Multiple Sequence Classifiers}
We utilized \textit{Real-finetune} and \textit{Joint-train} paradigms to integrate real and synthetic clips for downstream multi-class sequence classification training.
The efficacy of our framework was fully validated across three medical datasets using eleven common classifiers.
To demonstrate the generalizability of \textit{\M}, the selected networks involve 3D CNN, transformer, and CNN-Transformer hybrid designs, which specialize in capturing local, global, and mixed features, respectively.

Table~\ref{tab:results_full_shot_threesets} and Table~\ref{tab:results_full_shot_twosets} compare the average performance (i.e., accuracy and AUROC) on different training paradigms across a wide range of sequence classifiers, including \textbf{five} CNN-based classic models, \textbf{three} transformer-based models, and \textbf{three} CNN-Transformer hybrid-based methods.
It is evident that the performance ceiling of existing popular models, from simple CNN-based to complex CNN-Transformer hybrids, can be further elevated using synthetic samples obtained from our proposed framework, under \textit{Real-finetune} and \textit{Joint-train} paradigms, with most performance improvements being statistically significant ($p < 0.05$).
On the carotid dataset, the accuracy/AUROC can be most boosted by 5.44\%/0.077 and 6.80\%/0.086 through \textit{Real-finetune} and \textit{Joint-train} paradigms, respectively.
Due to high intra-class variations in TUSC, MosMedData, and MRNet datasets, the \textit{Baseline} exhibits visibly inferior results, with average accuracies of 62.42\%, 71.99\%, and 65.03\%, respectively, compared to other datasets.
Nevertheless, it can be seen from Table~\ref{tab:results_full_shot_threesets} and Table~\ref{tab:results_full_shot_twosets} that samples synthesized by \textit{\M} make significant contributions and enhance the classification performance to a practically acceptable level.
On TUSC, CSN harvests the largest performance gain among all classifiers, which is non-trivially enhanced by 5.41\%/0.061 and 3.61\%/0.040 accuracy/AUROC under \textit{Real-finetune} and \textit{Joint-train} paradigms, respectively.
On MosMedData and MRNet, high-quality synthetic samples similarly play a crucial role in driving performance improvements, for example, they improve the 11 classifiers by +2.02\% and +3.56\% in average accuracy under the \textit{Joint-train} paradigm, respectively.
Notably, for ACDC, the overall performance improvement remains significant even with a high baseline.
For instance, CSG-3DCT is boosted from 87.16\%$\rightarrow$88.63\% (+1.47\%) accuracy and from 0.797$\rightarrow$0.820 (+0.023) AUROC under the \textit{Joint-train} practice.
Extensive evaluations across multiple classifiers reveal that the gains brought by \textit{\M} are architecture-agnostic, further underscoring that it serves as a general-purpose augmentation tool, effectively strengthening diverse diagnostic networks across multiple medical sequence analysis tasks.

\begin{table*}
    \centering
    \scriptsize
    \caption{Out-domain performance comparison using eleven nets trained on \textit{Baseline} (A), \textit{Real-finetune} (B), and \textit{Joint-train} (C) paradigms in carotid dataset.
    \underline{Underlined results} show no significant difference compared with \textit{Baseline} (A), all others indicate significant differences ($p<0.05$).
    95\% confidence intervals for both metrics are reported.
    “Hybrid”, the net with CNN-Transformer design. Acc., accuracy ($\%$).}
    \label{tab:results_od}
    \resizebox{0.71\textwidth}{!}{
    \begin{tabular}{c c c c c c c}
        \toprule
        \multirow{2.5}{*}{\bf{Classifier}} & \multirow{2.5}{*}{\bf{Backbone}} & \multirow{2.5}{*}{\bf{\makecell[c]{Training\\Paradigm}}} & \multicolumn{4}{c}{\bf{Carotid}} \\
        \cmidrule(lr){4-7}
        & & & Acc. & \textcolor{delta}{$\Delta$} & AUROC & \textcolor{delta}{$\Delta$} \\
        \midrule
        \multirow{3}{*}{\makecell[c]{I3D\\~\citep{carreira2017quo}}} & \multirow{3}{*}{CNN} 
            & A
            & 70.00 [60.00, 80.00] & & 0.609 [0.509, 0.721] & 
            \\
            & & B
            & \cellcolor{gray!15}\underline{71.67} [61.67, 81.67] & \textcolor{delta}{$\uparrow$ 1.67} & \cellcolor{gray!15}\underline{0.648} [0.537, 0.771] & \textcolor{delta}{$\uparrow$ 0.039} 
            \\
            & & C 
            & \cellcolor{gray!15}75.00 [65.00, 85.00] & \textcolor{delta}{$\uparrow$ 5.00} & \cellcolor{gray!15}0.682 [0.565, 0.794] & \textcolor{delta}{$\uparrow$ 0.073} 
            \\
        \midrule
        \multirow{3}{*}{\makecell[c]{R(2+1)D\\~\citep{tran2018closer}}} & \multirow{3}{*}{CNN} 
            & A
            & 66.67 [56.67, 76.71] & & 0.577 [0.471, 0.690] &
            \\
            & & B
            & \cellcolor{gray!15}71.67 [61.67, 81.67] & \textcolor{delta}{$\uparrow$ 5.00} & \cellcolor{gray!15}0.661 [0.556, 0.773] & \textcolor{delta}{$\uparrow$ 0.084}  
            \\
            & & C 
            & \cellcolor{gray!15}75.00 [65.00, 85.00] & \textcolor{delta}{$\uparrow$ 8.33} & \cellcolor{gray!15}0.689 [0.575, 0.810] & \textcolor{delta}{$\uparrow$ 0.112} 
            \\
        \midrule
        \multirow{3}{*}{\makecell[c]{SlowFast\\~\citep{feichtenhofer2019slowfast}}} & \multirow{3}{*}{CNN} 
            & A
            & 73.33 [63.33, 83.33] & & 0.638 [0.528, 0.750] &
            \\
            & & B
            & \cellcolor{gray!15}\underline{75.00} [65.00, 85.00] & \textcolor{delta}{$\uparrow$ 1.67} & \cellcolor{gray!15}\underline{0.670} [0.562, 0.787] & \textcolor{delta}{$\uparrow$ 0.032}  
            \\
            & & C 
            & \cellcolor{gray!15}\underline{75.00} [65.00, 85.00] & \textcolor{delta}{$\uparrow$ 1.67} & \cellcolor{gray!15}\underline{0.706} [0.588, 0.817] & \textcolor{delta}{$\uparrow$ 0.068} 
            \\
        \midrule
        \multirow{3}{*}{\makecell[c]{CSN\\~\citep{tran2019video}}} & \multirow{3}{*}{CNN} 
            & A
            & 68.33 [58.33, 78.33] & & 0.592 [0.494, 0.703] &
            \\
            & & B
            & \cellcolor{gray!15}73.33 [63.33, 83.33] & \textcolor{delta}{$\uparrow$ 5.00} & \cellcolor{gray!15}0.664 [0.556, 0.786] & \textcolor{delta}{$\uparrow$ 0.072}  
            \\
            & & C 
            & \cellcolor{gray!15}71.67 [61.67, 81.67] & \textcolor{delta}{$\uparrow$ 3.34} & \cellcolor{gray!15}0.672 [0.552, 0.784] & \textcolor{delta}{$\uparrow$ 0.080} 
            \\
        \midrule
        \multirow{3}{*}{\makecell[c]{TPN\\~\citep{yang2020temporal}}} & \multirow{3}{*}{CNN} 
            & A
            & 75.00 [65.00, 85.00] & & 0.667 [0.563, 0.783] &
            \\
            & & B
            & \cellcolor{gray!15}\underline{76.67} [66.67, 86.67] & \textcolor{delta}{$\uparrow$ 1.67} & \cellcolor{gray!15}\underline{0.731} [0.624, 0.831] & \textcolor{delta}{$\uparrow$ 0.064}
            \\
            & & C 
            & \cellcolor{gray!15}\underline{76.67} [66.67, 86.67] & \textcolor{delta}{$\uparrow$ 1.67} & \cellcolor{gray!15}\underline{0.678} [0.569, 0.790] & \textcolor{delta}{$\uparrow$ 0.011} 
            \\
        \midrule
        \multirow{3}{*}{\makecell[c]{TimeSformer\\~\citep{bertasius2021space}}} & \multirow{3}{*}{Transformer} 
            & A
            & 68.33 [58.33, 78.33] & & 0.590 [0.489, 0.689] &
            \\
            & & B
            & \cellcolor{gray!15}71.67 [61.62, 81.67] & \textcolor{delta}{$\uparrow$ 3.34} & \cellcolor{gray!15}0.665 [0.543, 0.787] & \textcolor{delta}{$\uparrow$ 0.075}  
            \\
            & & C 
            & \cellcolor{gray!15}73.33 [63.33, 83.33] & \textcolor{delta}{$\uparrow$ 5.00} & \cellcolor{gray!15}0.664 [0.565, 0.773] & \textcolor{delta}{$\uparrow$ 0.074} 
            \\
        \midrule
        \multirow{3}{*}{\makecell[c]{MViTv2\\~\citep{li2022mvitv2}}} & \multirow{3}{*}{Transformer} 
            & A
            & 66.67 [56.67, 76.67] & & 0.590 [0.480, 0.699] &
            \\
            & & B
            & \cellcolor{gray!15}\underline{68.33} [58.33, 78.33] & \textcolor{delta}{$\uparrow$ 1.66} & \cellcolor{gray!15}\underline{0.592} [0.493, 0.699] & \textcolor{delta}{$\uparrow$ 0.002}  
            \\
            & & C 
            & \cellcolor{gray!15}70.00 [60.00, 80.00] & \textcolor{delta}{$\uparrow$ 3.33} & \cellcolor{gray!15}0.604 [0.508, 0.706] & \textcolor{delta}{$\uparrow$ 0.014} 
            \\
        \midrule
        \multirow{3}{*}{\makecell[c]{VideoSwin\\~\citep{liu2022video}}} & \multirow{3}{*}{Transformer} 
            & A
            & 65.00 [55.00, 75.00] & & 0.574 [0.462, 0.688] &
            \\
            & & B
            & \cellcolor{gray!15}70.00 [58.33, 80.00] & \textcolor{delta}{$\uparrow$ 5.00} & \cellcolor{gray!15}0.650 [0.527, 0.767] & \textcolor{delta}{$\uparrow$ 0.076}  
            \\
            & & C 
            & \cellcolor{gray!15}68.33 [58.33, 78.33] & \textcolor{delta}{$\uparrow$ 3.33} & \cellcolor{gray!15}0.609 [0.503, 0.728] & \textcolor{delta}{$\uparrow$ 0.035} 
            \\
        \midrule
        \multirow{3}{*}{\makecell[c]{UniFormerV2\\~\citep{li2022uniformerv2}}} & \multirow{3}{*}{Hybrid} 
            & A
            & 66.67 [56.67, 76.67] & & 0.617 [0.498, 0.735] &
            \\
            & & B
            & \cellcolor{gray!15}73.33 [63.33, 83.33] & \textcolor{delta}{$\uparrow$ 6.66} & \cellcolor{gray!15}0.681 [0.564, 0.803] & \textcolor{delta}{$\uparrow$ 0.064}  
            \\
            & & C 
            & \cellcolor{gray!15}\underline{68.33} [58.33, 78.33] & \textcolor{delta}{$\uparrow$ 1.66} & \cellcolor{gray!15}\underline{0.633} [0.521, 0.756] & \textcolor{delta}{$\uparrow$ 0.016} 
            \\
        \midrule
        \multirow{3}{*}{\makecell[c]{FTC\\~\citep{ahmadi2023transformer}}} & \multirow{3}{*}{Hybrid} 
            & A
            & 67.08 [56.67, 78.33] & & 0.616 [0.502, 0.741] &
            \\
            & & B
            & \cellcolor{gray!15}71.31 [66.67, 80.00] & \textcolor{delta}{$\uparrow$ 4.23} & \cellcolor{gray!15}0.632 [0.523, 0.752] & \textcolor{delta}{$\uparrow$ 0.016}  
            \\
            & & C 
            & \cellcolor{gray!15}74.68 [66.67, 83.33] & \textcolor{delta}{$\uparrow$ 7.60} & \cellcolor{gray!15}0.670 [0.553, 0.784] & \textcolor{delta}{$\uparrow$ 0.054} 
            \\
        \midrule
        \multirow{3}{*}{\makecell[c]{CSG-3DCT\\~\citep{zhou2023inflated}}} & \multirow{3}{*}{Hybrid} 
            & A
            & 68.33 [58.33, 78.33] & & 0.599 [0.491, 0.715] &
            \\
            & & B
            & \cellcolor{gray!15}73.33 [63.33, 83.33] & \textcolor{delta}{$\uparrow$ 5.00} & \cellcolor{gray!15}0.689 [0.569, 0.805] & \textcolor{delta}{$\uparrow$ 0.090} 
            \\
            & & C  
            & \cellcolor{gray!15}\underline{70.00} [60.00, 80.00] & \textcolor{delta}{$\uparrow$ 1.67} & \cellcolor{gray!15}\underline{0.651} [0.530, 0.772] & \textcolor{delta}{$\uparrow$ 0.052}
            \\
        \bottomrule
    \end{tabular}}
\end{table*}

\subsection{Improved Diagnosis in Underrepresented High-risk Sets}
We tested the effectiveness of \textit{\M} by evaluating the diagnostic performance in underrepresented high-risk situations using our carotid dataset.
In our study, \textit{underrepresented} represents the tail data with small amounts (i.e., moderate\&severe), and \textit{overrepresented} defines mild data with sufficient amounts.
To build a class-biased dataset, we skewed the real training dataset by randomly sampling 25\% clips from each high-risk class (moderate\&severe) in the underrepresented sets.
We leveraged \textit{\M} to expand the clips of each high-risk class to match the number of the low-risk class (mild), resulting in an augmented dataset.
Four downstream networks were trained on the augmented dataset with \textit{Joint-train} paradigm, and six metrics were used for class-level evaluation, including sensitivity, specificity, accuracy, F1-score, precision, and AUROC.

Fig.~\ref{fig:exp2_radar} quantitatively compares the carotid stenosis diagnostic performance of underrepresented high-risk sets on \textit{Baseline} and \textit{Joint-train} paradigms.
It shows that \textit{Baseline} performs poorly on underrepresented high-risk sets, especially in terms of sensitivity (average 18.52\% in moderate and 36.25\% in severe sets).
In comparison, the classifiers can be greatly enhanced by jointly training with synthetic data (e.g., \textit{SlowFast}: 11.11\%$\rightarrow$44.44\% sensitivity on moderate set).

We also provided the category-level analysis, and summarized the average F1 scores across four classifiers for a global observation and conclusion.
Specifically, \textit{Baseline} reached 69.92\%, 25.92\%, and 43.44\% average F1 scores for mild, moderate, and severe cases, respectively.
The F1 performance of \textit{Joint-train} in these three categories was 82.44\%, 52.73\%, and 60.40\%.
We found that well-designed classifiers jointly trained on synthetic samples from \textit{\M} and real ones, can effectively close diagnostic performance gaps between overrepresented (i.e., mild) and underrepresented sets (i.e., moderate and severe) while improving the former ($\sim$13\%$\uparrow$).

\subsection{Classification Robustness in Out-domain Conditions}
We further explored the impact of synthetic samples from \textit{\M} on diagnostic robustness in out-domain settings using the carotid dataset.
Training data of the sequence generator consisted of labeled ID and unlabeled OD subsets.
We added the hospital identifier in the descriptive text of the ID data to condition the domain distribution.
For the OD data, which did not contain the diagnostic class label, we padded the corresponding conditioning vector with zeros, while solely preserving hospital IDs to form descriptive texts.
As shown in Table~\ref{tab:results_od}, the average accuracy and AUROC of all eleven classifiers display better results on the \textit{Real-finetune} and \textit{Joint-train} paradigms compared to the \textit{Baseline}, with most of the improvements showing statistical significance ($p < 0.05$).
It validates that diagnostic robustness can be enhanced with the aid of \textit{\M} in scenarios where we only have access to unlabeled cases from additional medical centers due to limited resources.

\begin{table*}[!t]
    \centering
    \caption{Ablation study for different conditional controls in multi-organ and multi-modal datasets, including class label (C), text (T), image prior (I), and motion field (MF). SlowFast~\citep{feichtenhofer2019slowfast} (C1), CSN~\citep{tran2019video} (C2), and FTC~\citep{ahmadi2023transformer} (C3) trained on \textit{Joint-train} paradigm were used for downstream diagnosis. The accuracy and AUROC gains over the \textit{Baseline} are reported. “Hybrid”, the classifier with CNN-Transformer design.}
    \label{tab:ablation_cond}
    \resizebox{1.0\textwidth}{!}{
    \begin{tabular}{c c c c c c c c|c c c c}
        \toprule
        \multirow{4.25}{*}{\bf{Dataset}} & \multicolumn{7}{c|}{\bf{Synthesis}} & \multicolumn{4}{c}{\bf{Downstream Diagnosis}} \\
        \cmidrule(lr){2-12}
        & \multicolumn{4}{c}{Control} & \multicolumn{3}{c|}{Metric} & & \multirow{2.5}{*}{Backbone} & \multirow{2.5}{*}{$\Delta$Accuracy} & \multirow{2.5}{*}{$\Delta$AUROC} \\
        \cmidrule(lr){2-5} \cmidrule(lr){6-8}
        & C & T & I & MF & FVD$\downarrow$ & VAE-Seq$\uparrow$ & Dynamic Smoothness$\uparrow$ \\
        \midrule
        \multirow{13.5}{*}{\bf{Carotid}} & \multirow{3}{*}{\checkmark} & & \multirow{3}{*}{\checkmark} & \multirow{3}{*}{\checkmark} & \multirow{3}{*}{\bf{4.18}} & \multirow{3}{*}{89.65\%} & \multirow{3}{*}{93.27\%} & C1 & CNN & $\uparrow$ 2.04\% & $\uparrow$ 0.009  \\
        & & & & & & & & C2 & CNN & $\downarrow$ 0.68\% & $\downarrow$ 0.009 \\
        & & & & & & & & C3 & Hybrid & $\uparrow$ 0.56\% & $\uparrow$ 0.031  \\
        \cmidrule(lr){2-12}
        & \multirow{3}{*}{\checkmark} & \multirow{3}{*}{\checkmark} & & \multirow{3}{*}{\checkmark} & \multirow{3}{*}{6.50} & \multirow{3}{*}{89.08\%} & \multirow{3}{*}{92.85\%} & C1 & CNN & $\uparrow$ 1.36\% & $\uparrow$ 0.046 \\
        & & & & & & & & C2 & CNN & $\downarrow$ 0.68\% & $\downarrow$ 0.011 \\
        & & & & & & & & C3 & Hybrid & $\uparrow$ 1.12\% & $\uparrow$ 0.021 \\
        \cmidrule(lr){2-12}
        & \multirow{3}{*}{\checkmark} & \multirow{3}{*}{\checkmark} & \multirow{3}{*}{\checkmark} & & \multirow{3}{*}{4.29} & \multirow{3}{*}{88.74\%} & \multirow{3}{*}{93.09\%} & C1 & CNN & $\uparrow$ 2.72\% & $\uparrow$ 0.078  \\
        & & & & & & & & C2 & CNN & $\uparrow$ 4.09\% & $\uparrow$ 0.094 \\
        & & & & & & & & C3 & Hybrid & $\uparrow$ 3.28\% & $\uparrow$ 0.064  \\
        \cmidrule(lr){2-12}
        & \multirow{3}{*}{\checkmark} & \multirow{3}{*}{\checkmark} & \multirow{3}{*}{\checkmark} & \multirow{3}{*}{\checkmark} & \multirow{3}{*}{4.26} & \multirow{3}{*}{\bf{91.46\%}} & \multirow{3}{*}{\bf{94.29\%}} & C1 & CNN & \textcolor{delta}{$\uparrow$ 3.40\%} & \textcolor{delta}{$\uparrow$ 0.069} \\
        & & & & & & & & C2 & CNN & \textcolor{delta}{$\uparrow$ 6.80\%} & \textcolor{delta}{$\uparrow$ 0.086} \\
        & & & & & & & & C3 & Hybrid & \textcolor{delta}{$\uparrow$ 4.81\%} & \textcolor{delta}{$\uparrow$ 0.035} \\
        \midrule
        \multirow{13.5}{*}{\makecell[c]{\bf{TUSC}\\~\citep{TUSC}}} & \multirow{3}{*}{\checkmark} & & \multirow{3}{*}{\checkmark} & \multirow{3}{*}{\checkmark} & \multirow{3}{*}{4.18} & \multirow{3}{*}{90.40\%} & \multirow{3}{*}{95.03\%} & C1 & CNN & $\downarrow$ 1.80\% & $\downarrow$ 0.015 \\
        & & & & & & & & C2 & CNN & $\uparrow$ 1.44\% & $\uparrow$ 0.022 \\
        & & & & & & & & C3 & Hybrid & $\uparrow$ 1.08\% & $\uparrow$ 0.030 \\
        \cmidrule(lr){2-12}
        & \multirow{3}{*}{\checkmark} & \multirow{3}{*}{\checkmark} & & \multirow{3}{*}{\checkmark} & \multirow{3}{*}{7.01} & \multirow{3}{*}{90.51\%} & \multirow{3}{*}{\bf{96.91\%}} & C1 & CNN & $\downarrow$ 1.08\% & $\downarrow$ 0.001 \\
        & & & & & & & & C2 & CNN & $\downarrow$ 0.36\% & $\downarrow$ 0.008 \\
        & & & & & & & & C3 & Hybrid & $\downarrow$ 1.44\% & $\downarrow$ 0.011 \\
        \cmidrule(lr){2-12}
        & \multirow{3}{*}{\checkmark} & \multirow{3}{*}{\checkmark} & \multirow{3}{*}{\checkmark} & & \multirow{3}{*}{5.59} & \multirow{3}{*}{91.56\%} & \multirow{3}{*}{94.66\%} & C1 & CNN & $\uparrow$ 1.80\% & $\uparrow$ 0.017 \\
        & & & & & & & & C2 & CNN & $\uparrow$ 1.08\% & $\uparrow$ 0.014 \\
        & & & & & & & & C3 & Hybrid & $\uparrow$ 2.16\% & $\uparrow$ 0.016 \\
        \cmidrule(lr){2-12}
        & \multirow{3}{*}{\checkmark} & \multirow{3}{*}{\checkmark} & \multirow{3}{*}{\checkmark} & \multirow{3}{*}{\checkmark} & \multirow{3}{*}{\bf{3.69}} & \multirow{3}{*}{\bf{92.14\%}} & \multirow{3}{*}{95.59\%} & C1 & CNN & \textcolor{delta}{$\uparrow$ 2.53\%} & \textcolor{delta}{$\uparrow$ 0.030} \\
        & & & & & & & & C2 & CNN & \textcolor{delta}{$\uparrow$ 3.61\%} & \textcolor{delta}{$\uparrow$ 0.040} \\
        & & & & & & & & C3 & Hybrid & \textcolor{delta}{$\uparrow$ 3.57\%} & \textcolor{delta}{$\uparrow$ 0.030} \\
        \midrule
        \multirow{10.5}{*}{\makecell[c]{\bf{ACDC}\\~\citep{bernard2018deep}}} & \multirow{3}{*}{\checkmark} & \multirow{3}{*}{-} & \multirow{3}{*}{\checkmark} & & \multirow{3}{*}{8.70} & \multirow{3}{*}{76.74\%} & \multirow{3}{*}{92.50\%} & C1 & CNN & $\uparrow$ 1.46\% & $\uparrow$ 0.010\\
        & & & & & & & & C2 & CNN & $\uparrow$ 0.37\% & $\uparrow$  0.027\\
        & & & & & & & & C3 & Hybrid & $\uparrow$ 1.10\% & $\uparrow$ 0.051  \\
        \cmidrule(lr){2-12}
        & \multirow{3}{*}{\checkmark} & \multirow{3}{*}{-} & & \multirow{3}{*}{\checkmark} & \multirow{3}{*}{10.81} & \multirow{3}{*}{\bf{79.66\%}} & \multirow{3}{*}{\bf{93.89\%}} & C1 & CNN & $\uparrow$ 0.37\% & $\uparrow$ 0.020\\
        & & & & & & & & C2 & CNN & $\uparrow$ 0.74\% & $\uparrow$ 0.012\\
        & & & & & & & & C3 & Hybrid & $\uparrow$ 1.83\% & $\uparrow$ 0.029  \\
        \cmidrule(lr){2-12}
        & \multirow{3}{*}{\checkmark} & \multirow{3}{*}{-} & \multirow{3}{*}{\checkmark} & \multirow{3}{*}{\checkmark} & \multirow{3}{*}{\bf{8.18}} & \multirow{3}{*}{77.08\%} & \multirow{3}{*}{92.77\%} & C1 & CNN & \textcolor{delta}{$\uparrow$ 2.20\%} & \textcolor{delta}{$\uparrow$ 0.040} \\
        & & & & & & & & C2 & CNN & \textcolor{delta}{$\uparrow$ 2.93\%} & \textcolor{delta}{$\uparrow$ 0.049} \\
        & & & & & & & & C3 & Hybrid & \textcolor{delta}{$\uparrow$ 3.43\%} & \textcolor{delta}{$\uparrow$ 0.029} \\
        \midrule
        \multirow{10.5}{*}{\makecell[c]{\bf{MosMedData}\\~\citep{morozov2020mosmeddata}}} & \multirow{3}{*}{\checkmark} & \multirow{3}{*}{-} & \multirow{3}{*}{\checkmark} & & \multirow{3}{*}{2.98} & \multirow{3}{*}{83.06\%} & \multirow{3}{*}{95.27\%}
        & C1 & CNN &  $\uparrow$ 2.96\% & $\uparrow$ 0.004 \\
        & & & & & & & & C2 & CNN & $\uparrow$ 1.41\% & $\downarrow$ 0.091 \\
        & & & & & & & & C3 & Hybrid & $\uparrow$ 1.15\% & $\uparrow$ 0.070  \\
        \cmidrule(lr){2-12}
        & \multirow{3}{*}{\checkmark} & \multirow{3}{*}{-} & & \multirow{3}{*}{\checkmark} & \multirow{3}{*}{5.05} & \multirow{3}{*}{\bf{84.72\%}} & \multirow{3}{*}{\bf{96.35\%}} & C1 & CNN & $\uparrow$ 1.93\% & $\downarrow$ 0.026 \\
        & & & & & & & & C2 & CNN & $\uparrow$ 0.38\% & $\downarrow$ 0.114 \\
        & & & & & & & & C3 & Hybrid & $\uparrow$ 1.81\% & $\uparrow$ 0.054  \\
        \cmidrule(lr){2-12}
        & \multirow{3}{*}{\checkmark} & \multirow{3}{*}{-} & \multirow{3}{*}{\checkmark} & \multirow{3}{*}{\checkmark} & \multirow{3}{*}{\bf{2.91}} & \multirow{3}{*}{84.07\%} & \multirow{3}{*}{96.33\%} & C1 & CNN & \textcolor{delta}{$\uparrow$ 3.73\%} & \textcolor{delta}{$\uparrow$ 0.028} \\
        & & & & & & & & C2 & CNN & \textcolor{delta}{$\uparrow$ 2.57\%} & \textcolor{delta}{$\uparrow$ 0.005} \\
        & & & & & & & & C3 & Hybrid & \textcolor{delta}{$\uparrow$ 2.70\%} & \textcolor{delta}{$\uparrow$ 0.044} \\
        \midrule
        \multirow{10.5}{*}{\makecell[c]{\bf{MRNet}\\~\citep{bien2018deep}}} & \multirow{3}{*}{\checkmark} & \multirow{3}{*}{-} & \multirow{3}{*}{\checkmark} & & \multirow{3}{*}{8.48} & \multirow{3}{*}{75.48\%} & \multirow{3}{*}{92.90\%} 
        & C1 & CNN & $\uparrow$ 2.12\% & $\uparrow$ 0.018 \\
        & & & & & & & & C2 & CNN & $\uparrow$ 0.53\% & $\uparrow$ 0.045 \\
        & & & & & & & & C3 & Hybrid & $\uparrow$ 3.17\% & $\uparrow$ 0.057 \\
        \cmidrule(lr){2-12}
        & \multirow{3}{*}{\checkmark} & \multirow{3}{*}{-} & & \multirow{3}{*}{\checkmark} & \multirow{3}{*}{11.67} & \multirow{3}{*}{76.67\%} & \multirow{3}{*}{93.61\%} & C1 & CNN & $\uparrow$ 1.59\% & $\uparrow$ 0.017\\
        & & & & & & & & C2 & CNN & $\downarrow$ 2.12\% & $\downarrow$  0.020\\
        & & & & & & & & C3 & Hybrid & $\uparrow$ 1.06\% & $\uparrow$ 0.026 \\
        \cmidrule(lr){2-12}
        & \multirow{3}{*}{\checkmark} & \multirow{3}{*}{-} & \multirow{3}{*}{\checkmark} & \multirow{3}{*}{\checkmark} & \multirow{3}{*}{\bf{8.14}} & \multirow{3}{*}{\bf{77.10\%}} & \multirow{3}{*}{\bf{93.81\%}} & C1 & CNN & \textcolor{delta}{$\uparrow$ 2.12\%} & \textcolor{delta}{$\uparrow$ 0.030} \\
        & & & & & & & & C2 & CNN & \textcolor{delta}{$\uparrow$ 1.58\%} & \textcolor{delta}{$\uparrow$ 0.009} \\
        & & & & & & & & C3 & Hybrid & \textcolor{delta}{$\uparrow$ 4.76\%} & \textcolor{delta}{$\uparrow$ 0.035} \\
        \bottomrule
    \end{tabular}}
\end{table*}

\begin{table*}
    \centering
    \caption{Ablation study for the sequential augmentation module of the generator in multi-organ and multi-modal datasets. SlowFast~\citep{feichtenhofer2019slowfast} (C1), CSN~\citep{tran2019video} (C2), and FTC~\citep{ahmadi2023transformer} (C3) trained on \textit{Joint-train} paradigm were used for downstream diagnosis. The accuracy and AUROC gains over the \textit{Baseline} are reported. “Hybrid”, the classifier with CNN-Transformer design.}
    \label{tab:ablation_sam}
    \resizebox{1.0\textwidth}{!}{
    \begin{tabular}{c c c c c|c c c c}
        \toprule
        \multirow{2.5}{*}{\bf{Dataset}} & \multicolumn{4}{c|}{\bf{Synthesis}} & \multicolumn{4}{c}{\bf{Downstream Diagnosis}} \\
        \cmidrule(lr){2-9}
        & & FVD$\downarrow$ & VAE-Seq$\uparrow$ & Dynamic Smoothness$\uparrow$ & & Backbone & $\Delta$Accuracy & $\Delta$AUROC \\
        \midrule
        \multirow{10}{*}{\bf{Carotid}} & \multirow{3}{*}{Ours-S} & \multirow{3}{*}{4.53} & \multirow{3}{*}{85.39\%} & \multirow{3}{*}{87.45\%} & C1 & CNN & $\uparrow$ 1.40\% & $\uparrow$ 0.031  \\
        & & & & & C2 & CNN & $\uparrow$ 1.12\% & $\uparrow$ 0.009 \\
        & & & & & C3 & Hybrid & $\downarrow$ 0.68\% & $\downarrow$ 0.017 \\
        \cmidrule(lr){2-9}
        & \multirow{3}{*}{Ours-SK} & \multirow{3}{*}{4.75} & \multirow{3}{*}{89.03\%} & \multirow{3}{*}{93.38\%} & C1 & CNN & $\uparrow$ 2.04\% & $\uparrow$ 0.044  \\
        & & & & & C2 & CNN & $\uparrow$ 1.36\% & $\uparrow$ 0.053 \\
        & & & & & C3 & Hybrid & $\uparrow$ 2.72\% & $\uparrow$ 0.062\\
        \cmidrule(lr){2-9}
        & \multirow{3}{*}{Ours-SKM} & \multirow{3}{*}{\bf{4.26}} & \multirow{3}{*}{\bf{91.46\%}} & \multirow{3}{*}{\bf{94.29\%}} & C1 & CNN & \textcolor{delta}{$\uparrow$ 3.40\%} & \textcolor{delta}{$\uparrow$ 0.069} \\
        & & & & & C2 & CNN & \textcolor{delta}{$\uparrow$ 6.80\%} & \textcolor{delta}{$\uparrow$ 0.086} \\
        & & & & & C3 & Hybrid & \textcolor{delta}{$\uparrow$ 4.81\%} & \textcolor{delta}{$\uparrow$ 0.035} \\
        \midrule
        \multirow{10}{*}{\makecell[c]{\bf{TUSC}\\~\citep{TUSC}}} & \multirow{3}{*}{Ours-S} & \multirow{3}{*}{3.76} & \multirow{3}{*}{87.44\%} & \multirow{3}{*}{91.07\%} & C1 & CNN & $\downarrow$ 0.36\% & $\downarrow$ 0.009 \\
        & & & & & C2 & CNN & $\uparrow$ 1.08\% & $\uparrow$ 0.011 \\
        & & & & & C3 & Hybrid & $\uparrow$ 1.44\% & $\uparrow$ 0.007 \\
        \cmidrule(lr){2-9}
        & \multirow{3}{*}{Ours-SK} & \multirow{3}{*}{4.53} & \multirow{3}{*}{91.10\%} & \multirow{3}{*}{95.02\%} & C1 & CNN & $\uparrow$ 0.72\% & $\uparrow$ 0.007 \\
        & & & & & C2 & CNN & $\uparrow$ 2.88\% & $\uparrow$ 0.032 \\
        & & & & & C3 & Hybrid & $\uparrow$ 3.24\% & $\uparrow$ 0.021\\
        \cmidrule(lr){2-9}
        & \multirow{3}{*}{Ours-SKM} & \multirow{3}{*}{\bf{3.69}} & \multirow{3}{*}{\bf{92.14\%}} & \multirow{3}{*}{\bf{95.59\%}} & C1 & CNN & \textcolor{delta}{$\uparrow$ 2.53\%} & \textcolor{delta}{$\uparrow$ 0.030} \\
        & & & & & C2 & CNN & \textcolor{delta}{$\uparrow$ 3.61\%} & \textcolor{delta}{$\uparrow$ 0.040} \\
        & & & & & C3 & Hybrid & \textcolor{delta}{$\uparrow$ 3.57\%} & \textcolor{delta}{$\uparrow$ 0.030} \\
        \midrule
        \multirow{10}{*}{\makecell[c]{\bf{ACDC}\\~\citep{bernard2018deep}}} & \multirow{3}{*}{Ours-S} & \multirow{3}{*}{8.75} & \multirow{3}{*}{70.11\%} & \multirow{3}{*}{87.62\%} & C1 & CNN & $\uparrow$ 0.73\% & $\uparrow$ 0.015 \\
        & & & & & C2 & CNN & $\downarrow$ 0.37\% & $\uparrow$ 0.000 \\
        & & & & & C3 & Hybrid & $\downarrow$ 0.73\% & $\downarrow$ 0.005 \\
        \cmidrule(lr){2-9}
        & \multirow{3}{*}{Ours-SK} & \multirow{3}{*}{8.52} & \multirow{3}{*}{\bf{77.29\%}} & \multirow{3}{*}{92.60\%} & C1 & CNN & $\uparrow$ 1.47\% & $\uparrow$ 0.021 \\
        & & & & & C2 & CNN & $\uparrow$ 1.83\% & $\uparrow$ 0.032 \\
        & & & & & C3 & Hybrid & $\uparrow$ 1.83\% & $\uparrow$ 0.014\\
        \cmidrule(lr){2-9}
        & \multirow{3}{*}{Ours-SKM} & \multirow{3}{*}{\bf{8.18}} & \multirow{3}{*}{77.08\%} & \multirow{3}{*}{\bf{92.77\%}} & C1 & CNN & \textcolor{delta}{$\uparrow$ 2.20\%} & \textcolor{delta}{$\uparrow$ 0.040} \\
        & & & & & C2 & CNN & \textcolor{delta}{$\uparrow$ 2.93\%} & \textcolor{delta}{$\uparrow$ 0.049} \\
        & & & & & C3 & Hybrid & \textcolor{delta}{$\uparrow$ 3.43\%} & \textcolor{delta}{$\uparrow$ 0.029} \\
        \midrule
        \multirow{10}{*}{\makecell[c]{\bf{MosMedData}\\~\citep{morozov2020mosmeddata}}} & \multirow{3}{*}{Ours-S} & \multirow{3}{*}{5.17} & \multirow{3}{*}{75.23\%} & \multirow{3}{*}{89.05\%} & C1 & CNN & $\uparrow$ 1.54\% & $\downarrow$ 0.002 \\
        & & & & & C2 & CNN & $\downarrow$ 0.13\% & $\downarrow$ 0.102 \\
        & & & & & C3 & Hybrid & $\downarrow$ 0.76\% & $\downarrow$ 0.013  \\
        \cmidrule(lr){2-9}
        & \multirow{3}{*}{Ours-SK} & \multirow{3}{*}{2.98} & \multirow{3}{*}{\bf{84.19\%}} & \multirow{3}{*}{96.16\%} & C1 & CNN & $\uparrow$ 2.83\% & $\uparrow$ 0.008\\
        & & & & & C2 & CNN & $\uparrow$ 1.28\% & $\downarrow$ 0.017 \\
        & & & & & C3 & Hybrid & $\uparrow$ 1.41\% & $\uparrow$ 0.022 \\ 
        \cmidrule(lr){2-9}
        & \multirow{3}{*}{Ours-SKM} & \multirow{3}{*}{\bf{2.91}} & \multirow{3}{*}{84.07\%} & \multirow{3}{*}{\bf{96.33\%}} & C1 & CNN & \textcolor{delta}{$\uparrow$ 3.73\%} & \textcolor{delta}{$\uparrow$ 0.028} \\
        & & & & & C2 & CNN & \textcolor{delta}{$\uparrow$ 2.57\%} & \textcolor{delta}{$\uparrow$ 0.005} \\
        & & & & & C3 & Hybrid & \textcolor{delta}{$\uparrow$ 2.70\%} & \textcolor{delta}{$\uparrow$ 0.044} \\
        \midrule
        \multirow{10}{*}{\makecell[c]{\bf{MRNet}\\~\citep{bien2018deep}}} & \multirow{3}{*}{Ours-S} & \multirow{3}{*}{8.46} & \multirow{3}{*}{73.88\%} 
        & \multirow{3}{*}{89.68\%} & C1 & CNN & $\downarrow$ 0.52\% & $\downarrow$ 0.060 \\
        & & & & & C2 & CNN & $\uparrow$ 0.00\% & $\downarrow$ 0.001 \\
        & & & & & C3 & Hybrid & $\uparrow$ 0.53\% & $\uparrow$ 0.004 \\
        \cmidrule(lr){2-9}
        & \multirow{3}{*}{Ours-SK} & \multirow{3}{*}{9.20} & \multirow{3}{*}{76.81\%} & \multirow{3}{*}{93.34\%} & C1 & CNN & $\uparrow$ 2.65\% & $\uparrow$ 0.036\\
        & & & & & C2 & CNN & $\uparrow$ 1.58\% & $\uparrow$ 0.056 \\
        & & & & & C3 & Hybrid & $\uparrow$ 2.12\% & $\uparrow$ 0.029 \\
        \cmidrule(lr){2-9}
        & \multirow{3}{*}{Ours-SKM} & \multirow{3}{*}{\bf{8.14}} & \multirow{3}{*}{\bf{77.10\%}} & \multirow{3}{*}{\bf{93.81\%}} & C1 & CNN & \textcolor{delta}{$\uparrow$ 2.12\%} & \textcolor{delta}{$\uparrow$ 0.030} \\
        & & & & & C2 & CNN & \textcolor{delta}{$\uparrow$ 1.58\%} & \textcolor{delta}{$\uparrow$ 0.009} \\
        & & & & & C3 & Hybrid & \textcolor{delta}{$\uparrow$ 4.76\%} & \textcolor{delta}{$\uparrow$ 0.035} \\
        \bottomrule
    \end{tabular}}
\end{table*}

\subsection{Ablation Study}

\subsubsection{Analysis of Multimodal Conditions Guidance}
To illustrate the role of different conditional guidance, we used five datasets to conduct comparative experiments with generators that resort to various banks of conditions for training and sampling.
Table~\ref{tab:ablation_cond} validates that all proposed conditions are effective for enhancing synthesis and diagnosis tasks.
Specifically, without text control, the downstream diagnostic performance is comparatively poor.
Fig.~\ref{fig:syn_visual}(c-d) compares typical synthetic thyroid nodule sequences generated under condition banks with and without text guidance.
The former is visibly high-fidelity and faithful to the given text prompt (e.g., smooth margin), while the latter exhibits less distinguishable features for diagnosis (e.g., blurry margin).
That is to say, the proposed \textit{\M} can create more diagnosis-reliable samples by enhancing semantic steerability in the generation process.
Besides, by incorporating image prior knowledge in the sampling process, the three classifiers achieve consistent performance improvements on all datasets, with average accuracy/AUROC gains of Carotid (+4.40\%, +0.045), TUSC (+4.20\%, +0.040), ACDC (+1.87\%, +0.019), MosMedData (+1.63\%, +0.054), and MRNet (+2.64\%, +0.017).
This proves the domain gap between synthetic and real samples is mitigated by introducing the image prior.
Moreover, as shown in Table~\ref{tab:ablation_cond}, conditioning the generator on the motion field produces more content-consistent and smoother samples, leading to continuous improvements in diagnostic performance.
It can also be observed that the FVD results show a limited correlation with downstream evaluation metrics, confirming the finding in~\cite{luo2024measurement}.

\subsubsection{Impact of the Sequential Augmentation Module}
To verify the impact of the proposed sequential augmentation module, as shown in Table~\ref{tab:ablation_sam}, we compared the performance of three generator variants on the synthesis and downstream tasks.
Ours-S, Ours-SK, and Ours-SKM denote our ablation studies, including gradually adding sequential attention ('-S'), key-frame/slice attention ('-K'), and motion field attention ('-M') to the generator with spatial inflation only.
Table~\ref{tab:ablation_sam} shows that the holistic downstream accuracy and AUROC improvements of Ours-SKM are significantly higher than those of Ours-S and Ours-SK in five datasets.
In terms of quantitative assessment of sequential coherence, Ours-SKM achieves better results overall than other variants.
These prove that this module can help the generator synthesize cross-frame/slice consistent and dynamic-smooth clips, which are diagnosis-promotive for downstream diagnostic tasks.
Moreover, it can be observed that the FVD results across these three generators show no significant difference on four datasets, revealing its limited value for evaluating diagnosis-oriented medical sequence synthesis tasks.

\begin{figure}[!htbp]
	\centering
    \includegraphics[width=1.0\linewidth]{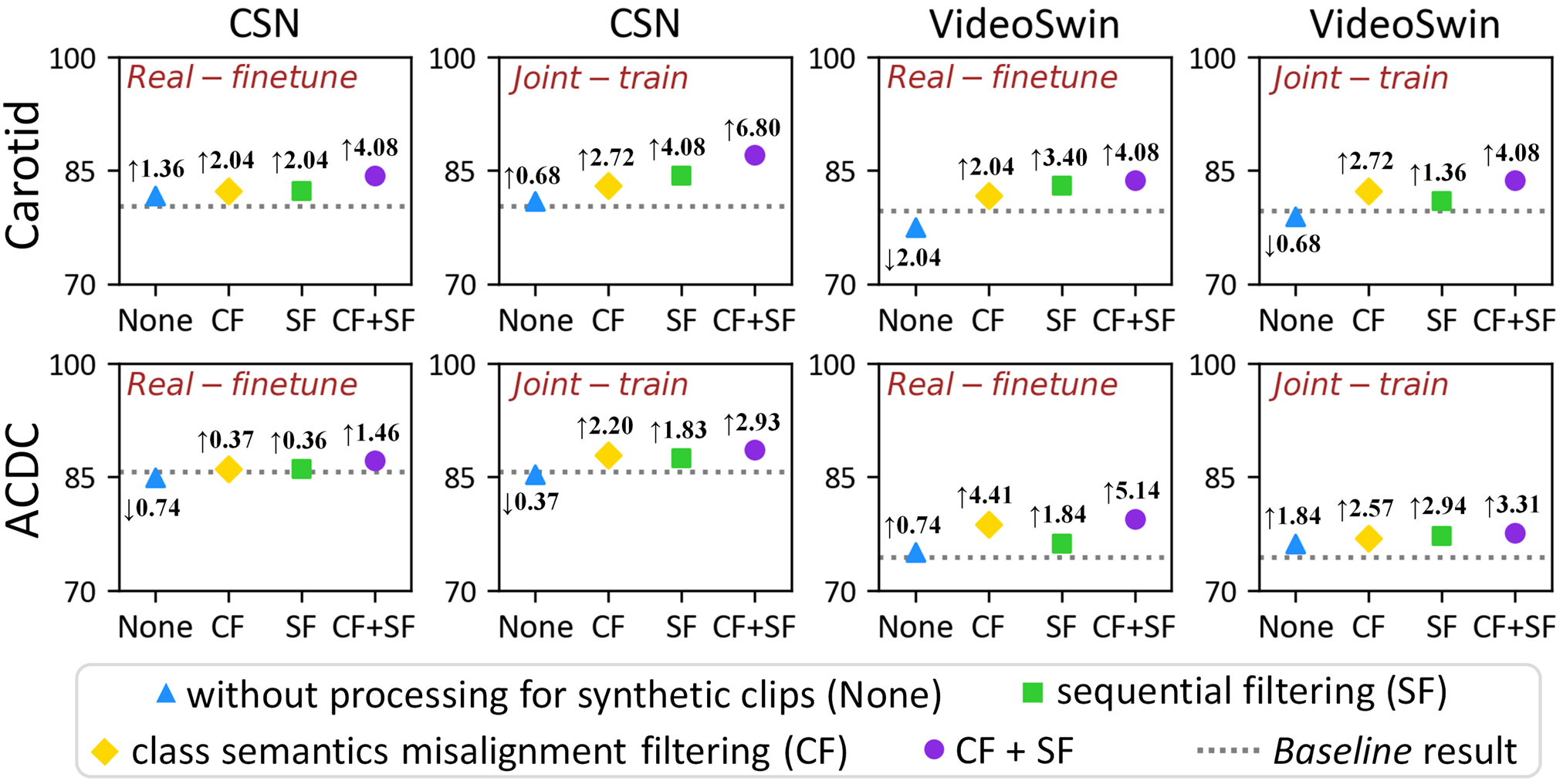}
	\caption{Carotid stenosis and heart disease diagnostic accuracy (\%) comparison using two classifiers on our filtering strategies under \textit{Real-finetune} and \textit{Joint-train} paradigms. The variant 'CF+SF' indicates our final filter, and the accuracy gains over the \textit{Baseline} are depicted.}
	\label{fig:res_data_filter}
\end{figure}

\subsubsection{Effectiveness of the Noisy Synthetic Data Filter}
As shown in Fig.~\ref{fig:res_data_filter}, we validated the indispensable role of our proposed noisy synthetic data filter by assessing its impact on downstream diagnostic accuracy, using two classifiers trained on \textit{Real-finetune} and \textit{Joint-train} paradigms across the carotid and ACDC datasets.
For instance, on the carotid dataset, the \textit{Joint-train} CSN resorting to non-quality controlled synthetic clips (i.e., 'None'), displays no significant difference with its \textit{Baseline} (80.95\% \textit{vs}. 80.27\%), even an accuracy degradation for its VideoSwin counterpart (78.91\% \textit{vs}. 79.59\%).  
Then, by implementing our CF and SF strategy, diagnostic accuracy is significantly improved by 2.04\% and 3.40\%, respectively.
Equipped with both, it achieves an accuracy of 87.07\%, outperforming the 'None' by 6.12\%.

\begin{table*}
    \centering
    \scriptsize
    \caption{Ablation study for performance-oriented diffusion settings in TUSC~\citep{TUSC} and ACDC~\citep{bernard2018deep} datasets. I3D~\citep{carreira2017quo} trained on \textit{Real-finetune} paradigm was used for downstream diagnosis. The blue-highlighted block represents our default configuration.}
    \label{tab:ablation_diff_factor1}
    \resizebox{1.0\textwidth}{!}{
    \begin{tabular}{c c c|c c c|c c}
        \toprule
        \multirow{2.5}{*}{\bf{Dataset}} & \multicolumn{2}{c|}{\bf{Diffusion Settings}} & \multicolumn{3}{c|}{\bf{Synthesis}} & \multicolumn{2}{c}{\bf{Diagnosis}} \\
        \cmidrule(lr){2-8}
        & Noise Schedule & Guidance Scale & FVD$\downarrow$ & VAE-Seq$\uparrow$ & Dynamic Smoothness$\uparrow$ & Accuracy & AUROC \\
        \midrule
        \multirow{5}{*}{\bf{TUSC}} & Cosine & 12.5 & 5.19 & 92.79\% & 95.30\% & 63.24\% & 0.588 \\
        & Cosine & 7.5 & 3.90 & 92.57\% & 95.76\% & 64.69\% & 0.600 \\
        & Scaled Linear & 12.5 & 4.57 & 92.16\% & 95.45\% & 64.33\% & 0.597 \\
        & \cellcolor{cyan!10}Scaled Linear & \cellcolor{cyan!10}7.5 & \cellcolor{cyan!10}3.69 & \cellcolor{cyan!10}92.14\% & \cellcolor{cyan!10}95.59\% & \cellcolor{cyan!10}65.41\% & \cellcolor{cyan!10}0.611 \\
        \midrule
        \multirow{5}{*}{\bf{ACDC}} & Cosine & 12.5 & 8.42 & 73.95\% & 90.71\% & 85.32\% & 0.762 \\
        & Cosine & 7.5 & 8.63 & 73.88\% & 90.74\% & 86.05\% & 0.775 \\
        & Scaled Linear & 12.5 & 8.16 & 76.75\% & 92.56\% & 86.06\% & 0.774 \\
        & \cellcolor{cyan!10}Scaled Linear & \cellcolor{cyan!10}7.5 & \cellcolor{cyan!10}8.18 & \cellcolor{cyan!10}77.08\% & \cellcolor{cyan!10}92.77\% & \cellcolor{cyan!10}87.16\% & \cellcolor{cyan!10}0.795 \\
        \bottomrule
    \end{tabular}}
\end{table*}

\begin{table*}
    \centering
    \caption{Ablation study for efficiency-performance trade-off across diffusion settings in TUSC~\citep{TUSC} and ACDC~\citep{bernard2018deep} datasets. I3D~\citep{carreira2017quo} trained on \textit{Real-finetune} paradigm was used for downstream diagnosis. 
    All diffusion models are tested on a single NVIDIA GeForce RTX 3090 GPU.
    The blue-highlighted block represents our default configuration. DS, Dynamic Smoothness.}
    \label{tab:ablation_diff_factor2}
    \resizebox{1.0\textwidth}{!}{
    \begin{tabular}{c c c|c c c c|c c}
        \toprule
        \multirow{2.5}{*}{\bf{Dataset}} & \multicolumn{2}{c|}{\bf{Diffusion Settings}} & \multicolumn{4}{c|}{\bf{Synthesis}} & \multicolumn{2}{c}{\bf{Diagnosis}} \\
        \cmidrule(lr){2-9}
        & UNet Depth & Sampling Steps & FVD$\downarrow$ & VAE-Seq$\uparrow$ & DS$\uparrow$ & Sampling Time (per sequence)$\downarrow$ & Accuracy & AUROC \\
        \midrule
        \multirow{5}{*}{\bf{TUSC}} & 3 & 200 & 13.26 & 86.29\% & 90.30\% & 53.5s & 62.52\% & 0.575 \\
        & 3 & 300 & 12.33 & 83.42\% & 88.19\% & 87.4s & 62.88\% & 0.580 \\
        & 4 & 100 & 3.70 & 92.02\% & 95.53\% & 39.8s & 63.97\% & 0.592 \\
        & \cellcolor{cyan!10}4 & \cellcolor{cyan!10}200 & \cellcolor{cyan!10}3.69 & \cellcolor{cyan!10}92.14\% & \cellcolor{cyan!10}95.59\% & \cellcolor{cyan!10}68.8s & \cellcolor{cyan!10}65.41\% & \cellcolor{cyan!10}0.611 \\
        & 4 & 300 & 2.69 & 91.15\% & 95.23\% & 100.3s & 65.77\% & 0.622 \\
        \midrule
        \multirow{5}{*}{\bf{ACDC}} & 3 & 200 & 10.81 & 75.15\% & 88.73\% & 51.3s & 83.85\% & 0.741 \\
        & 3 & 300 & 10.17 & 73.52\% & 88.41\% & 79.9s & 83.85\% & 0.739 \\
        & 4 & 100 & 8.20 & 76.86\% & 92.74\% & 33.7s & 85.69\% & 0.771 \\
        & \cellcolor{cyan!10}4 & \cellcolor{cyan!10}200 & \cellcolor{cyan!10}8.18 & \cellcolor{cyan!10}77.08\% & \cellcolor{cyan!10}92.77\% & \cellcolor{cyan!10}70.7s & \cellcolor{cyan!10}87.16\% & \cellcolor{cyan!10}0.795 \\
        & 4 & 300 & 7.55 & 74.59\% & 91.90\% & 103.8s & 88.06\% & 0.806 \\
        \bottomrule
    \end{tabular}}
\end{table*}

\subsubsection{Impact of Diffusion Factors in Sequence Generator}
To systematically analyze how the key diffusion factors affect the quality of the synthetic databases, we conducted two ablation studies that examine performance-oriented diffusion settings, including the noise schedule and guidance scale (Table~\ref{tab:ablation_diff_factor1}), as well as efficiency-performance trade-off analyses under the settings of different UNet depths and sampling steps (Table~\ref{tab:ablation_diff_factor2}).

As shown in Table~\ref{tab:ablation_diff_factor1}, the scaled linear noise schedule consistently yields stronger downstream augmentation benefits and also achieves lower FVD than the cosine variant across both datasets.
We believe that this advantage stems from its smoother and more stable noise evolution during training, which allows the model to more effectively learn and preserve anatomical details throughout the diffusion process.
Moreover, the results show that using a high guidance scale of 12.5 leads to a noticeable degradation in diagnostic performance compared with the moderate setting of 7.5, with accuracy decreasing from 65.41\% to 64.33\% and AUROC dropping from 0.611 to 0.597 under the scaled linear schedule on the TUSC dataset.
This trend suggests that excessive guidance may constrain the sampling trajectory, contributing to in-distribution (i.e., overly close to the real training distribution) synthetic samples, which in turn limit their usefulness for downstream model optimization.
In contrast, a moderate guidance scale strikes a more favorable balance between semantic alignment and diversity, enabling the synthesizer to generate more diagnosis-promotive samples.

The depth of the UNet denoiser and the number of sampling steps both exert dual impacts on overall performance and synthesis efficiency.
Table~\ref{tab:ablation_diff_factor2} demonstrates that increasing the UNet depth consistently improves generation fidelity and downstream classification, since deeper architectures can capture richer anatomical patterns and finer dynamic cues during training. 
However, these performance gains come with additional computational cost.
A similar trend is observed for sampling steps, where additional iterations yield more diagnosis-reliable sequences but also require longer sampling time.
Notably, configurations with insufficient feature representation capacity, such as a UNet depth of three, benefit minimally from increasing the sampling steps, indicating that additional steps cannot compensate for limited feature extraction ability. 
We further observe that the sequential consistency and smoothness of our synthetic samples are substantially more sensitive to the UNet depth than to other diffusion factors, as it directly determines the network’s capacity to model and propagate temporal/stereoscopic dependencies.
Overall, our default configuration achieves a balanced compromise between generation quality and inference efficiency, providing a practical choice for large-scale customized medical sequence synthesis and generative augmentation.

\begin{table*}
    \centering
    \caption{Sensitivity analysis on four representative classifier backbones used in the class semantics misalignment filtering across the Carotid, ACDC~\citep{bernard2018deep}, and MosMedData~\citep{morozov2020mosmeddata} datasets. I3D~\citep{carreira2017quo} trained on \textit{Real-finetune} and \textit{Joint-train} paradigms was adopted for downstream diagnosis. The accuracy and AUROC improvements over the \textit{Baseline} are reported, with the \underline{best gains} underlined.}
    \label{tab:ablation_filter_sensitivity}
    \resizebox{1.0\textwidth}{!}{
    \begin{tabular}{c c c c c c c c}
        \toprule
        \multirow{2.5}{*}{\bf{Training Paradigm}} & \multirow{2.5}{*}{\bf{Semantic Filtering Classifier}} & \multicolumn{2}{c}{\bf{Carotid}} & \multicolumn{2}{c}{\bf{ACDC}} & \multicolumn{2}{c}{\bf{MosMedData}} \\
        \cmidrule(lr){3-4} \cmidrule(lr){5-6} \cmidrule(lr){7-8}
        & & $\Delta$Accuracy & $\Delta$AUROC & $\Delta$Accuracy & $\Delta$AUROC & $\Delta$Accuracy & $\Delta$AUROC \\
        \midrule
        \multirow{5}{*}{\it{Real-finetune}} & CSN & $\uparrow$ 4.08\% & $\uparrow$ 0.064 & $\uparrow$ 2.56\% & $\uparrow$ 0.041 & $\uparrow$ 4.76\% & $\uparrow$ 0.027 \\
        \addlinespace[2.5pt]
        & VideoSwin & $\uparrow$ 4.08\% & $\uparrow$ 0.066 & $\uparrow$ 1.83\% & $\uparrow$ 0.035 & $\uparrow$ \underline{5.66}\% & $\uparrow$ 0.035 \\
        \addlinespace[2.5pt]
        & CSG-3DCT & $\uparrow$ 5.44\% & $\uparrow$ 0.056 & $\uparrow$ 2.57\% & $\uparrow$ 0.043 & $\uparrow$ 4.25\% & $\uparrow$ 0.018 \\
        \addlinespace[2.5pt]
        & I3D & $\uparrow$ \underline{5.44}\% & $\uparrow$ \underline{0.076} & $\uparrow$ \underline{3.67}\% & $\uparrow$ \underline{0.062} & $\uparrow$ 5.41\% & $\uparrow$ \underline{0.043} \\
        \midrule
        \multirow{5}{*}{\it{Joint-train}} & CSN & $\uparrow$ 3.40\% & $\uparrow$ 0.042 & $\uparrow$ 1.10\% & $\uparrow$ 0.021 & $\uparrow$ 2.32\% & $\uparrow$ 0.007 \\
        \addlinespace[2.5pt]
        & VideoSwin & $\uparrow$ 4.76\% & $\uparrow$ 0.045 & $\uparrow$ 1.47\% & $\uparrow$ 0.027 & $\uparrow$ \underline{2.58}\% & $\uparrow$ 0.005 \\
        \addlinespace[2.5pt]
        & CSG-3DCT & $\uparrow$ 4.08\% & $\uparrow$ 0.071 & $\uparrow$ 1.83\% & $\uparrow$ 0.028 & $\uparrow$ 2.45\% & $\uparrow$ 0.006 \\
        \addlinespace[2.5pt]
        & I3D & $\uparrow$ \underline{4.76}\% & $\uparrow$ \underline{0.074} & $\uparrow$ \underline{1.83}\% & $\uparrow$ \underline{0.029} & $\uparrow$ 2.32\% & $\uparrow$ \underline{0.019} \\
        \bottomrule
    \end{tabular}}
\end{table*}

\subsubsection{Analysis of Classifier Backbone Choice in the Semantic Filter}
To investigate whether the observed diagnostic performance gains depend on the choice of classifier backbone used in the semantics misalignment filtering stage (refer to Sec.~\ref{semantic_filter}), we performed sensitivity analysis using four backbones across three datasets (Table~\ref{tab:ablation_filter_sensitivity}).
Specifically, we fixed the downstream classifier to I3D and selected four representative backbones in the semantic filtering stage.
It can be observed that, under both training paradigms on three datasets, the diagnostic performance consistently improves with all backbone selections, and their gains exhibit only marginal variations (with deviations in accuracy gains of at most 1.84\% relative to our default setting).
Overall, these results demonstrate that the performance benefits are not tied to a particular semantic filtering backbone, affirming the robustness and general applicability of our method.

\begin{figure*}[!t]
	\centering
    \includegraphics[width=0.85\linewidth]{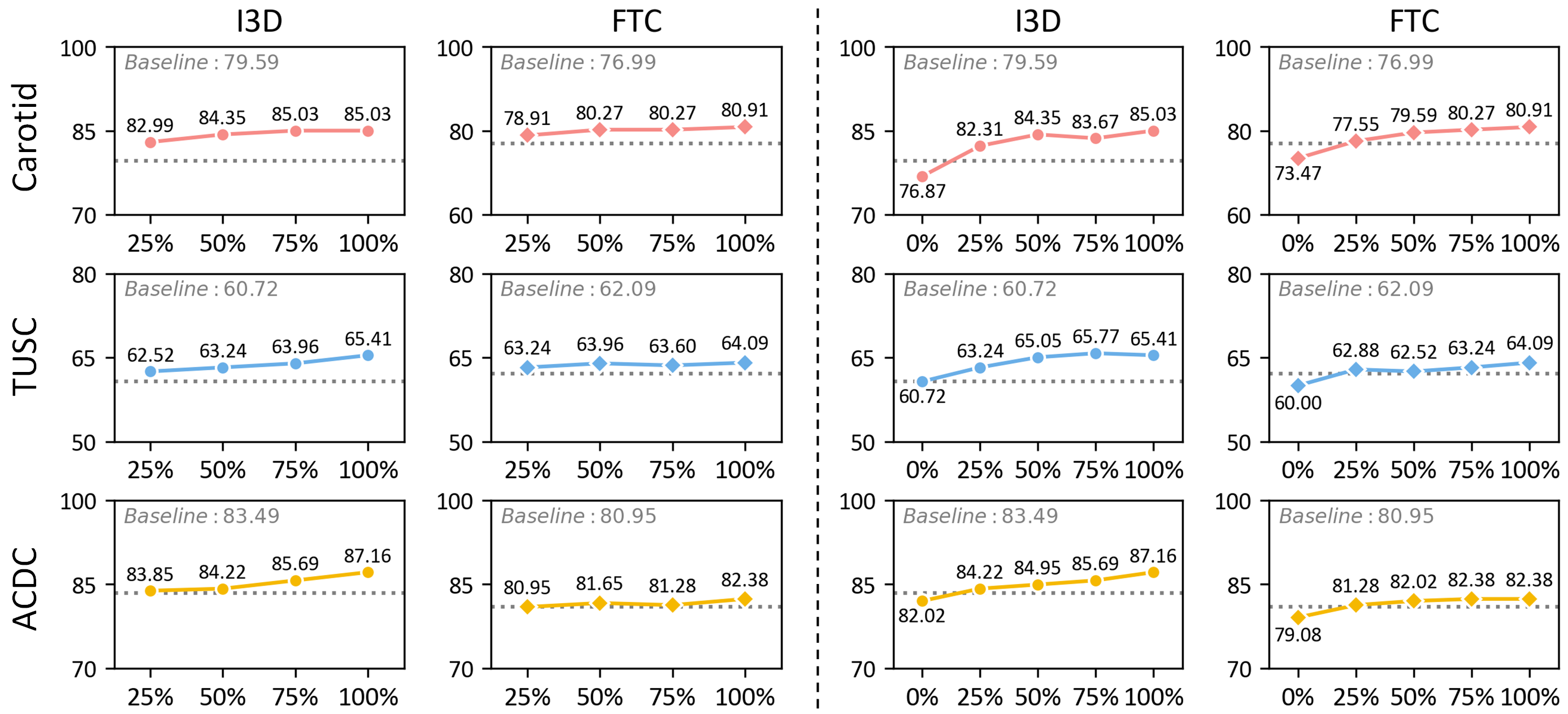}
	\caption{Diagnostic accuracy (\%) comparison using I3D~\citep{carreira2017quo} and FTC~\citep{ahmadi2023transformer} on three datasets under two annotation settings, both trained under the \textit{Real-finetune} paradigm. \textit{Left:} generator training with varying proportions of class-labeled real data. \textit{Right:} downstream classifiers trained with different fractions of real labeled data. The gray dashed line indicates the \textit{Baseline} result. Our \textit{\M} can substantially reduce the overall annotation burden across both the generator training (\textit{Left}) and downstream diagnosis stages (\textit{Right}).}
	\label{fig:labeled_data_ratio}
\end{figure*}

\subsubsection{Analysis of Annotation Workload Reduction}
To investigate the advantages of \textit{\M} in reducing annotation burden, we conducted two dedicated ablation studies, including 1) examining the synthesizer’s dependency on real annotations during training (Figure~\ref{fig:labeled_data_ratio}, \textit{left}), and 2) assessing whether the trained generator can reduce labeling demand in downstream applications (Figure~\ref{fig:labeled_data_ratio}, \textit{right}).
For the \textbf{first study}, we trained the generator using 25\%, 50\%, 75\%, and 100\% of the class-labeled real data, and subsequently used the corresponding synthetic datasets to compare their impact on downstream diagnostic accuracy under identical classifier settings.
For the \textbf{second study}, we first obtained a synthetic dataset using the generator trained on all real labeled data, and then trained downstream classifiers with 25\%, 50\%, 75\%, and 100\% of the class-labeled real data under the \textit{Real-finetune} paradigm to compare the resulting diagnostic accuracy.
We also report the performance of classifiers trained exclusively on synthetic data (i.e., using 0\% real annotated samples).

As shown in the left graph of Figure~\ref{fig:labeled_data_ratio}, across all three datasets, synthetic data produced by a generator trained with only 25\% of the real labeled cases lead to competitive downstream performance, with most results surpassing the \textit{Baseline} and the largest accuracy gain reaching 3.40\%.
These results indicate that the proposed sequence generator can be effectively trained with only a small fraction of labeled samples, thereby substantially alleviating the annotation burden during the synthesizer training stage.
Besides, as shown in the right graph of Figure~\ref{fig:labeled_data_ratio}, when the availability of real labeled data becomes limited, the synthetic dataset produced by \textit{\M} continues to provide noticeable performance gains.
For instance, with only a 25\% real-labeled training subset, the synthetic data improve the I3D by +2.52\% accuracy (60.72\%$\rightarrow$63.24\%).
Notably, even without any real labeled data, training solely on synthetic sequences yields a comparable performance to the \textit{Baseline}.
These findings highlight that our customized synthetic samples can significantly ease the annotation workload in downstream applications, underscoring their great potential to complement real data in label-sparse scenarios.

\section{Conclusion and discussion}
In this study, we present a new and general diffusion-based generative augmentation framework, named \textit{\M}, to facilitate medical sequence classification by leveraging customized and diagnosis-reliable synthetic sequences.
To the best of our knowledge, this is the first comprehensive empirical study on controllable generative augmentation for medical sequence classification.

To improve the quality of synthetic data for promoting downstream classification, we propose a multimodal conditions-guided medical sequence generator that ensures flexibly controllable synthesis across semantic, sequential, and data distribution aspects.
Moreover, we propose a highly effective noisy synthetic data filter to better learn from synthetic data, including adaptively filtering diagnosis-inhibitive synthetic sequences at class semantics and sequential levels.
This can better connect the synthesis task and the downstream one, thus further enhancing our sequence classification performance.
Extensive experiments on 5 medical datasets spanning 4 distinct modalities, including comparisons against 15 augmentation strategies and evaluations using 11 classifiers trained on 3 paradigms, comprehensively validate the robustness and generality of \textit{\M}.
Furthermore, the empirical analysis validates that \textit{\M} can be effectively leveraged to improve diagnostic performance in underrepresented high-risk populations and out-domain robustness.
We believe that \textit{\M} can serve as a powerful and practical data augmentation tool for various clinical scenarios.

Even though our experimental results demonstrate that synthetic data can effectively enhance medical AI models, we recognize that clinical adoption ultimately depends on rigorous validation showing that generated sequences are diagnosis-reliable and interpretable to radiologists.
Expert evaluation provides an intuitive starting point, where radiologists assess fidelity, sequential coherence, and adherence to conditioning cues~\citep{duan2025fetalflex}.
Recording clear provenance for each synthetic sequence (such as its associated disease category and image priors) further enhances transparency and interpretability, which is particularly important for conditional generative methods.
However, human review alone cannot scale to large synthetic databases, underscoring the need for automated and standardized quality-assessment pipelines.
Building on prior insights~\citep{huang2024vbench}, we believe that a hierarchical and multidimensional evaluation suite tailored to medical sequence generation, complemented by small-scale user studies to ensure alignment between automated metrics and radiologists' judgments, is urgently needed in the future to guarantee data quality and establish clinical credibility.
This represents an interesting avenue for future exploration.

It is also worth mentioning that our proposed noisy synthetic data filter contributes an initial effort toward assessing the reliability of synthetic samples before being used in downstream diagnosis, and it can be seamlessly integrated into future large-scale quality-assessment pipelines.
Ultimately, we suggest that synthetic data should complement rather than replace real clinical data, and its practical use relies on robust quality metrics, rigorous validation protocols, and appropriate regulatory oversight to ensure trustworthy and effective integration into medical AI workflows.

We further discuss the limitations of \textit{\M}.
First, the denoising diffusion implicit model (DDIM) sampling process is time-consuming due to the large inference steps.
This limits the clinical practicality of \textit{\M} for fast and large-scale production.
Second, in out-domain conditions, the proposed framework requires target domain data during generator training, which can be tough to obtain in advance in clinical practice.
In future work, we will adopt fast-sampling strategies (e.g., AMED-Solver~\citep{zhou2024fast}) to better balance the trade-off between sampling time and sample quality.
Besides, we will incorporate test-time adaptation methods~\citep{huang2023test,luo2023recon,huang2023fourier,zhang2024unsupervised} to provide a more user-friendly tool for out-domain applications in clinical settings.
Last, we will further extend the proposed method to segmentation and detection tasks~\citep{huang2025flip,tao2025enhancing} and work toward establishing a comprehensive benchmark suite tailored for medical sequence generative models.

\myparagraph{Data Availability Statement.}
Our data includes four publicly available datasets: \textbf{TUSC}~\citep{TUSC}, \textbf{ACDC}~\citep{bernard2018deep}, \textbf{MosMedData}~\citep{morozov2020mosmeddata}, and \textbf{MRNet}~\citep{bien2018deep},
as well as a private \textbf{carotid} dataset, which is available from the corresponding author upon reasonable request.

\myparagraph{Acknowledgments.}
This work was supported by the grant from National Natural Science Foundation of China (Nos. 12326619, 62471313, 62471305), Science and Technology Planning Project of Guangdong Province (Nos. 2023A0505020002, 2024A1515030143), Frontier Technology Development Program of Jiangsu Province (No. BF2024078), Guangdong Basic and Applied Basic Research Foundation (Nos. 2025A1515011448, 2025A1515011821), Shenzhen Science and Technology Program (No. JCYJ20240813143302004), Shenzhen Natural Science Foundation (No. JCYJ20250604181708011), the Royal Academy of Engineering, United Kingdom under the RAEng Chair in Emerging Technologies (INSILEXCiET1919/19), the ERC Advanced Grant UKRI Frontier Research Guarantee (INSILICO EP/Y030494/1), the UK Centre of Excellence on in-silico Regulatory Scienceand Innovation (UK CEiRSI) (10139527), the National Institute for Health and CareResearch (NIHR) Manchester Biomedical Research Centre (BRC) (NIHR203308), the BHF Manchester Centre of Research Excellence (RE/24/130017), and the CRUKRadNet Manchester (C1994/A28701).

{\small
\bibliographystyle{spbasic}
\bibliography{egbib}
}

\newpage

\setcounter{table}{0}
\setcounter{figure}{0}
\renewcommand{\thetable}{S\arabic{table}}  
\renewcommand{\thefigure}{S\arabic{figure}} 

\begin{figure*}[t]
    \centering
    \LARGE Supplementary Material
    \vspace{1em}
\end{figure*}

\begin{figure*}[!b]
    \centering
    \includegraphics[width=0.88\linewidth]{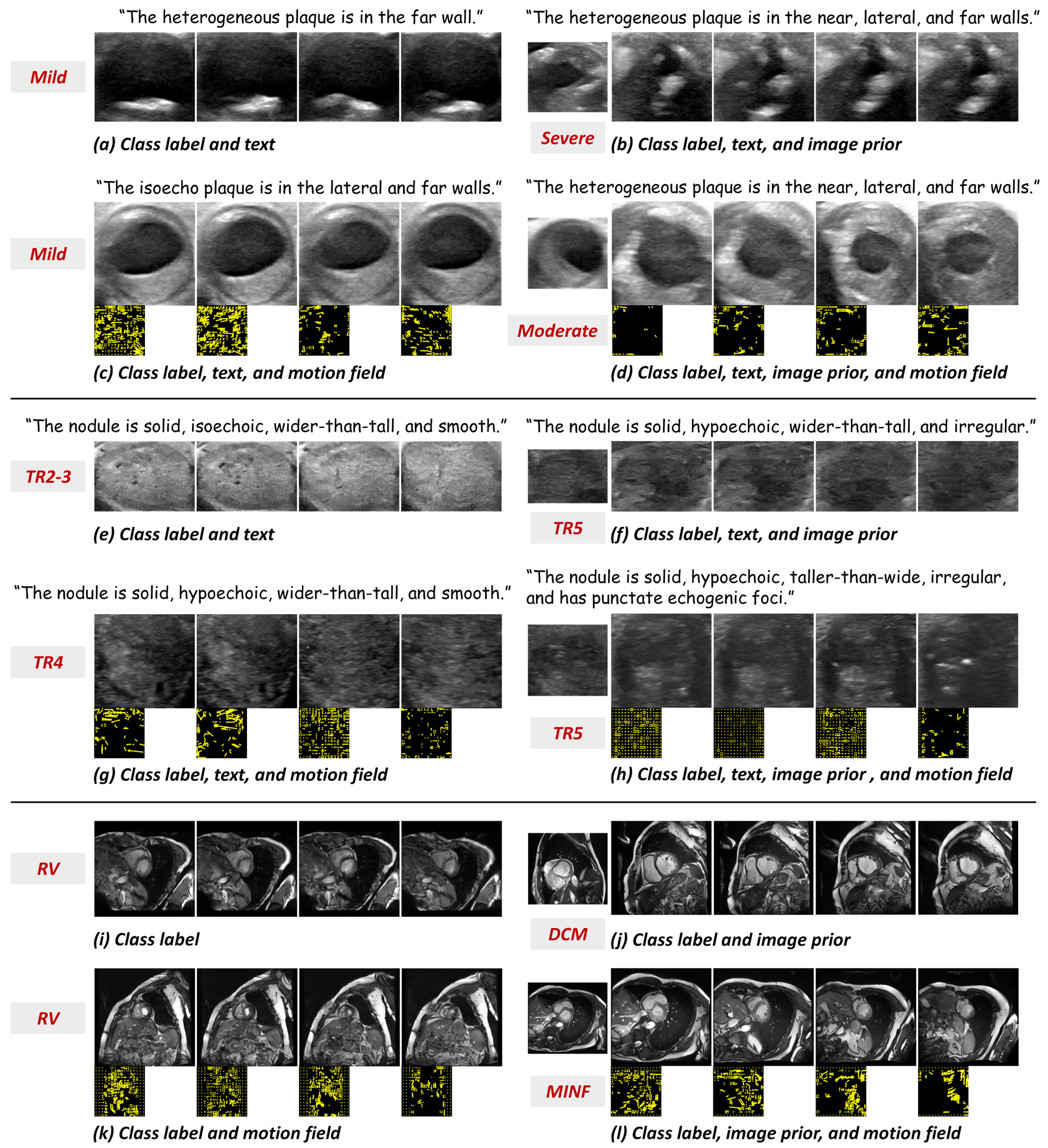}	
    \caption{Additional synthetic results on Carotid (a-d), TUSC~\citep{TUSC} (e-h), and ACDC~\citep{bernard2018deep} (i-l). We used the multimodal conditions joint training strategy~\citep{huang2023composer} during training and flexibly dropped several during sampling.}
\end{figure*}

\begin{figure*}[!t]
    \centering
    \includegraphics[width=0.88\linewidth]{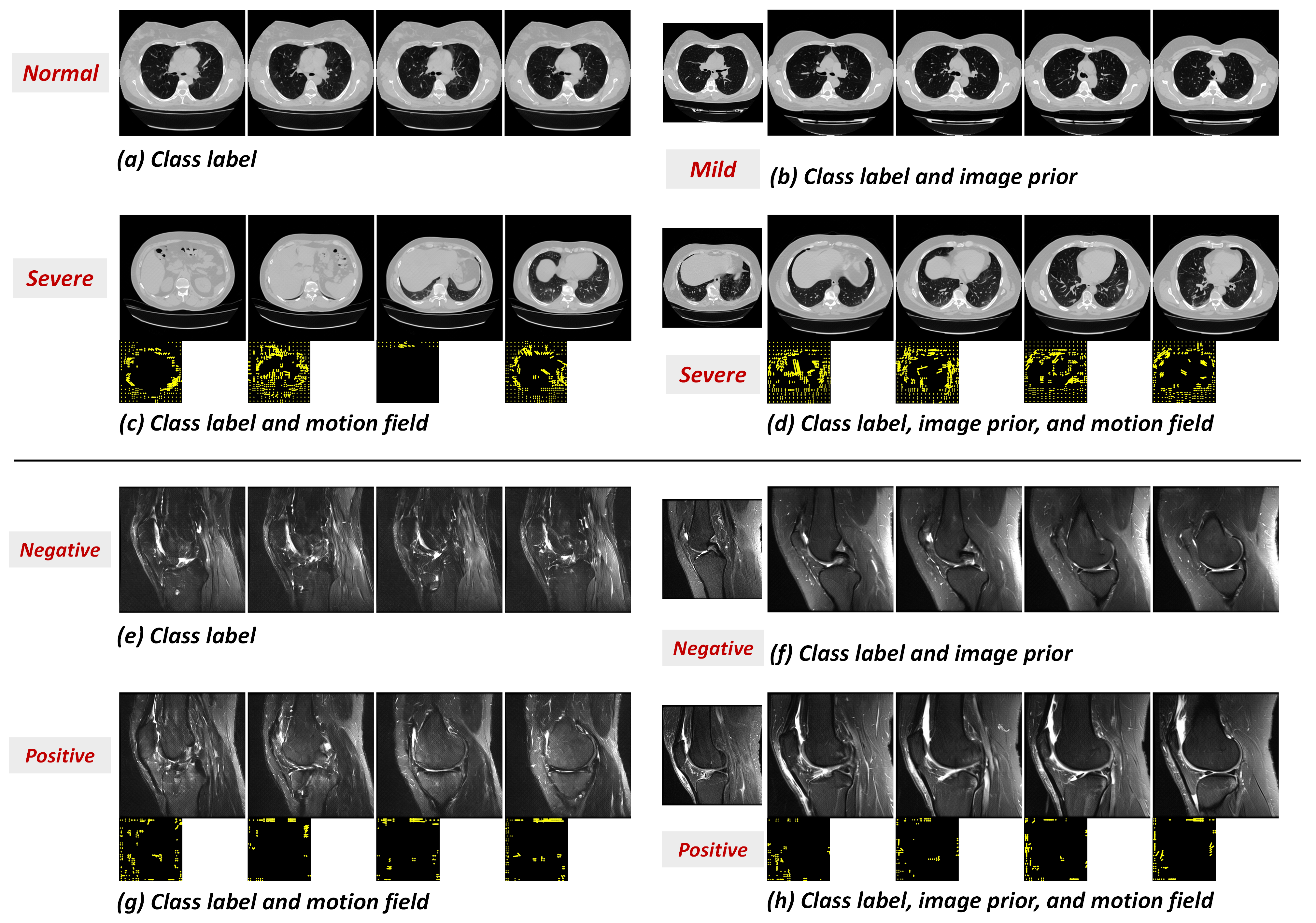}	
    \caption{Additional synthetic results on MosMedData~\citep{morozov2020mosmeddata} (a-d) and MRNet~\citep{bien2018deep} (e-h). We used the multimodal conditions joint training strategy~\citep{huang2023composer} during training and flexibly dropped several during sampling.}
\end{figure*}

\clearpage

\begin{table*}[!t]
  \centering
  \caption{A brief review of studies in the medical field utilizing diffusion-based generative models to promote downstream tasks. We consider six aspects: the conditional control, the downstream task, the underlying architecture of the downstream network, the number of the downstream network, and the modality and type of the medical dataset on which the surveyed approaches were applied. Arch., architecture. Num., number. OCT, optical coherence tomography.}
  \label{tab:addlabel}
  \resizebox{1.0\textwidth}{!}{
    \begin{tabular}{p{3.0cm}|p{3.8cm}|p{1.8cm}|p{1.8cm}|p{1.8cm}|p{2.7cm}|p{1.6cm}}
    \hline
    \bf{Study} & \bf{Conditional Control} & \bf{Downstream Task} & \bf{Downstream Arch.} & \bf{Downstream Net Num.} & \bf{Medical Modality} & \bf{Data Type} \\
    \hline
    \cite{shang2023synfundus} & disease class, \newline image readability & classification & CNN-based, \newline Transformer-based & 2 & Fundus & image \\
    \hline
    \cite{farooq2024derm} & text & classification & CNN-based, \newline Transformer-based & 2 & Dermoscopy & image \\
    \hline
    \cite{zhang2024diffboost} & text, edge & segmentation & CNN-Transformer hybrid-based & 1 & US, CT, MRI & image \\
    \hline
    \cite{sagers2023augmenting} & text & classification & CNN-based & 1 & Dermoscopy & image \\
    \hline
    \cite{khosravi2024synthetically} & attribute, 14 pathology labels & classification & CNN-based & 1 & X-ray & image \\
    \hline
    \cite{yu2024knowledge} & pathology labels, \newline lesion bounding boxes, \newline device types, \newline pathology-specific knowledge (NCM/CAL labels) & classification & Transformer-based & 1 & US & image \\
    \hline
    \cite{ktena2024generative} & disease class, attribute & classification & CNN-based & 1 & Histopathology, \newline X-ray, \newline Dermoscopy & image \\
    \hline
    \cite{luo2024measurement} & disease class & classification & CNN-based & 10 & Fundus, \newline Dermoscopy, \newline X-ray, \newline US & image \\
    \hline
    \cite{li2024endora} & multi-scale latent \newline representation & classification & Transformer-based & 1 & Endoscopy & video \\
    \hline
    \cite{gong2024diffuse} & mask & segmentation & CNN-based & 1 & MRI & volume \\
    \hline
    \cite{peng2024optimizing} & mask & segmentation & CNN-based, CNN-Transformer hybrid-based & 3 & CT & volume \\
    \hline
    \cite{kebaili20243d} & tumor attribute, mask & segmentation & CNN-based & 1 & MRI & volume \\
    \hline
    \cite{yu2024ct} & mask & segmentation & CNN-based & 1 & CT & volume \\
    \hline
    \cite{chen2024towards} & mask & segmentation & CNN-based, \newline CNN-Transformer hybrid-based & 3 & CT & volume \\
    \hline
    \cite{dorjsembe2024conditional} & mask & segmentation & CNN-based & 1 & MRI & volume \\
    \hline
    \cite{huang2024memory} & mask & segmentation & CNN-based & 2 & OCT & volume \\
    \hline
    Ours & disease class, text, \newline image prior, motion field & classification & CNN-based, Transformer-based, CNN-Transformer hybrid-based & 11 & US, MRI & video, \newline volume \\
    \hline
    \end{tabular}
    }
\end{table*}

\end{document}